\documentclass[11pt]{article}
\usepackage[margin=1in]{geometry}

\usepackage{amsmath,amsthm,amssymb}
\usepackage{commath,mathtools}
\usepackage{algorithm,algorithmic}
\usepackage{graphicx}
\usepackage{subcaption}
\usepackage{hyperref}
\usepackage{color}
\usepackage{amsfonts}
\usepackage{epstopdf}
\usepackage{algorithmic}

\usepackage{enumitem}
\usepackage{commath,mathtools}
\usepackage{algorithm,algorithmic}
\usepackage{braket,amsfonts,amsopn}
\usepackage{graphicx,subcaption,hyperref,color}
\usepackage[sort]{cite}
\usepackage{tikz}
\usepackage{booktabs}
\usepackage{bm}
\usepackage{multirow}

\usepackage{mathfont}

\newtheorem{theorem}{Theorem}
\newtheorem{proposition}{Proposition}[theorem]

\newcommand{\algrule}[1][.2pt]{\par\vskip.5\baselineskip\hrule height #1\par\vskip.5\baselineskip}

\title{Influence Estimation and Maximization via Neural Mean-Field Dynamics
}

\author{
Shushan He\thanks{Department of Mathematics and Statistics, Georgia State University, Atlanta, Georgia 30303, USA. (\url{she4@student.gsu.edu}).}
\and
Hongyuan Zha\thanks{School of Data Science, Shenzhen Research Institute of Big Data, The Chinese University of Hong Kong, Shenzhen, Guangdong, China, 518172. (\url{zhahy@cuhk.edu.cn}).}  
\and
Xiaojing Ye\thanks{Department of Mathematics and Statistics, Georgia State University, Atlanta, Georgia 30303, USA. (\url{xye@gsu.edu}).}
}

\usepackage{amsopn}
\DeclareMathOperator{\diag}{diag}

\newcommand{\ex}{\mathbb{E}}
\newcommand{\pr}{\mathrm{Pr}}

\DeclareMathOperator*{\argmin}{\mathrm{arg\,min}}

\providecommand{\keywords}[1]
{
  \small	
  \textbf{\textit{Keywords---}} #1
}

\date{}

\begin{document}

\maketitle

% REQUIRED
\begin{abstract}
We propose a novel learning framework using neural mean-field (NMF) dynamics for inference and estimation problems on heterogeneous diffusion networks. Our new framework leverages the Mori-Zwanzig formalism to obtain an exact evolution equation of the individual node infection probabilities, which renders a delay differential equation with memory integral approximated by learnable time convolution operators. Directly using information diffusion cascade data, our framework can \emph{simultaneously} learn the structure of the diffusion network and the evolution of node infection probabilities. Connections between parameter learning and optimal control are also established, leading to a rigorous and implementable algorithm for training NMF. Moreover, we show that the projected gradient descent method can be employed to solve the challenging influence maximization problem, where the gradient is computed extremely fast by integrating NMF forward in time just once in each iteration. Extensive empirical studies show that our approach is versatile and robust to variations of the underlying diffusion network models, and significantly outperform existing approaches in accuracy and efficiency on both synthetic and real-world data.
\end{abstract}


% REQUIRED
%\begin{AMS}
%68U35, 65D99, 65Z05.
%\end{AMS}

\section{Introduction}
Continuous-time information diffusion on heterogenous networks is a prevalent phenomenon \cite{boguna2002epidemic,newman2010networks,pastor-satorras2015epidemic}.
News spreading on social media \cite{du2013scalable, farajtabar2016multistage,vergeer2013online}, viral marketing \cite{kempe2003maximizing,kempe2005influential,wortman2008viral}, computer malware propagation, and 
epidemics of contagious diseases \cite{bodo2016sis,miller2014epidemic,pastor-satorras2015epidemic,sahneh2011epidemic} are all examples of diffusion on networks, among many others.
For instance, a piece of information (such as a tweet) can be retweeted by users (nodes) with followee-follower relationships (edge) on the Twitter network.
We call a user \emph{infected} if she retweets, and her followers see her retweet and can also become infected if they retweet in turn, and so on.
Such information diffusion mimics the epidemic spread where an infectious virus can spread to individuals (human, animal, or plant) and then to many others upon their close contact.
The study of heterogeneous diffusion networks only emerged in the past decade and is considered very challenging, mainly because of the extremely large scale of modern networks, the heterogeneous inter-dependencies between the nodes, and the randomness exhibited in cascade data.

In the remainder of this section, we provide the mathematical formulations of the inference, influence estimation, and influence maximization problems on an arbitrary diffusion network.
Throughout this paper, we use boldfaced lower (upper) letter to denote vector (matrix) or vector-valued (matrix-valued) function, and $(\cdot)_k$ (or $(\cdot)_{ij}$) for its $k$th component (or $(i,j)$-th entry). All vectors are column vectors unless otherwise noted. We follow the Matlab syntax and use $[\xbm;\ybm]$ to denote the vector that stacks $\xbm$ and $\ybm$ vertically, and $\xbm \cdot \ybm$ or $\xbm^{\top}\ybm$ for the inner product. Time is denoted by $t$ in either continuous ($t\in[0,T]$) or discrete case ($t=0,1,\dots,T$) for some time horizon $T \in \mathbb{R}_+$ ($\mathbb{N}$ in discrete case). Derivative $'$ is with respect to $t$, and gradient $\nabla_{\xbm}$ is with respect to $\xbm$. Probability is denoted by $\pr(\cdot)$, and expectation with respect to $X$ (or its distribution function $p_X$) is denoted by $\ex_{X}[\,\cdot\,]$. The $n$-vectors $\onebm_n,\zerobm_n \in \mathbb{R}^{n}$ stand for the vectors of ones and zeros respectively, and we often omit the subscript $n$ when their dimensions are obvious from the context.

\subsection{Diffusion network models}
Consider a diffusion network model, which consists of a network (directed graph) $\Gcal=(\Vcal,\Ecal)$ with node set $\Vcal=[n]:=\{1,\dots,n\}$ and edge set $\Ecal\subset \Vcal\times \Vcal$, and a \emph{diffusion model} that describes the distribution $p(t;\alpha_{ij})$ of the time $t$ that an infected node $i$ takes to infect her healthy neighbor $j \in \{j':(i,j')\in \Ecal\}$. Here $\alpha_{ij}$ is the infection rate of $i$ on $j$ which \emph{vary across different edges}. That is, $t_{ij}$ is a random variable following the density function $p(t;\alpha_{ij})$ for each $(i,j) \in \Ecal$. We assume that the infection is \emph{progressive}, i.e., a node will not be infected again nor recover once infected, since generalization to the case with recovery is straightforward.
Then, given a source set $\Scal$ (a subset of $\Vcal$) of nodes that are infected at time $0$, they will infect their healthy neighbors with random infection times described above; and the infected neighbors will then infect their healthy neighbors, and so on. As such, the infection initiated by $\Scal$ at time $0$ propagates to other nodes of the network. 
We call one course of such propagation a \emph{cascade}. For simplicity, it is common to assume that the infection times across different edges are independent, known as the \emph{continuous-time independent cascade} (CIC) model \cite{gomez-rodriguez2016influence,du2013scalable,gomez-rodriguez2012influence}.

\begin{figure}[t]
\centering
    \begin{tikzpicture}
    \node[inner sep=0pt] (diff) at (7,2.8)
        {\includegraphics[width=.5\textwidth]{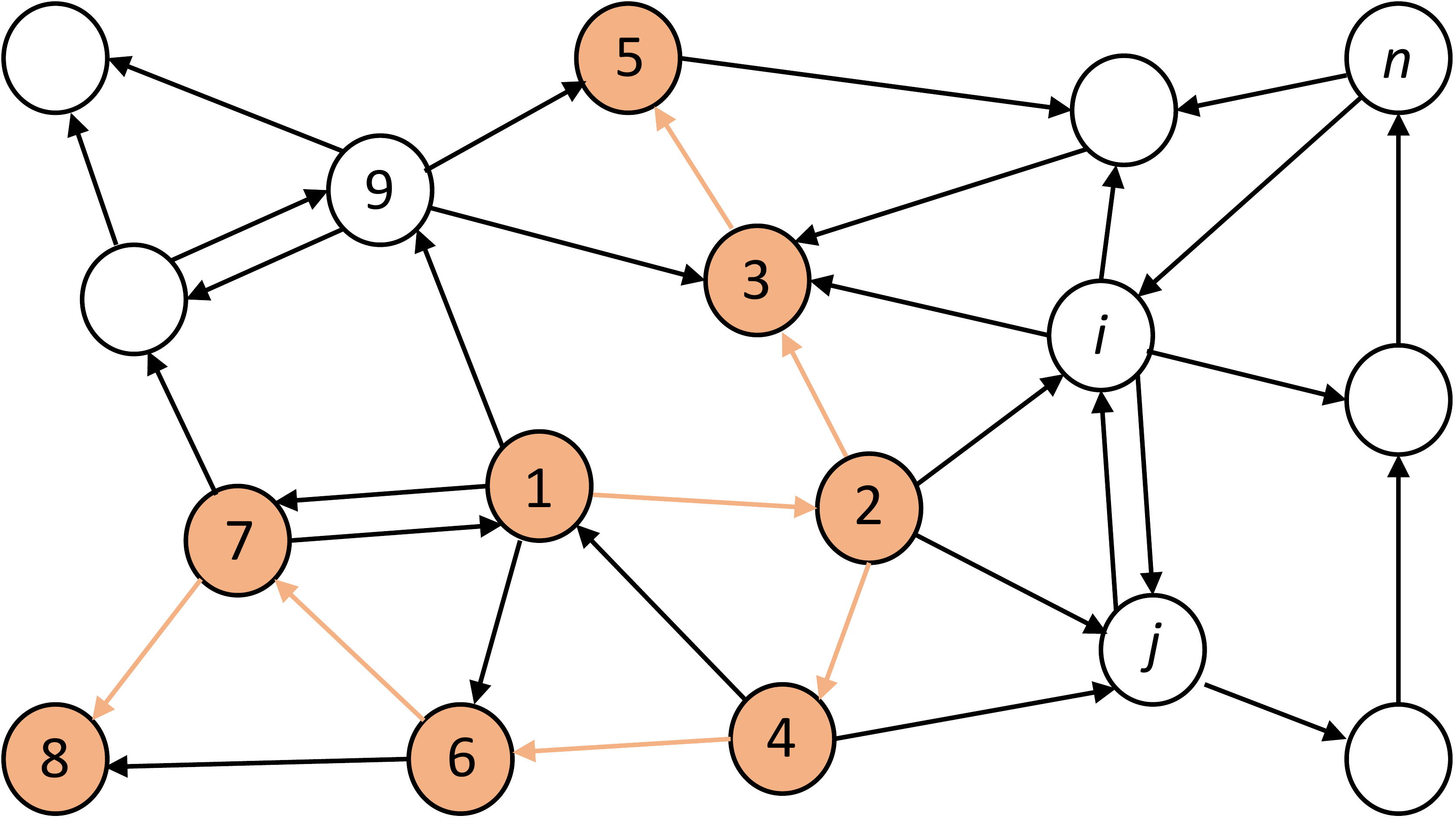}};
    \draw[-latex] (3,0) -- (11,0) ; %edit here for the axis
    \foreach \x in  {3,4,4.5,5.8,7,9,9.5,10.2} % edit here for the vertical lines
    \draw[shift={(\x,0)},color=black] (0pt,3pt) -- (0pt,-3pt);
    \node at (3,-0.5) {$t_1$};\node at (4,-0.5) {$t_2$}; \node at (4.5,-0.5) {$t_3$}; \node at (5.8,-0.5) {$t_4$};
    \node at (7,-0.5) {$t_5$};\node at (9,-0.5) {$t_6$}; \node at (9.5,-0.5) {$t_7$}; \node at (10.2,-0.5) {$t_8$};
    \end{tikzpicture}
    \caption{Example of a sample cascade on a diffusion network. The cascade was originated from the source set $\Scal = \set{1}$ and gradually propagates to other nodes through their directed edge connections. The time line below the network shows the wall-clock time $t_i$ that each node $i$ was infected during the cascade with $t_1=0$. The orange edges indicate whom each node got infection from, and $t_{ij}:=t_j-t_i$ is the time that node $i$ took to infect node $j$.}
    \label{fig:diff}
\end{figure}
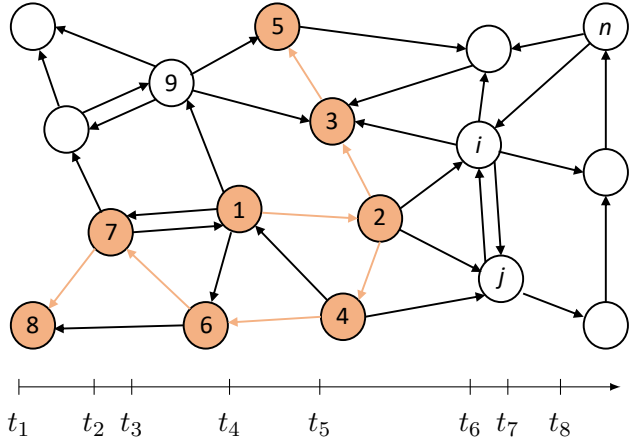
In Figure \ref{fig:diff}, we illustrate one such cascade originated from a singleton source set $\Scal = \set{1}$, which spreads to other nodes during the propagation. The orange edges indicate whom a node got infection from, for example, node 4 succeeded in infecting node 6 before node 1 did. The time line below the network indicates the wall-clock time $t_i$ of each node $i$ got infected in this cascade. In particular, $t_1 = 0$. Moreover, $t_{ij} := t_j-t_i$ is the time node $i$ took to infect node $j$. Note that this is one sample cascade of $\Scal=\set{1}$, and a different sample cascade of the same source $\Scal$ may yield different infected nodes and infection times due to the randomness of $t_{ij}$.

The standard diffusion model with exponential distribution $p(t;\alpha) = \alpha e^{-\alpha t}$ is mostly widely used in the literature. That is, $t_{ij} \sim p(t;\alpha_{ij})$ for each $(i,j) \in \Ecal$. Note that the parameter $\alpha_{ij} > 0$ in the exponential distribution indicates the \emph{strength} of impact node $i$ has on $j$---the expectation of $t_{ij} \sim p(t;\alpha_ij)$ is $1/\alpha_{ij}$---and the larger $\alpha_{ij}$ is, the sooner node $j$ will be infected by $i$ on expectation. 
We focus on the diffusion model with exponential distribution in this work. 
Other distributions, such as Rayleigh and general Weibull distributions, are also experimented in our empirical studies in this work.

\subsection{Cascade data}

Observation data $\Dcal$ of a diffusion network are often in the form of {\it sample cascades} $\Dcal:=\{\Ccal_k=(\Scal_k,\taubm_k)\in \Vcal \times \mathbb{R}_+^n:k\in[K]\}$, where the $k$th cascade $\Ccal_k$ records its source set $\Scal_k \subset \Vcal$ and the time $(\taubm_k)_i \ge 0$ which indicates when node $i$ was infected (if $i$ was not infected during $\Ccal_k$ then $(\taubm_k)_i=\infty$). See Figure \ref{fig:diff} for one of such sample cascades, where we have $\taubm=\{t_1,\dots,t_8,\infty,\dots,\infty\}$ if no other nodes were infected in this cascade. 
Cascade data are collected from historical events for training purposes.

\subsection{Network inference and influence estimation}

Suppose $\Gcal=(\Vcal,\Ecal)$ is a diffusion network with transmission matrix $\Abm$, where $(\Abm)_{ji}=\alpha_{ij}$ is the parameter of $p(t;\alpha_{ij})$ for edge $(i,j)$. Then the goal of \emph{infection probability estimation} (\emph{influence estimation}, or \emph{influence prediction}, for short) is to compute
\begin{equation}\label{eq:x}
\xbm(t;{\chibm}_{\Scal}) = [x_1(t;{\chibm}_{\Scal}),\dots,x_n(t;{\chibm}_{\Scal})]^{\top} \in [0,1]^n
\end{equation}
for all time $t>0$ and any given source set $\Scal \subset \Vcal$.
In \eqref{eq:x}, $x_i(t;{\chibm}_{\Scal})$ is the probability of node $i$ being infected at time $t$ given the source set $\Scal$, and ${\chibm}_{\Scal} \in \set{0,1}^n$ indicates the identities of $\Scal$, i.e., $({\chibm}_{\Scal})_i = 1$ if $i \in \Scal$ and $0$ otherwise. Note that we use ${\chibm}_{\Scal}$ and $\Scal$ interchangeably hereafter.
The probability $\xbm(t;{\chibm}_{\Scal})$ can also be used to compute the \emph{influence} function $\sigma(t;{\chibm}_{\Scal}):=\onebm_n^{\top} \xbm(t;{\chibm}_{\Scal})$, the expected number of infected nodes at time $t$.
Our method can be readily applied to influence functions defined with uneven weights (rather than 1's on all nodes) if the severity of infection varies at different nodes, but we focus on the even weight case for the sake of conciseness.
%
%Note that \eqref{eq:x} is intractable to compute due to the exponentially large state space of the complete dynamical system of the diffusion problem \cite{van-mieghem2009virus,gomez-rodriguez2012structure}.

Most influence estimation problems do not assume knowledge of $\Abm$. Instead, only cascade data $\Dcal$ are available. In this case, \emph{network inference} is often needed. Network inference refers to the problem of uncovering $\Ecal$ and $\Abm$ from cascade data $\Dcal$, and is of independent research interests in the literature.
Now influence estimation can be tackled by a \emph{two-stage} approach, where a network inference is performed first to learn the network structure $\Ecal$ and the diffusion model parameters $\Abm$, and then an influence estimation is used to compute the influence of the source set $\Scal$. However, both influence estimation and network inference problems are very challenging, and the approximation errors and biases in the two stages will certainly accumulate. Alternately, one can use a \emph{one-stage} approach to directly estimate $\xbm(t;{\chibm}_{\Scal})$ of any $\Scal$ from the cascade data $\Dcal$, which is more versatile and less prone to diffusion model misspecification. Our method is a such kind of one-stage method. Additionally, it allows knowledge of $\Ecal$ and/or $\Abm$, if available, to be integrated for further performance improvement.

\subsection{Influence maximization}
Given a budget size $n_0 \in \{1,\dots,n-1\}$, the goal of \emph{influence maximization} is to find the source set $\Scal$ which generates the maximal influence $\sigma(t;{\chibm}_{\Scal})$ at a prescribed time $t$ among all source sets of size $n_0$.
Influence maximization can be formulated as follows:
\begin{equation}\label{eq:im}
\max_{{\chibm}_{\Scal}}\ \sigma(t;{\chibm}_{\Scal}), \quad \mathrm{s.t.}\quad {\chibm}_{\Scal} \in \{0,1\}^n, \quad \onebm_n^{\top} {\chibm}_{\Scal} = n_0.
\end{equation}
There are two main ingredients of an influence maximization method for solving \eqref{eq:im}: an influence estimation subroutine that evaluates the influence $\sigma(t;{\chibm}_{\Scal})$ for any given source set $\Scal$, and an (approximate) combinatorial optimization solver to find the optimal set $\Scal$ of \eqref{eq:im} that repeatedly calls the subroutine. The combinatorial optimization problem is NP-hard and is often approximately solved by greedy algorithms with guaranteed sub-optimality when $\sigma(t;{\chibm}_{\Scal})$ is submodular in ${\chibm}_{\Scal}$. In this work, we propose a variational method based on the continuous relaxation of \eqref{eq:im}, and show that it can tackle this challenging problem very efficiently using our solution framework.

\subsection{Summary of contribution}
In this paper, we develop a comprehensive framework, called neural mean-field (NMF) dynamics, for simultaneous influence estimation and network inference from cascade data on a diffusion network. We substantially extend our preliminary work \cite{he2020network} which first proposed NMF for influence estimation and network influence with a discrete-time setting.  
The novelty and contribution of the present work in contrast to existing ones, including \cite{he2020network}, are summarized as follows:
\begin{enumerate}
    \item 
    We extend the NMF dynamics developed in \cite{he2020network} to the continuous-time setting which is more suitable for real-world applications of diffusion networks. We show that the continuous-time NMF dynamics can be naturally incorporated into the likelihood function of the corresponding point process, which in turn plays the role of loss function, whereas \cite{he2020network} directly uses cross-entropy of the observed discrete-time data as the loss function. 
    \item 
    We prove rigorously the connections between parameter learning in continuous-time NMF and optimal control, where the NMF parameter serves as the time invariant control. The derivations lead to a fast algorithm based on numerical ordinary differential equation (ODE) solver that is easy to implement. Unlike the standard deep residual network training used in \cite{he2020network}, we prove rigorously that the gradients in continuous-time NMF training can be efficiently computed by solving the ODE defined by NMF forward in time and an augmented co-state ODE backward in time. 
    \item 
    Based on our continuous-time NMF framework, we develop a highly efficient algorithm for the very challenging influence maximization problem. In each iteration, our algorithm only requires solving one augmented ODE based on NMF dynamics forward in time and one quadratic program, both of which can be computed very efficiently.
\end{enumerate}
All the theoretical and algorithm developments mentioned above are supported by extensive empirical studies in this work. The numerical results show that our approach is robust to the variation of the unknown underlying diffusion models, and it also significantly outperforms existing approaches on both synthetic and real-world diffusion networks.

\subsection{Paper outline}
The remainder of this paper is organized as follows. In Section \ref{sec:nmf}, we develop the proposed neural mean-field dynamics for network inference and influence estimation on diffusion networks, as well as an optimal control formulation for parameter training. We show that our new solution framework leads to an efficient influence maximization algorithm in Section \ref{sec:infmax}. We demonstrate the performance of the proposed method on influence estimation and maximization on a variety of synthetic and real-world networks in  Section \ref{sec:experiment}. A comprehensive review of related work in the literature is provided in Section \ref{sec:related}. Section \ref{sec:conclusion} concludes the paper.

\section{Neural Mean-Field Dynamics}
\label{sec:nmf}

\subsection{Modeling diffusion by stochastic jump processes}
\label{subsec:jump}
We begin with the jump process formulation of network diffusion.
Given a source set ${\chibm}_{\Scal}$, let $X_i(t;{\chibm}_{\Scal})$ denote the infection status of the node $i$ at time $t$. Namely, $X_i(t)=1$ if node $i$ is infected by time $t$, and $0$ otherwise.
Then $\{X_i(t):i\in [n]\}$ are a set of $n$ coupled jump processes, such that $X_i(t;{\chibm}_{\Scal})$ jumps from $0$ to $1$ when the node $i$ is infected at $t$.
Let $\lambda_i^*(t)$ be the conditional intensity of $X_i(t;{\chibm}_{\Scal})$ given history $\Hcal(t)=\{X_i(s;{\chibm}_{\Scal}):s\le t,\, i \in [n]\}$, i.e., 
\begin{equation}\label{eq:lambda}
\lambda_i^*(t) := \lim_{\tau \to 0^+} \frac{\ex[X_i(t+\tau;{\chibm}_{\Scal}) - X_i(t;{\chibm}_{\Scal}) | \Hcal(t)]}{\tau}.
\end{equation}
In influence estimation, our goal is to compute the probability $\xbm(t;{\chibm}_{\Scal})=[x_i(t;{\chibm}_{\Scal})]$ in \eqref{eq:x}, which is the expectation of $X_i(t;{\chibm}_{\Scal})$ conditioning on $\Hcal(t)$:
\begin{equation}\label{eq:def_x}
    x_i(t;{\chibm}_{\Scal}) = \ex_{\Hcal(t)}[X_i(t;{\chibm}_{\Scal})|\Hcal(t)].
\end{equation}
To this end, we adopt the following notations (for notation simplicity we temporarily drop ${\chibm}_{\Scal}$ in this subsection as the source set $\Scal$ is arbitrary but fixed):
\begin{equation}\label{eq:xy}
x_I(t) = \ex_{\Hcal(t)} \sbr[1]{\textstyle\prod\nolimits_{i\in I} X_i(t;{\chibm}_{\Scal}) \big\vert \Hcal(t)}, \quad y_I(t) = \textstyle\prod\nolimits_{i \in I} x_i(t), \quad e_I(t) = x_I(t) - y_I(t)
\end{equation}
for all $I \subset [n]$ and $|I| \ge 2$. Then we can derive the evolution of $\zbm := [\xbm; \ebm]$.
Here $\xbm(t)\in [0,1]^n$ is the \emph{resolved} variable whose value is of interests and samples can be observed in cascade data $\Dcal$, and $\ebm(t) = [\cdots; e_I(t); \dots] \in \mathbb{R}^{2^n-n-1}$ is the \emph{unresolved} variable.
The evolution of $\zbm$ is given by the following theorem, and the proof is provided in Section \ref{subsec:pf_z_ode}.
\begin{theorem}\label{thm:z_ode}
The evolution of $\zbm(t) = [\xbm(t); \ebm(t)]$ follows the nonlinear differential equation:
\begin{equation}\label{eq:z_ode}
\zbm' = \bar{\fbm}(\zbm),\quad \mbox{where} \quad
\bar{\fbm}(\zbm) = \bar{\fbm} (\xbm,\ebm) = 
\begin{bmatrix}
\fbm(\xbm; \Abm) - (\Abm\odot \Ebm) \onebm;\ \cdots, f_{I}(\xbm, \ebm); \cdots
\end{bmatrix},
\end{equation}
with initial value $\zbm_0 = [{\chibm}_{\Scal}; \zerobm]$, $\Ebm = [e_{ij}]\in \mathbb{R}^{n \times n}$, and 
\begin{align}
\fbm(\xbm; \Abm) &= \Abm \xbm - \diag(\xbm) \Abm \xbm, \label{eq:f}\\
f_I(\xbm,\ebm) &= \sum_{i \in I} \sum_{j \notin I} \alpha_{ji}(y_I - y_{I \cup \{j\}} + e_I - e_{I \cup \{j\}}) - \sum_{i \in I} y_{I \setminus \{i\}}\sum_{j \ne i} \alpha_{ji}(x_j - y_{ij} - e_{ij}). \label{eq:F}
\end{align}
\end{theorem}

We remark that the dimension of $\zbm$ is $2^n-1$ which grows exponentially fast in $n$ and hence renders the computation infeasible in practice.
To overcome this issue, we employ the Mori-Zwanzig formalism \cite{chorin2000optimal} to derive a reduced-order model of $\xbm$ that has dimensionality $n$ only, as shown in the next subsection.

\subsection{Mori-Zwanzig memory closure} 

We employ the Mori-Zwanzig (MZ) formalism\cite{chorin2000optimal} that allows to introduce a generalized Langevin equation (GLE) of the $\xbm$ part of the dynamics. %\eqref{eq:z_ode}. The main contribution of the MZ theory is to deduce this system of $\xbm$ of 
The GLE of $\xbm$ is derived from the original equation \eqref{eq:z_ode} describing the evolution of $\zbm = [\xbm; \ebm]$, while maintaining the effect of the unresolved part $\ebm$. This is particularly useful in our case, as we only need $\xbm$ for infection probability and influence estimation. 

Define the Liouville operator $\Lcal$ such that $\Lcal[g](\zbm) := \bar{\fbm}(\zbm) \cdot \nabla_{\zbm}g(\zbm)$ for any real-valued function $g$ of $\zbm$.
Let $e^{t\Lcal}$ be the Koopman operator associated with $\Lcal$ such that $e^{t\Lcal}g(\zbm(0)) = g(\zbm(s))$ where $\zbm(t)$ solves \eqref{eq:z_ode}.
Then $\Lcal$ is known to satisfy the semi-group property for all $g$, i.e.,
$e^{t \Lcal} g (\zbm) = g( e^{t \Lcal} \zbm)$.
Now consider the projection operator $\Pcal$ as the truncation such that $(\Pcal g)(\zbm) = (\Pcal g)(\xbm,\ebm) = g(\xbm,0)$ for any $\zbm=(\xbm,\ebm)$, and its orthogonal complement as $\Qcal = I - \Pcal$ where $I$ is the identity operator. The following theorem describes the \emph{exact} evolution of $\xbm(t)$, and the proof is given in Section \ref{subsec:pf_mz}.
\begin{theorem}\label{thm:mz}
The evolution of  $\xbm$ specified in  \eqref{eq:z_ode} can also be described by the following GLE:
\begin{equation}\label{eq:mz}
\xbm' = \fbm (\xbm; \Abm) + \int_{0}^t \kbm(t-s, \xbm(s)) \dif s,
\end{equation}
where $\fbm$ is given in \eqref{eq:f}, and $\kbm(t,\xbm) := \Pcal \Lcal e^{t \Qcal \Lcal} \Qcal \Lcal \xbm$.
\end{theorem}

Note that, \eqref{eq:mz} is \emph{not} an approximation---it is an \emph{exact} representation of the $\xbm$ part of the original problem \eqref{eq:z_ode}.
The equation \eqref{eq:mz} can be interpreted as a \emph{mean-field} equation, where the two terms on the right hand side are called the \emph{streaming term} (corresponding to the mean-field dynamics) and \emph{memory term}, respectively. The mean-field dynamics provide the \emph{main drift} of the evolution, and the memory term in a convolution form is for vital \emph{adjustment}.
This inspires us to approximate the memory term as a time convolution on $\xbm$, which naturally yields a delay differential equation reduced a continuous-time neural network, as shown in the next subsection.

\subsection{Memory approximation and delay differential equation}

To compute the evolution \eqref{eq:mz} of $\xbm$, we consider an approximation of the Mori-Zwanzig memory term by a neural network $\varepsilonbm$ with time convolution of $\xbm$ as follows,
\begin{equation}\label{eq:eps}
\int_{0}^t \kbm(t-s, \xbm(s)) \dif s \approx \varepsilonbm(\xbm(t),\hbm(t); \etabm)\quad  \mbox{where}\quad \hbm(t) = \int_{0}^t \Kbm(t-s;\wbm)  \xbm(s) \dif s .
\end{equation}
In \eqref{eq:eps}, $\Kbm(\cdot;\wbm)$ is a convolutional operator with parameter $\wbm$, and $\varepsilonbm(\xbm,\hbm; \etabm)$ is a deep neural network with $(\xbm,\hbm)$ as input and $\etabm$ as parameter. Both $\wbm$ and $\etabm$ are to be trained by the cascade data $\Dcal$.
Hence, \eqref{eq:mz} reduces to the \emph{delay differential equation} which involves a time integral $\hbm(t)$ of past $\xbm$:
\begin{equation}\label{eq:dde}
\xbm' = \tilde{\fbm}(\xbm,\hbm; \thetabm) := \fbm(\xbm;\Abm) + \varepsilonbm (\xbm, \hbm; \etabm).
\end{equation}
The initial condition of \eqref{eq:dde} with source set $\Scal$ is given by
\begin{equation}\label{eq:dde_init}
    \xbm(0) = {\chibm}_{\Scal},\quad \hbm(0) = \zerobm, \quad \mbox{and}\quad \xbm(t) = \hbm(t) = \zerobm, \quad \forall\, t<0.
\end{equation}
We call the system \eqref{eq:dde} with initial \eqref{eq:dde_init} the \emph{neural mean-field} (NMF) dynamics.

The delay differential equation \eqref{eq:dde} is equivalent to a coupled system of $(\xbm,\hbm)$ which is shown in the following theorem, whose proof is provided in Section \ref{subsec:pf_rnn}.

\begin{proposition}\label{prop:rnn}
The delay differential equation \eqref{eq:dde} is equivalent to the following coupled system of $(\xbm, \hbm)$:
\begin{subequations}\label{eq:xh}
\begin{align}
\xbm'(t) &= \tilde{\fbm}(\xbm(t),\hbm(t); \Abm, \etabm) = \fbm(\xbm(t);\Abm) + \varepsilonbm(\xbm(t),\hbm(t);\etabm) \label{eq:xh_x}\\
\hbm'(t) &= \int_{0}^{t} \Kbm(t-s;\wbm) \tilde{\fbm}(\xbm(s),\hbm(s);\Abm,\etabm) \dif s \label{eq:xh_h}
\end{align}
\end{subequations}
with initial condition \eqref{eq:dde_init}. In particular, if $\Kbm(t;\wbm)=\sum_{l=1}^L \Bbm_l e^{-\Cbm_l t}$ for some $L\in \mathbb{N}$ with $\wbm = \{(\Bbm_l, \Cbm_l)_l: \Bbm_l\Cbm_l=\Cbm_l\Bbm_l,\,\forall\, l\in[L]\}$, then \eqref{eq:xh} can be solved by a non-delay system of $(\xbm,\hbm)$ with \eqref{eq:xh_x} and $\hbm' = \sum_{l=1}^L (\Bbm_l \xbm - \Cbm_l \hbm)$.
\end{proposition}

In the remainder of this paper, we only consider the linear kernel $\Kbm(t;\wbm) = \Bbm e^{-\Cbm t}$ where $\Bbm$ and $\Cbm$ commute for simplicity. As shown in Proposition \ref{prop:rnn}, NMF \eqref{eq:dde} reduces to a non-deday ODE system of $(\xbm,\hbm)$ with \eqref{eq:xh_x} and $\hbm' = \Bbm \xbm - \Cbm \hbm$. Solving such a system for the optimal parameter $\thetabm = (\Abm, \wbm, \etabm)$ has been cast as the so-called neural ODE (NODE) in \cite{chen2018neural}. In the following subsection, we establish a direction connection between mathematical optimal control and NODE, and provide a rigorous proof that NODE exactly evaluates the gradient of the target payoff function (likelihood function in our case) during optimization from the optimal control point of view. Compared to \cite{chen2018neural}, our proof is based on calculus of variation which is more mathematically rigorous. Moreover, we show how to incorporate the running payoff (or loss) function at scattered observation times through a rigorous derivation, as needed in continuous-time NMF training.

Note that, once the optimal $\thetabm$ is obtained, we can extract $\Abm$ for network inference. Moreover, we can compute $\xbm(t)$ for all $t$ using \eqref{eq:dde} with any given source set $\Scal$, which solves the influence estimation problem. Therefore, the network inference and influence estimation problems can be tackled simultaneously by the parameter training of NMF.

\subsection{Parameter training and optimal control}
To obtain explicit form of NMF \eqref{eq:dde} for influence estimation and network inference, we need to know the network parameters $\thetabm = (\Abm, \etabm, \wbm)$.
Let $[0,T]$ be the time horizon of the cascade data $\Dcal = \{\Ccal_k = (\Scal_k, \taubm_k): k \in [K]\}$, i.e., all cascades are recorded up to time $T$.  
Given any particular $\Ccal = (\Scal, \taubm) \in \Dcal$ where ${\taubm}=\{t_i \in [0,T] \cup \{\infty\}: i \in [n]\}$, it suffices to derive the negative log-likelihood function of the infection probabilities $\xbm$ given \eqref{eq:dde} with parameter $\thetabm$ for this cascade $\Ccal$. The total negative log-likelihood function is thus the sum of such function over all the $K$ cascades in $\Dcal$.

Recall from Section \ref{subsec:jump} that $\Xbm(t)$ is the jump stochastic process describing the infection state of the nodes and $\xbm(t)$ is the infection probabilities. Therefore, $\xbm'(t)$ is essentially the (non-conditional) intensity of $\Xbm(t)$. In other words, $\Xbm(t)$ is identical to a non-homogeneous Poisson process with intensity function $\xbm'(t)$ for $t$ almost everywhere. Due to the relation between the intensity function and the likelihood function of a point process \cite{gulddahl-rasmussen2018lecture}, the negative log-likelihood function of $\xbm'(t)$ given the cascade $\Ccal = (\Scal, \taubm)$ can be easily obtained, and it is also the ``loss function'' $\ell$ we need to minimize:
\begin{equation}
    \label{eq:loss}
    \ell(\xbm; \Ccal)= \sum_{i=1}^{n} \del[2]{-\log x_i'(t_i) + {x}_i(T)} = \int_{0}^T r(\xbm(t),\thetabm)\dif t+ \onebm^{\top}\xbm(T),
\end{equation}
where the running loss function is defined by
\begin{equation}
\label{eq:runloss}
    r(\xbm(t),\thetabm)=\sum_{i=1}^{n}-\delta(t-t_i) \log x_i'(t) = \sum_{i=1}^{n}-\delta(t-t_i) \log (\tilde{\fbm}(\xbm, \hbm; \thetabm))_i,
\end{equation}
and $\delta(\cdot)$ is the Dirac delta function.
The running loss takes into account the changes of $\xbm(t)$ at intermediate times during $[0,T]$.

We can further add regularization or incorporate prior information to \eqref{eq:loss}. 
In particular, if $\Ecal$ is given, we know that $\Abm$ must be supported on $\Ecal$, which serves as the constraint of $\Abm$.
If we know the network has low density (sparse edges), then we can enforce a sparsity regularization such as $\|\Abm\|_1$ (the $l_1$ norm of the vectorized $\Abm \in \mathbb{R}^{n^2}$). 
In general, $\Abm$ can be interpreted as the convolution to be learned from a graph convolution network (GCN)\cite{kipf2017semi-supervised,wu2020comprehensive}. The support and magnitude of $\Abm$ implies the network structure and strength of interaction between pairs of nodes, respectively. We will provide more details of our choice of regularization and its parameter setting in Section \ref{sec:experiment}.

To summarize, the optimal parameter $\thetabm$ of \eqref{eq:dde} can be obtained by minimizing the loss function in \eqref{eq:loss} for the given cascade $\Ccal$:
% %
\begin{subequations}\label{eq:oc}
\begin{align}
\min_{\thetabm} \quad & \ell(\thetabm;\Ccal) :=\int_0^T r(\xbm(t), \thetabm) \dif t+ \onebm^{\top}\xbm(T), \label{eq:oc_obj}\\ 
\mathrm{s.t.}\quad & \mbm'(t) = \gbm(\mbm(t);\thetabm), \quad \mbm(0) = [ {\chibm}_{\Scal_{k}};\zerobm],\quad t\in [0,T], \label{eq:oc_m}
\end{align}
\end{subequations}
where $\mbm(t) = [\xbm(t); \hbm(t)] \in \mathbb{R}^{2n}$ and 
\begin{equation}
    \label{eq:g}
    \gbm(\mbm; \thetabm) = \begin{pmatrix} \Abm \xbm - \diag(\xbm) \Abm \xbm + \varepsilonbm(\xbm, \hbm; \etabm) \\ \Bbm \xbm - \Cbm \hbm \end{pmatrix}.
\end{equation}
This is the parameter training problem given one cascade $\Ccal$, and can be trivially extended to the case $\Dcal$ which consists of $K$ cascades. In what follows, we drop the notation $\Ccal$ for conciseness.

In \eqref{eq:g}, $\gbm(\mbm;\thetabm)$ is the NMF dynamics derived in \eqref{eq:xh} with parameter $\thetabm = (\Abm, \wbm, \etabm)$ and $\wbm = (\Bbm, \Cbm)$. In particular, $\etabm$ stands for the network parameters of the deep neural network $\varepsilonbm$ that wraps the augmented state $\mbm$ to approximate the MZ memory term \eqref{eq:mz}.
It is worth noting that, the so-called control variable $\thetabm$ is constant and time-invariant in NODE \cite{chen2018neural} as well as in NMF. Therefore, it is considerably easier to handle than that in classical optimal control $\thetabm$. Specifically, we can develop an algorithm for solving $\thetabm$ in \eqref{eq:oc} which is easy to implement. Moreover, we can derive rigorous proof of the relation between the gradient of the loss function and the solution of the augmented dynamics.

As we can see, to find the optimal $\thetabm$ of \eqref{eq:oc}, the key is to compute $\nabla_{\thetabm} \ell (\thetabm)$ for any $\thetabm$. To this end, we recall that the \emph{Hamiltonian} function associated with the control problem \eqref{eq:oc} is
\begin{equation}
    \label{eq:hamiltonian}
    H(\mbm(t), \pbm(t); \thetabm) = \pbm(t) \cdot \gbm(\mbm(t); \thetabm) + r(\mbm(t),\thetabm),
\end{equation}
where $\pbm(t) \in \mathbb{R}^{2n}$ is the co-state variable (also known as the adjoint variable) associated with $\mbm(t)$. Here, $\pbm(t)$ plays the role of Lagrange multiplier for the ODE constraint \eqref{eq:oc_m}. The standard optimal control theory states that the co-state $\pbm(t)$ follows the ODE backward in time as follows:
\begin{equation}
\label{eq:ode-p}
\begin{cases}
\pbm'(t) = -\nabla_{\mbm} \gbm(\mbm(t); \thetabm) \pbm(t) - \nabla_{\mbm} r(\mbm(t),\thetabm), & \quad T \ge t \ge 0, \\
\pbm(T) = [\onebm; \zerobm].
\end{cases}
\end{equation}
The terminal condition $\pbm(T) = [\onebm; \zerobm]$ has this simple form because the ``terminal loss'' in \eqref{eq:oc} is given by $[\onebm; \zerobm] \cdot \mbm(T) = \onebm \cdot \xbm(T)$.

Now we show that $\nabla_{\thetabm} \ell$ can be obtained by solving the ODE \eqref{eq:oc_m} forward in time and an augmented ODE backward in time. To this end, we need the following theorem, whose proof is given in Appendix \ref{subsec:pf_pmp}.
\begin{theorem}\label{thm:pmp}
The gradient $\nabla_{\thetabm} \ell(\thetabm)$ of the loss function $\ell$ defined in \eqref{eq:oc} for any parameter $\thetabm$ and cascade data $\Ccal$ is given by 
\begin{equation}
\label{eq:grad-P}
\nabla_{\thetabm}\ell(\thetabm)=\int_{0}^{T}\del[2]{\nabla_{\thetabm} \gbm(\mbm(t);\thetabm)\pbm(t) + \nabla_{\thetabm} r (\mbm(t),\thetabm)} \dif t .
\end{equation}
Moreover, if $\mbm^*$ is the solution of \eqref{eq:oc_m} using the optimal solution $\thetabm^*$ to \eqref{eq:oc}, and $\pbm^*$ is the co-state determined by \eqref{eq:ode-p} with $\mbm^*$ and $\thetabm^*$, then $\int_0^{T} \nabla_{\thetabm} H(\mbm^*(t),\pbm^*(t);\thetabm) \dif t = \zerobm$.
\end{theorem}

The formula \eqref{eq:grad-P} in Theorem \ref{thm:pmp} suggests that we can compute $\nabla_{\thetabm} \ell$ by tracking an auxiliary variable $\qbm$ that follows the backward differential equation and terminal condition:
\begin{equation}
\label{eq:ode-ptheta}
\begin{cases}
\qbm'(t) = -\nabla_{\thetabm} \gbm(\mbm(t),\thetabm)^{\top} \pbm(t) - \nabla_{\thetabm} r(\thetabm,\mbm(t)), & \quad T \ge t \ge 0, \\
\qbm(T) = \zerobm.
\end{cases}
\end{equation}
Then  \eqref{eq:grad-P} implies that $\nabla_{\thetabm} \ell(\thetabm) = \qbm(T) - \int_{0}^{T} \qbm'(t) \dif t = \qbm(0)$. 

Before closing this section, we need to clarify one implementation issue with the running loss $r$. Suppose that the infection times in the cascade $\Ccal$ can be sorted $0 < t^{(1)} < t^{(2)} < \cdots < t^{(m)} < t^{(m+1)} := T$. That is, there are $m$ infections (excluding the infections at the source nodes) during the cascade $\Ccal$. (Note that any two infection times coincide with probability 0 since the point process is simple.) For notation simplicity, suppose that at time $t^{(i)}$, the new infected node is $i$.

Then the integral of the running loss reduces to
\begin{align}\label{eq:rl}
      \int_{0}^T \nabla_{\thetabm}r(\thetabm,\mbm)\dif t
    &= \sum_{i=0}^{m} \nabla_{\thetabm} \left(-\log  \gbm_i(\mbm(t^{(i)});\thetabm)\right),
\end{align}
where $\gbm_i(\mbm(t),\thetabm)$ is the $i$th component of $\gbm(\mbm(t),\thetabm)$. 
Hence, we need to compute $\qbm(0)$ by solving the ODE \eqref{eq:ode-ptheta} backward in each time interval as
\begin{align}\label{eq:back_theta}
\qbm(t^{(i-1)}) &= \qbm(t^{(i)}) - \int_{t^{(i)}}^{t^{(i-1)}} \nabla_{\thetabm} \gbm(\mbm(t),\thetabm)^{\top} \pbm(t)\dif t-\nabla_{\thetabm} \log \gbm_{i-1}(\mbm(t^{(i-1)});\thetabm).
\end{align}
Similarly, we have
\begin{align}\label{eq:back_p}
 \pbm(t^{(i-1)}) &=
 \pbm(t^{(i)}) - \int_{t^{(i)}}^{t^{(i-1)}}\nabla_{\mbm}\gbm(\mbm(t),\thetabm)^{\top} \pbm(t)\dif t- \nabla_{\mbm} \log \gbm_{i-1}(\mbm(t^{(i-1)});\thetabm).
\end{align}
The ODE of $\mbm(t)$ remains the same as in \eqref{eq:oc_m} since it does not involve the running loss $r$.

To summarize, in order to compute $\nabla_{\thetabm} \ell(\thetabm)$ for any given $\thetabm$, we need to first solve the ODE \eqref{eq:oc_m} of $\mbm(t)$ forward in time from $0$ to $T$; Then we need to solve the ODE system \eqref{eq:oc_m}, \eqref{eq:ode-p}, and
\eqref{eq:ode-ptheta} of $(\mbm(t), \pbm(t), \qbm(t))$ backward in time from $T$ to $0$. In particular, we need to solve the backward ODE such that the last term \eqref{eq:back_p} and \eqref{eq:back_theta} are added for $\pbm(t)$ and $\qbm(t)$ in each time interval $(t^{(i-1)},t^{(i)}]$. Finally, we obtain $\nabla_{\thetabm} \ell(\thetabm) = \qbm(0)$.
The complete training process is summarized in Algorithm \ref{alg:nmf}, where mini-batches of cascades are used to compute the stochastic gradient in searching the (local) minimizer $\thetabm$.
We did not include the gradient of the regularization of $\thetabm$, but its computation is standard and can be easily added to $\nabla_{\thetabm} \ell(\thetabm)$.

\begin{algorithm}[t]
\caption{Neural mean-field (NMF) dynamics}
\label{alg:nmf}
\begin{algorithmic}[1]
\STATE \textbf{Input:} $\Dcal = \{\Ccal_k = (\Scal_k,\taubm_k): k \in [K]\}$.
\STATE \textbf{Initialization:} Network architecture $\gbm(\cdot; \thetabm)$ and parameter  $\thetabm=(\Abm, \etabm, \wbm)$.
\FOR{$k=1,\dots,\text{MaxIterations}$}
\STATE Sample a mini-batch of cascades $\hat{\Dcal} \subset \Dcal $.
\STATE Compute $\mbm(t)$ in \eqref{eq:oc_m} forward in time for each $\Ccal \in \hat{\Dcal}$. \hfill (Forward pass)
\STATE Compute $\sum_{\Ccal \in \hat{\Dcal}} \nabla_{\thetabm} \ell(\thetabm; \Ccal)$ using the \textbf{BackwardMode} below. \hfill (Backward pass)
\STATE Update parameter $\thetabm$ using ADAM with stochastic gradient $\sum_{\Ccal \in \hat{\Dcal}} \nabla_{\thetabm} \ell(\thetabm; \Ccal)$.
\ENDFOR
\STATE \textbf{Output:} Network parameter $\thetabm$.
\algrule
\textbf{BackwardMode}
\STATE \textbf{Input:} Cascade $\Ccal = (\Scal,\taubm)$ with $\taubm: 0 =t^{(0)} < t^{(1)} < \cdots < t^{(m+1)} = T$ and $\mbm(T)$. 
\STATE \textbf{Terminal augmented state:} $[\mbm(T); \pbm(T); \qbm(T)] = [\mbm(T); [\onebm;\zerobm]; \zerobm]$.
\FOR{$i=m+1,\dots,1$}
\STATE Solve the ODE below backward in time $(t^{(i-1)}, t^{(i)}]$:
\[
\begin{pmatrix}
\mbm'(t) \\ \pbm'(t) \\ \qbm'(t)
\end{pmatrix} =
\begin{pmatrix}
    \gbm(\mbm(t);\thetabm) \\
    -\nabla_{\mbm}\gbm(\mbm(t);\thetabm)\pbm(t) \\
    -\nabla_{\thetabm}\gbm(\mbm(t);\thetabm)\pbm(t)
\end{pmatrix}
\]
with terminal condition $[\mbm(t^{(i)}); \pbm(t^{(i)}); \qbm(t^{(i)})]$.
\STATE $\pbm(t^{(i-1)}) \leftarrow \pbm(t^{(i-1)}) - \nabla_{\mbm} \log  \gbm_{i-1}(\mbm(t^{(i-1)});\thetabm)$.
\STATE $\qbm(t^{(i-1)}) \leftarrow \qbm(t^{(i-1)}) - \nabla_{\thetabm} \log  \gbm_{i-1}(\mbm(t^{(i-1)});\thetabm)$.
\ENDFOR
\STATE \textbf{Output:} $\nabla_{\thetabm}\ell(\thetabm;\Ccal) \leftarrow \qbm(0)$.
\end{algorithmic}
\end{algorithm}

\section{Influence Maximization with Learned NMF}
\label{sec:infmax}
In this section, we show how the proposed NMF can be used to tackle an important but very challenging problem known as the influence maximization.
Suppose we have trained an NMF with parameters $\thetabm$ in Algorithm \ref{alg:nmf}, such that we can estimate $\xbm(t)$ for any $t\in[0,T]$ and any given source node set ${\chibm}_{\Scal}$.
Then the goal of influence maximization is to identify ${\chibm}_{\Scal} \in \{0,1\}^n$ such that its influence at the prescribed time $T$ (or any other prescribed $t \in (0,T)$) is maximized.
Namely, our goal is to solve the following optimization problem
\begin{equation}
    \label{eq:im_control}
    \max_{{\chibm}_{\Scal}}\ \sigma(T;{\chibm}_{\Scal}) := \onebm_n^{\top}\xbm(T;{\chibm}_{\Scal}), \quad \mbox{s.t.}\quad  {\chibm}_{\Scal} \in \{0,1\}^n,\quad \onebm_n^{\top} {\chibm}_{\Scal} = n_0,
\end{equation}
where $n_0 \in \mathbb{N}$ is the given budget size. 
Note that $\xbm(T;{\chibm}_{\Scal})$ is the first $n$ components of $\mbm(T)$ computed by forward NMF dynamics with initial value $\mbm(0) = [{\chibm}_{\Scal};\zerobm]$.
However, \eqref{eq:im_control} is an NP-hard combinatorial optimization problem \cite{gomez-rodriguez2012structure}, we propose to relax the binary-valued decision vector ${\chibm}_{\Scal}$ to $\ubm \in [0,1]^n$ in the continuous hypercube $[0,1]^n$ as
\begin{equation}
    \label{eq:im_relax}
    \min_{\ubm \in \Ucal}\ L(\ubm) := \mathcal{R}(\ubm) - \onebm_n^{\top}\xbm(T;\ubm),  \quad \mbox{where}\quad \Ucal := \{ \ubm \in [0,1]^n :  \onebm_n^{\top} \ubm = n_0 \},
\end{equation}
and $\mathcal{R}(\ubm)$ is a regularizer that encourages all components of $\ubm$ to take values close to either $0$ or $1$. In our experiments, we simply set $\mathcal{R}(\ubm) = \sum_{i=1}^{n} u_i (1-u_i)$.
Then we employ the projected gradient descent (PGD) method to solve \eqref{eq:im_relax}:
\begin{equation}
    \label{eq:im_qp}
    \ubm_{l+1} = \Pi_{\Ucal}(\ubm_{l} - \gamma_l \nabla_{\ubm} L(\ubm_{l})) := \argmin_{\ubm \in \Ucal}\,\|\ubm - (\ubm_l -\gamma_l \nabla_{\ubm} L(\ubm_{l}))\|^2,
\end{equation}  
where $l$ is the iteration counter of PGD, $\tau_l>0$ is the step size, and $\Pi_{\Ucal}$ denotes the orthogonal projection onto $\Ucal$. If $\nabla_{\ubm} L(\ubm_{l})$ is known, then \eqref{eq:im_qp} is a standard quadratic program (QP) and can be solved efficiently by off-the-shelf solvers.
Therefore, the only remaining question is to compute $\nabla_{\ubm} L(\ubm)$ for any given $\ubm$.
The following theorem states that this quantity can be computed very efficiently using the proposed NMF dynamics. The proof is provided in Appendix \ref{subsec:pf_infmax-grad}.
\begin{theorem}
\label{thm:infmax-grad}
Let $[\mbm(t); \sbm(t)]$ be the solution of the augmented NMF system:
\begin{equation}
\label{eq:infmax-NMF}
    \begin{pmatrix}
        \mbm'(t) \\ \sbm'(t)
    \end{pmatrix}
    =
    \begin{pmatrix}
        \gbm(\mbm(t);\thetabm) \\ \nabla_{\xbm} \gbm_{\xbm}(\mbm;\thetabm)^{\top} \sbm(t)
    \end{pmatrix}
\end{equation}
with initial value $[\mbm(0);\sbm(0)] = [[\ubm; \zerobm]; \onebm]$ forward in time $[0,T]$, where $\gbm_{\xbm}$ is the first $n$ components of $\gbm$. Then $\nabla_{\ubm} L(\ubm) = \nabla_{\ubm}\Rcal(\ubm) - \sbm(T)$.
\end{theorem}

Theorem \ref{thm:infmax-grad} implies that $\nabla_{\ubm} L(\ubm)$ can be easily computed by solving NMF augmented by an auxiliary variable $\sbm(t)$ forward in time $[0,T]$ as in \eqref{eq:infmax-NMF}. Note that the computation complexity  of \eqref{eq:infmax-NMF} is linear in the network size $n$ and standard numerical ODE integrators can quickly solve the ODE to high accuracy. We summarize the steps for solving \eqref{eq:im_relax} in  Algorithm \ref{alg:nmf_infmax}. Note that the output $\ubm$ may not be binary, and thus we can set the largest $n_0$ components of $\ubm$ to $1$ and the rest to $0$ as the final source set selection.

\begin{algorithm}[t]
\caption{Influence maximization via neural mean-field dynamics (NMF-InfMax)}
\label{alg:nmf_infmax}
\begin{algorithmic}[1]
\STATE \textbf{Input:} Trained NMF with $\gbm(\cdot;\thetabm)$ from Algorithm \ref{alg:nmf}, budget $n_0\in \{1,\dots,n-1\}$
\STATE \textbf{Initialization:} $\ubm\in \Ucal$.
\FOR{$l=1,\dots,\text{MaxIterations}$}
\STATE Solve $[\mbm(T),\sbm(T)]$ from \eqref{eq:infmax-NMF} forward in time with initial $[[\ubm;\zerobm];\onebm]$.  
\hfill (Forward pass)
\STATE Set $\hat{\ubm} \leftarrow \ubm - \gamma \nabla_{\ubm} L(\ubm)$ where $\nabla_{\ubm} L(\ubm) = \sbm(T)$.
\STATE Solve a QP: $\ubm \leftarrow \argmin_{\ubm \in \Ucal}\,\|\ubm - \hat{\ubm}\|^2$.
\ENDFOR
\STATE 
\textbf{Output:} Source set selection $\ubm$.
\end{algorithmic}
\end{algorithm}

\section{Numerical Experiments}
\label{sec:experiment}

\subsection{Implementation details} 

In our NMF implementation, the neural mean field dynamic  $\gbm(\cdot;\thetabm)$ derived from Proposition \ref{prop:rnn} is learned as Algorithm \ref{alg:nmf}, where $\varepsilonbm(\xbm,\hbm;\etabm)$ is a three-layer fully connected network. Specifically, the input layer size of $\varepsilonbm$ is $2n$, and both of the hidden and output layer sizes are $n$. We use Exponential Linear Unit (ELU) as the activation function. The output is truncated into $[0,1]$. We use the $\ell_0$-norm regularization approximated by log-sum introduced by \cite{qiao2020log-sum}. 
The NMF networks are trained and tested in PyTorch \cite{paszke2019pytorch} by Adam optimizer \cite{kingma2014adam} with default parameters (lr=0.001, $\beta_1$=0.9, $\beta_2$=0.999, $\epsilon$=1e-8) on a Linux workstation with Intel i9 8-Core Turbo 5GHz CPU, 64GB of memory, and an Nvidia RTX 2080Ti GPU. InfluLearner and N\textsc{et}R\textsc{ate} are both trained by the Matlab code published by the original authors. 
All experiments are performed on the same machine. 
Given ground truth node infection probability $\xbm^*$, the Mean Absolute Error (MAE) of influence (Inf) and infection probability (Prob) of the estimated $\xbm$ are defined by $|\onebm \cdot (\xbm(t) -\xbm(t)^*)|$ and $\|\xbm(t)-\xbm(t)^*\|_1/n$ for every $t$, respectively. 
For all the experiments related on the influence estimation, we also use the scaled influence MAE $|\onebm \cdot (\xbm(t) -\xbm(t)^*)|/n$ as an evaluation metric.

%Due to space limitation in the main text, we present the numerical experiments on influence estimation problem, including parameter setting, compared methods, and numerical results in Section \ref{subsec:exp1}.

\subsection{Infection probability and influence function estimation}\label{subsec:exp1}

We first apply the proposed NMF to synthetic diffusion networks where ground truth node infection probabilities are available for quantitative evaluation.

\paragraph{Networks} We use three types of network models \cite{leskovec2010kronecker} to generate these synthetic networks: hierarchical (Hier) network \cite{clauset2008hierarchical}, core-periphery (Core) network \cite{leskovec2008statistical} and Random (Rand) network with parameter matrices [0.9,0.1;0.1,0.9], [0.9,0.5;0.5,0.3], and [0.5,0.5;0.5,0.5], respectively. 

For each of these three types of networks, we randomly generate 5 networks of $(n,d)=(128,4)$ and another 5 networks of $(n,d)=(1024,4)$, where $n$ is the total number of nodes on the network and $d$ is the average out-degree per node.

\paragraph{Diffusion models and parameters} 
We simulate the diffusion on these networks such that the infection time are modeled by exponential distribution (Exp), Rayleigh distribution (Ray), and general Weibull distribution (Wbl).

Note that our theoretical results in this work are based on diffusion models using exponential distribution, however, we still conduct experiments on other distributions to test the performance of NMF empirically. In particular, we draw the parameters $\alpha_{ji}$ from Unif[0.1,1] to simulate the heterogeneous interactions between nodes for exponential and Reyleigh distributions. We generate both of the shape and scale parameters of Weibull distribution from Unif[1,10] randomly.

\paragraph{Training and testing data}
We randomly generate 900 source node sets of size varying between 1 and 10, and simulate 10 diffusion cascades for each source set for training. Thus the training data consists of $K$=9,000 cascades, all of which are truncated into time window $[0,T]$ with $T=20$.

We generate 100 additional source sets in a similar way, and then split them as 50\%-validation and 50\%-test with ground truth of infection probability and influence estimated by simulating 10,000 cascades for each source set. This setting on validation and test data will be used for all the experiments related to influence estimation and all networks and cascades are generated using the SNAP package \cite{leskovec2016snap}. 

\paragraph{Algorithm and parameter settings}
In the training of NMF, the batch size of cascade data is set to 300 and the number of epochs is 50. The coefficients of the regularization term on $\Abm$ and weight decay in Adam optimizer are set to (0.01,1) and (0.001,0) for network of size 128 and 1024, respectively. We use Runge-Kutta 4th order (rk4) method with 40 time steps to solve the ODEs numerically.

\paragraph{Comparison algorithm}
For comparison, we use InfluLearner \cite{du2014influence}, which is a state-of-the-art method that can estimate individual node infection probability directly from cascade data in the CIC setting as our method.

InfluLearner draws a set of random binary features from certain distribution for each node $j$ indicating the reachabilities of $j$ by other nodes, and then uses a convex combination of random basis function to parameterize the conditional infection probability of the node given a source set over these binary vectors. To estimate the reachability distribution, InfluLearner calculates the mean frequency of node $j$ being influenced by a source node $s$, average over all cascades in the training dataset with the source $s$.

In our test, we set the number of random features to 200 as suggested in \cite{du2014influence}.

It is worth noting that InfluLearner requires additionally the source identity for each infection to estimate the coverage functions. That is, InfluLearner also needs to know the original source node in the source set for each and every new infection occurred in the cascade in the training data. This additional information is provided in our simulated data in favor of InfluLearner. However, it is often unavailable in real-world applications such as epidemic spreads. The proposed NMF method does not have such restriction.

Moreover, to quantify estimation error, we compute the MAE of node infection probability and influence at $t_l=l$ for $l=1,\dots,20$, and average each over the 50 test source sets. Since InfluLearner needs to learn the coverage function for a prescribed time $t$, we have to run it for each of the 20 time points one by one. In contrast, the proposed NMF is more advantageous since it can directly estimate the entire evolution of infection probabilities during $[0,T]$, which is more computationally efficient. 

\paragraph{Comparison results}

\begin{figure}
\centering

\begin{subfigure}[b]{.24\textwidth}
\includegraphics[width=\textwidth]{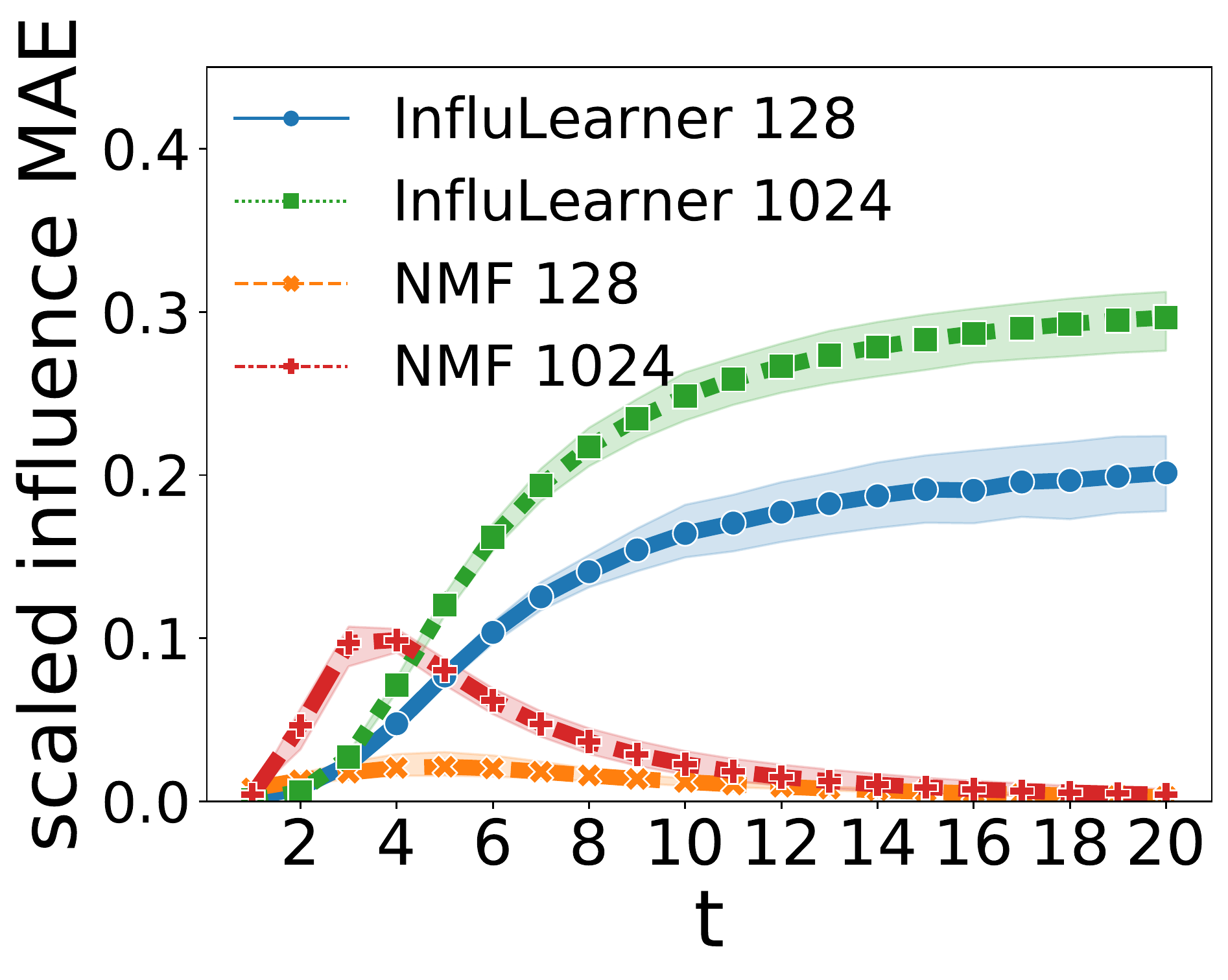}
\end{subfigure}
\begin{subfigure}[b]{.24\textwidth}
\includegraphics[width=\textwidth]{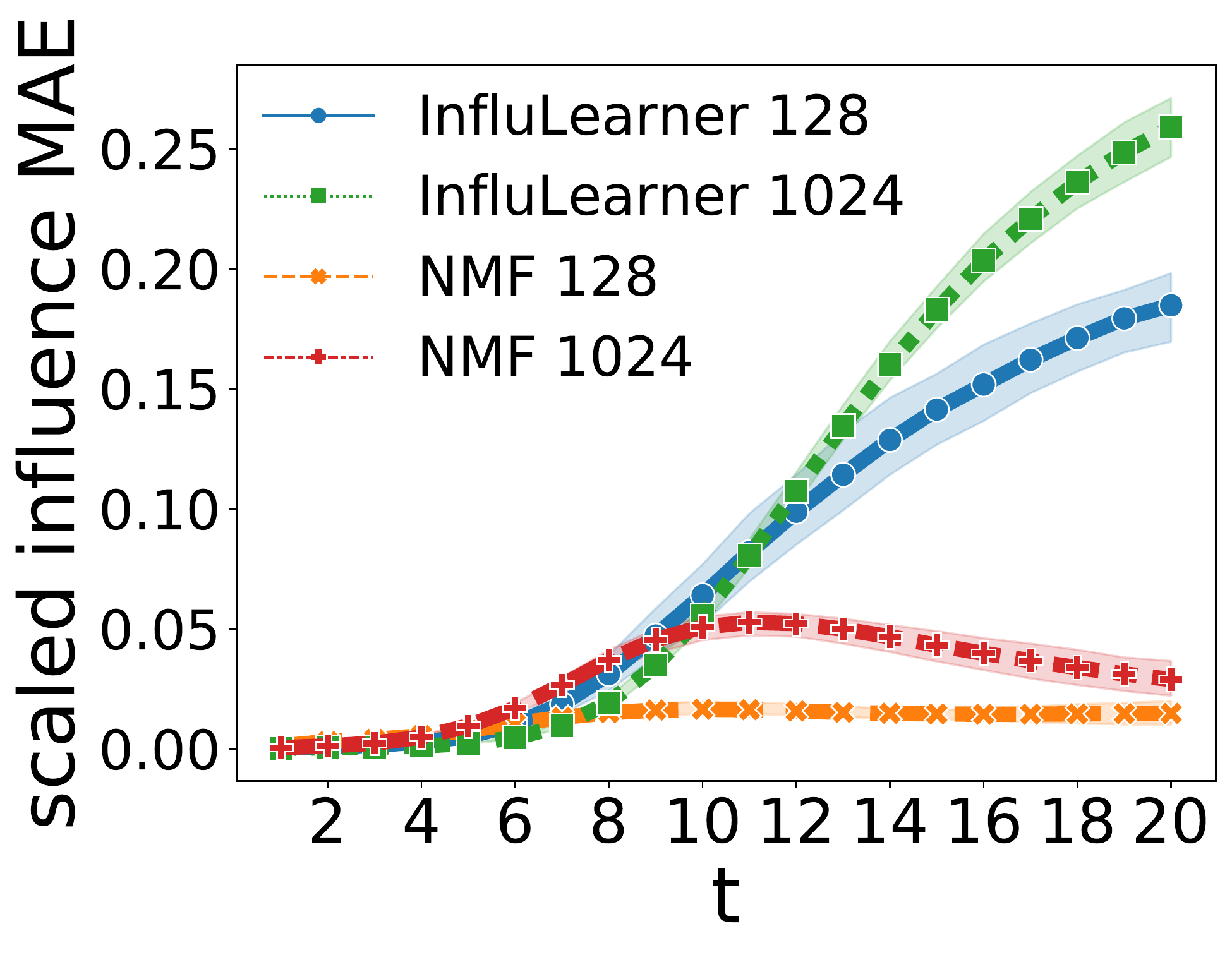}
\end{subfigure}
\begin{subfigure}[b]{.24\textwidth}
\includegraphics[width=\textwidth]{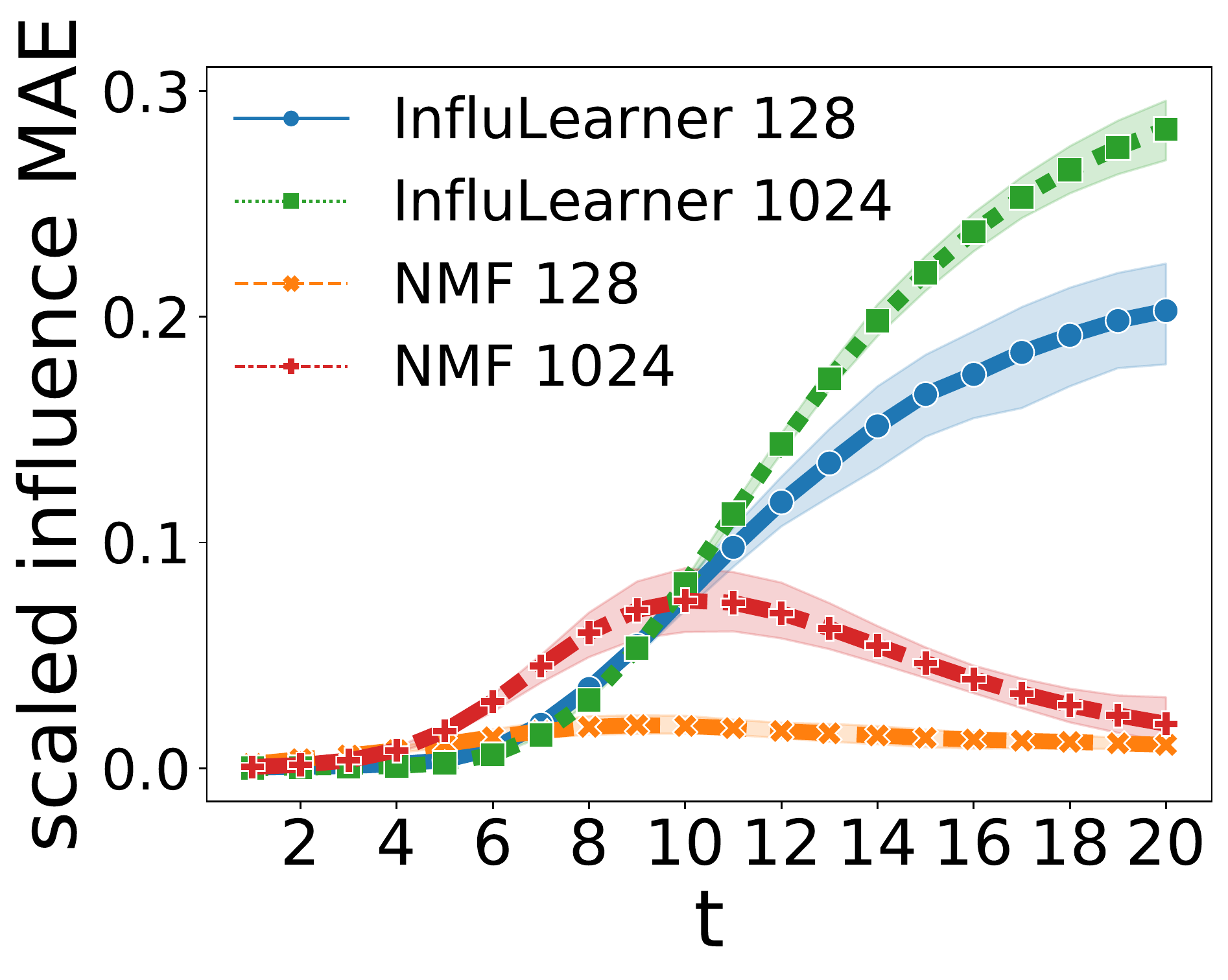}
\end{subfigure}\\
\begin{subfigure}[b]{.24\textwidth}
\includegraphics[width=\textwidth]{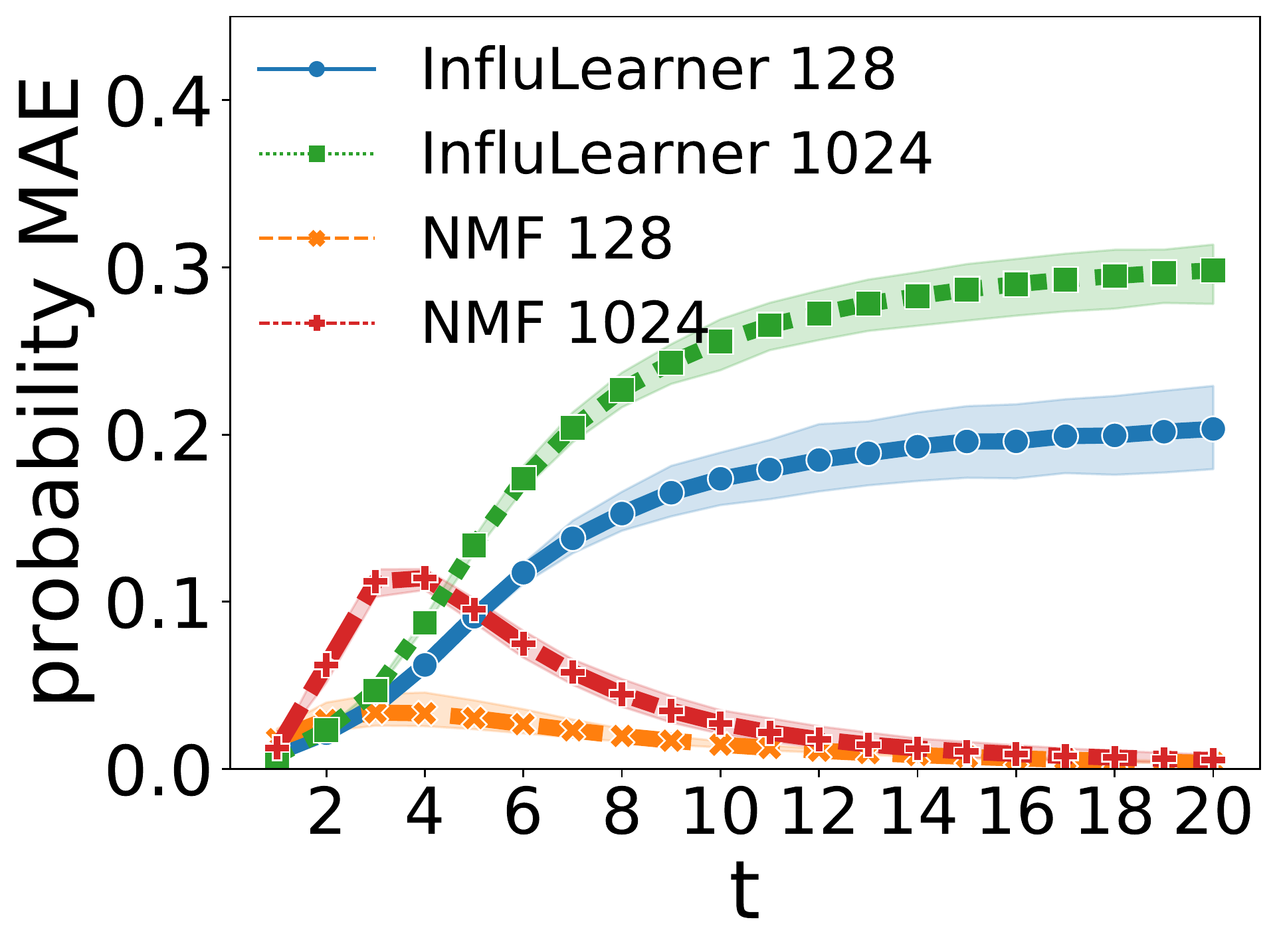}
\caption{Core + Exp}
\end{subfigure}
\begin{subfigure}[b]{.24\textwidth}
\includegraphics[width=\textwidth]{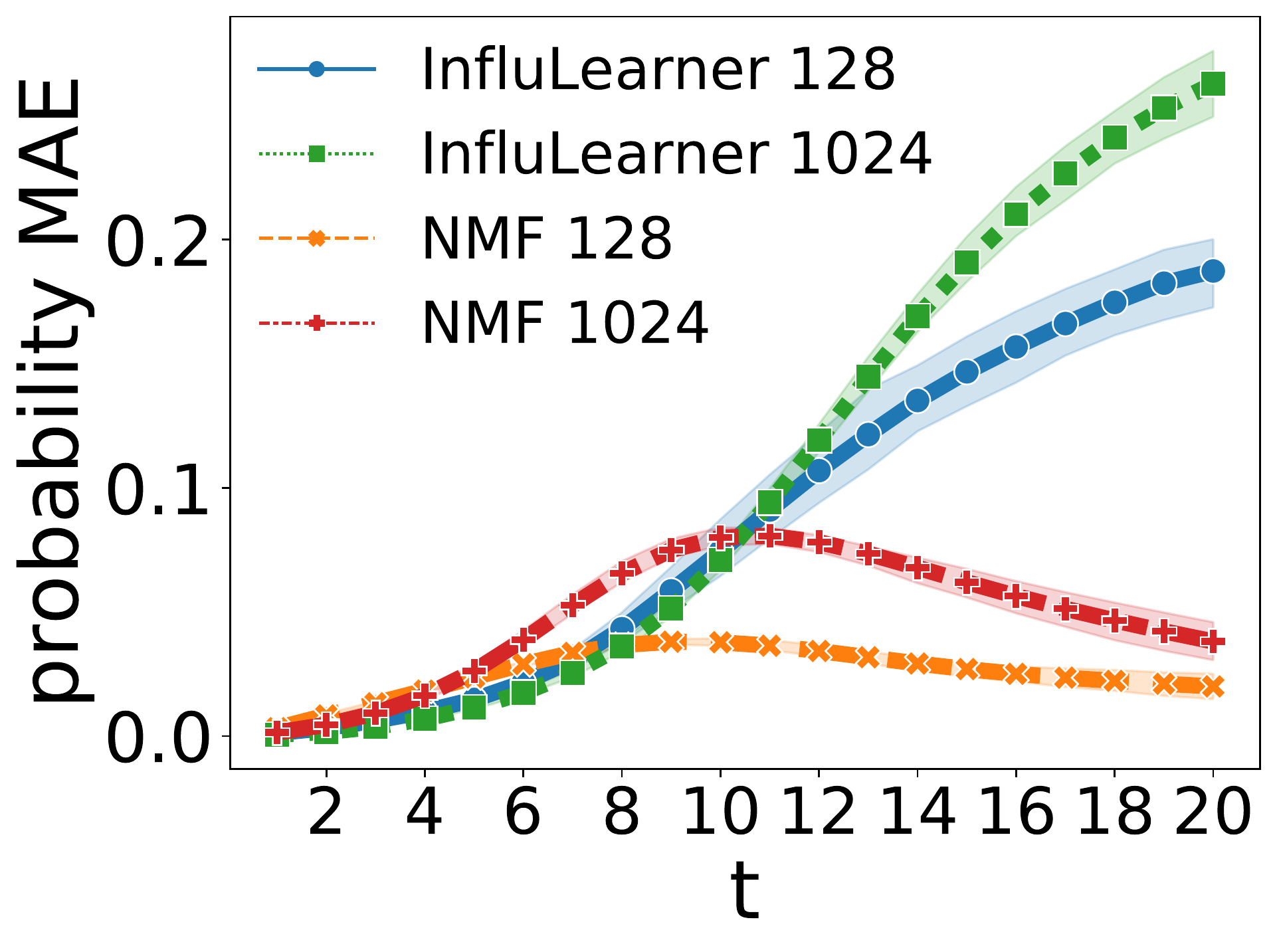}
\caption{Core + Ray}
\end{subfigure}
\begin{subfigure}[b]{.24\textwidth}
\includegraphics[width=\textwidth]{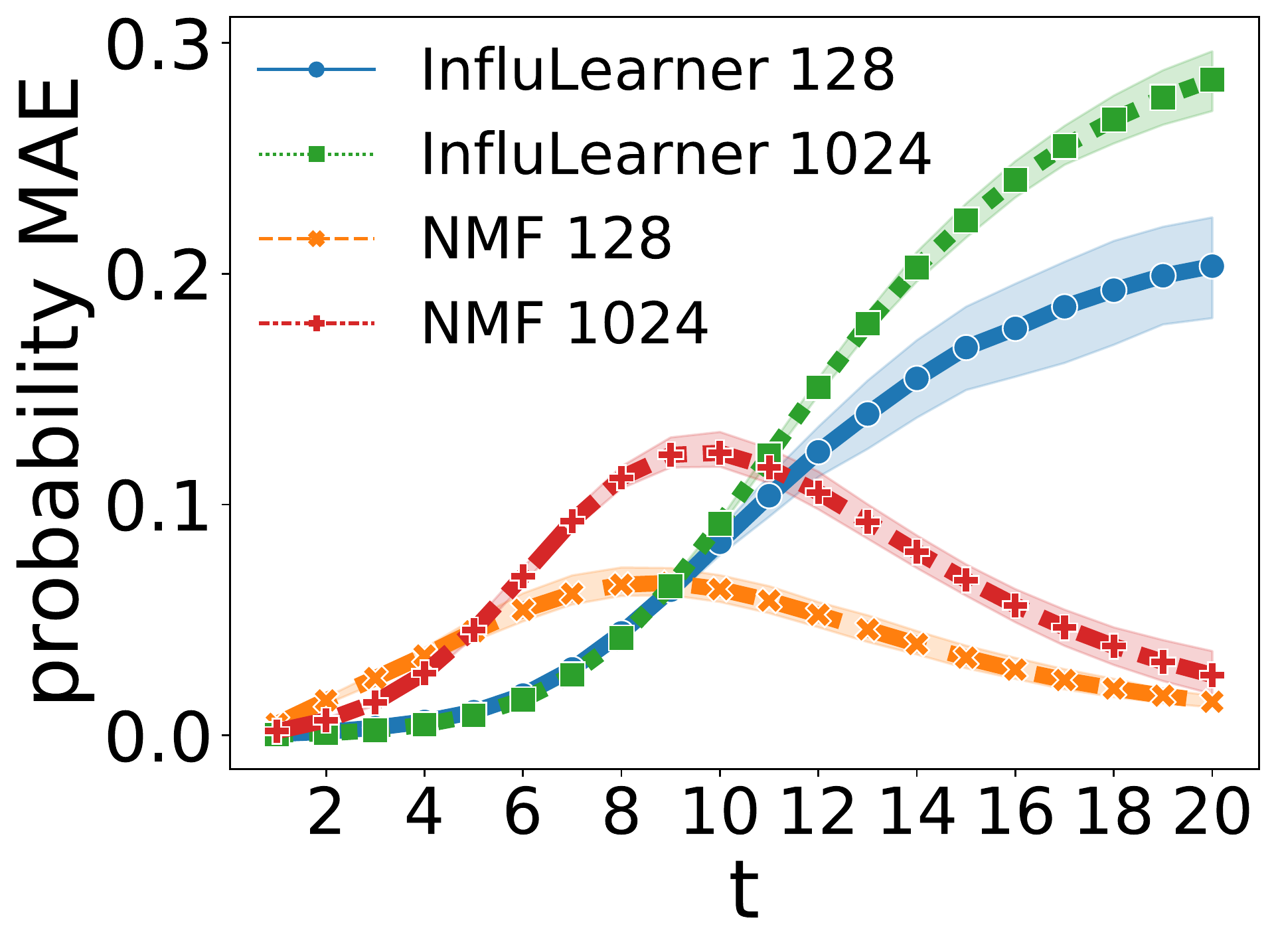}
\caption{Core + Wbl}
\end{subfigure}\\
\begin{subfigure}[b]{.24\textwidth}
\includegraphics[width=\textwidth]{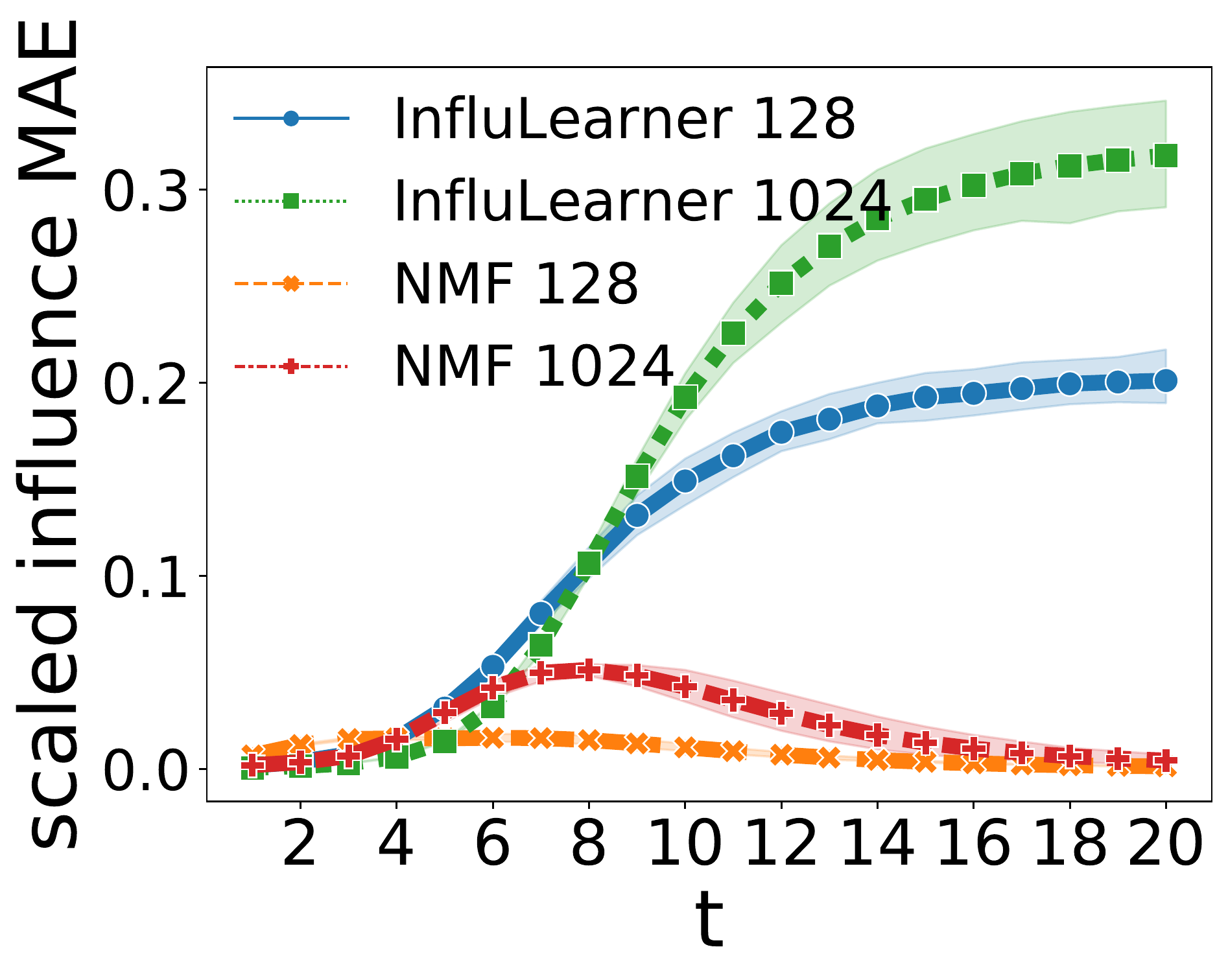}
\end{subfigure}
\begin{subfigure}[b]{.24\textwidth}
\includegraphics[width=\textwidth]{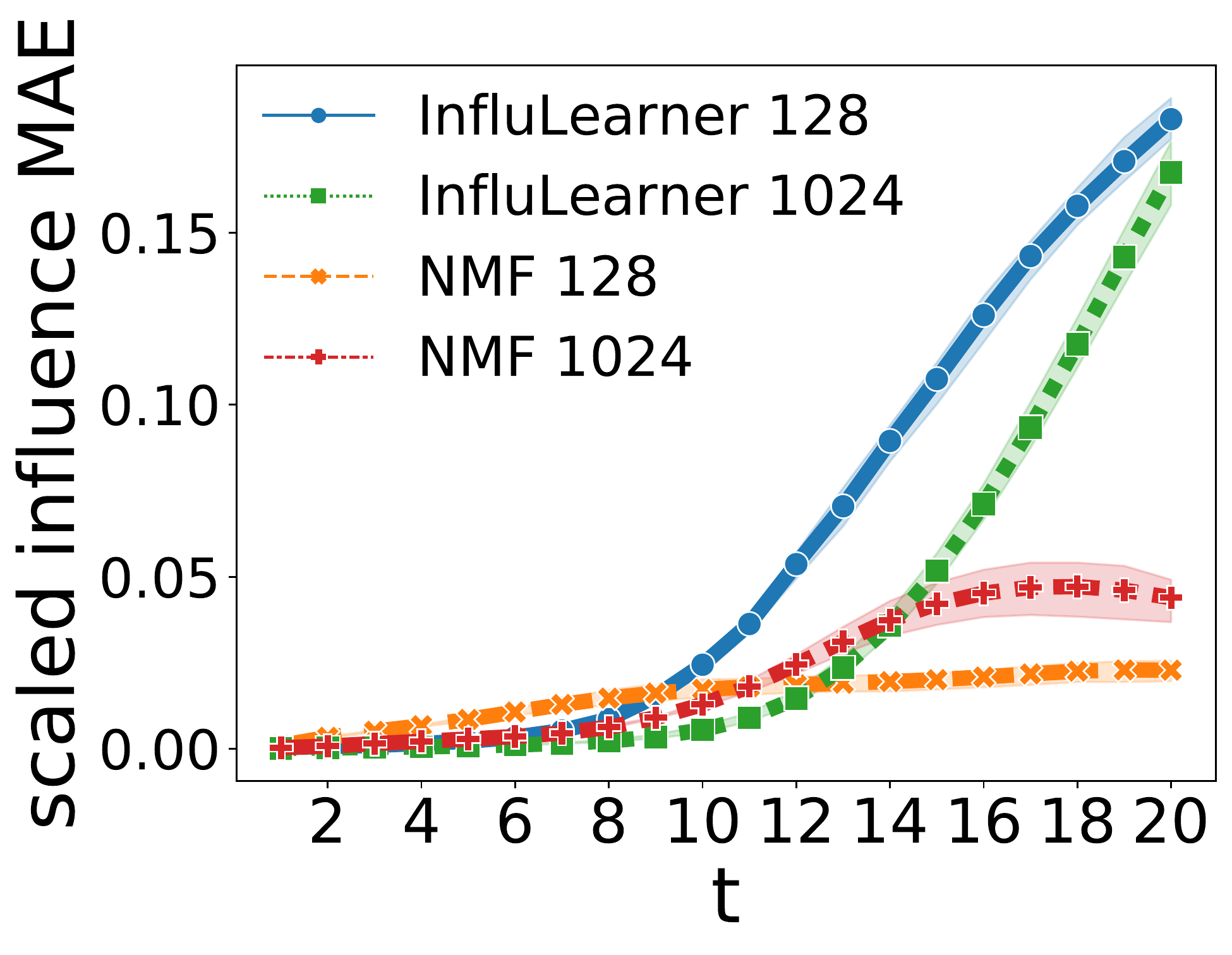}
\end{subfigure}
\begin{subfigure}[b]{.24\textwidth}
\includegraphics[width=\textwidth]{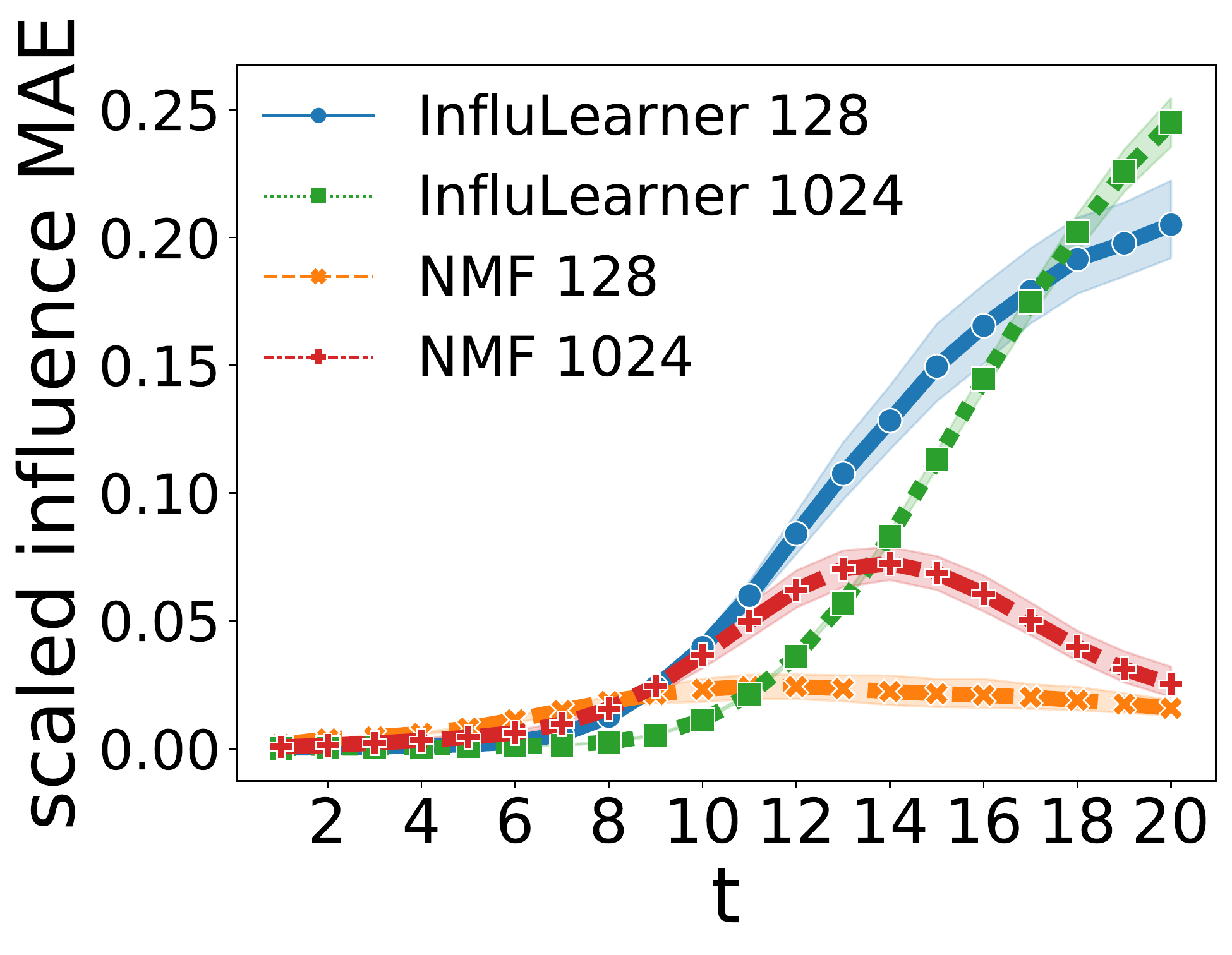}
\end{subfigure}\\
\begin{subfigure}[b]{.24\textwidth}
\includegraphics[width=\textwidth]{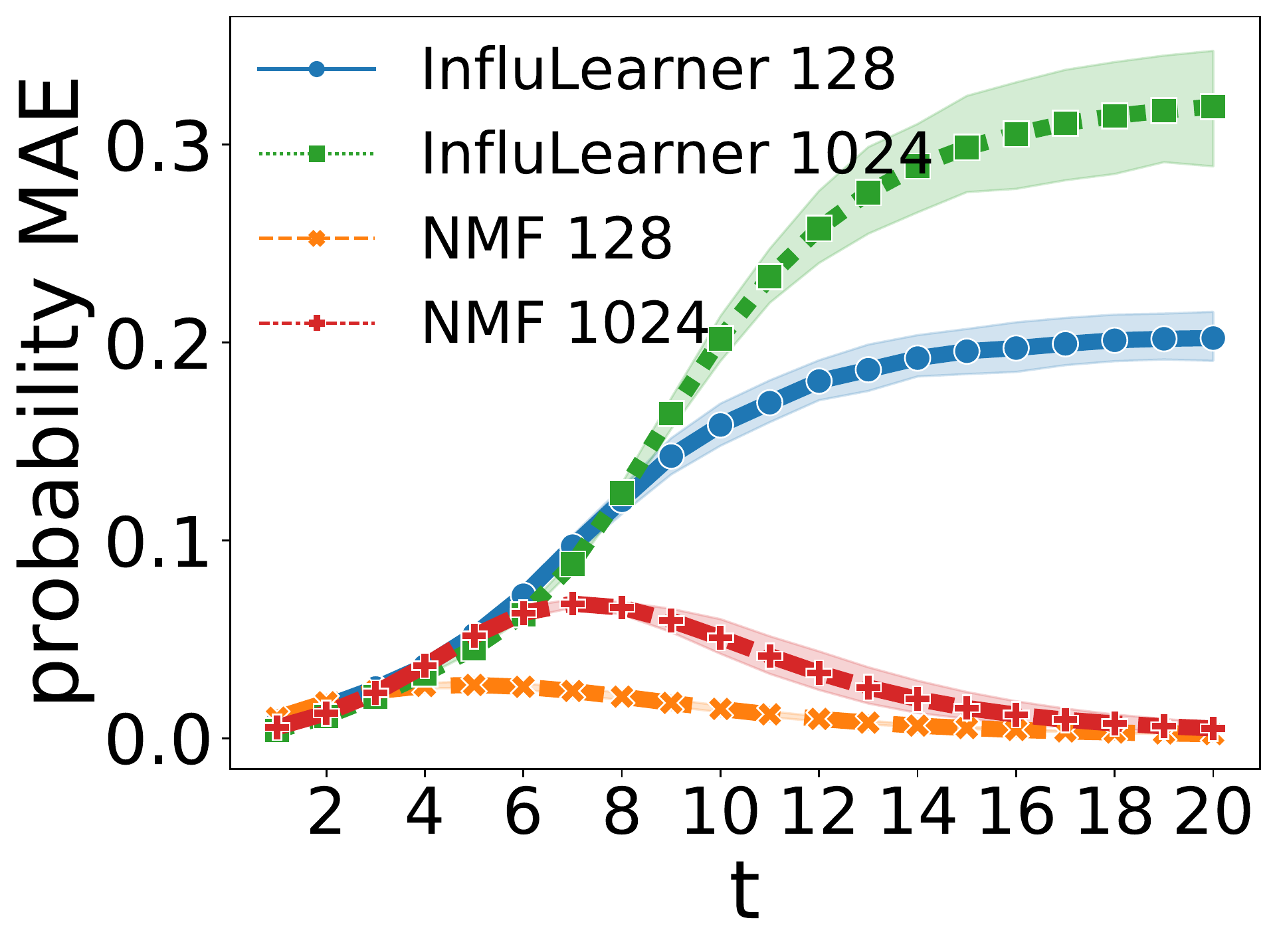}
\caption{Rand + Exp}
\end{subfigure}
\begin{subfigure}[b]{.24\textwidth}
\includegraphics[width=\textwidth]{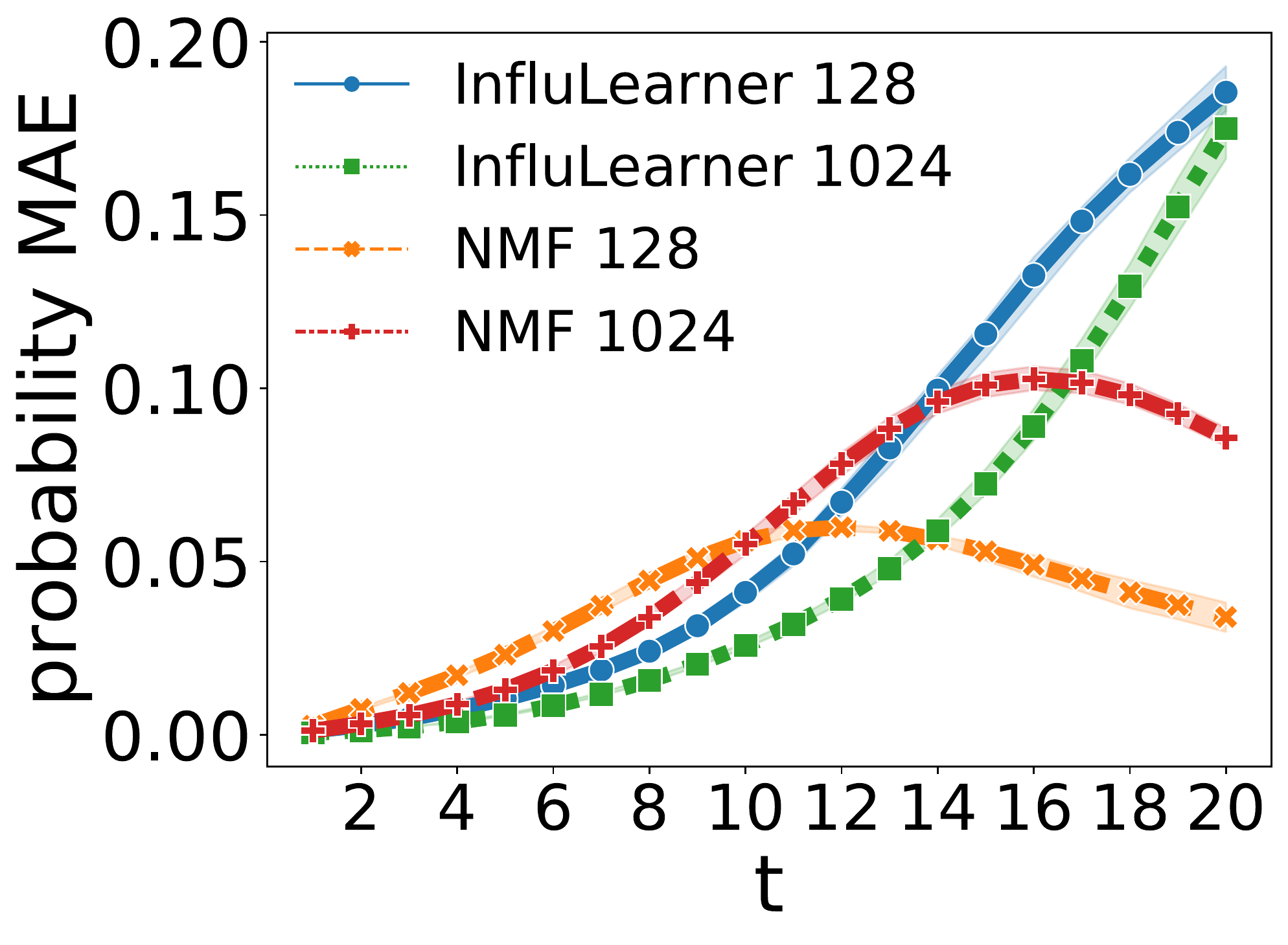}
\caption{Rand + Ray}
\end{subfigure}
\begin{subfigure}[b]{.24\textwidth}
\includegraphics[width=\textwidth]{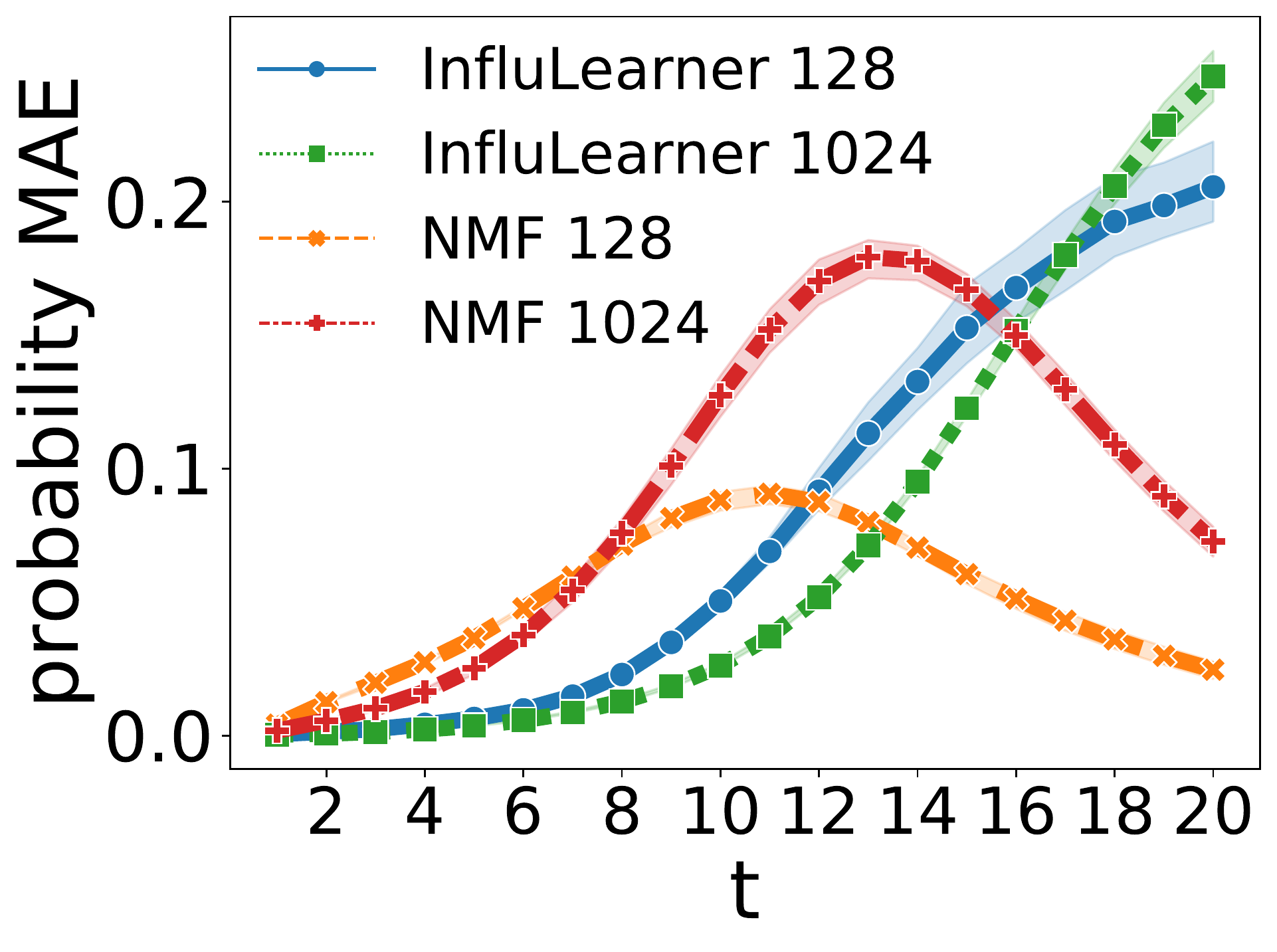}
\caption{Rand + Wbl}
\end{subfigure}\\
\begin{subfigure}[b]{.24\textwidth}
\includegraphics[width=\textwidth]{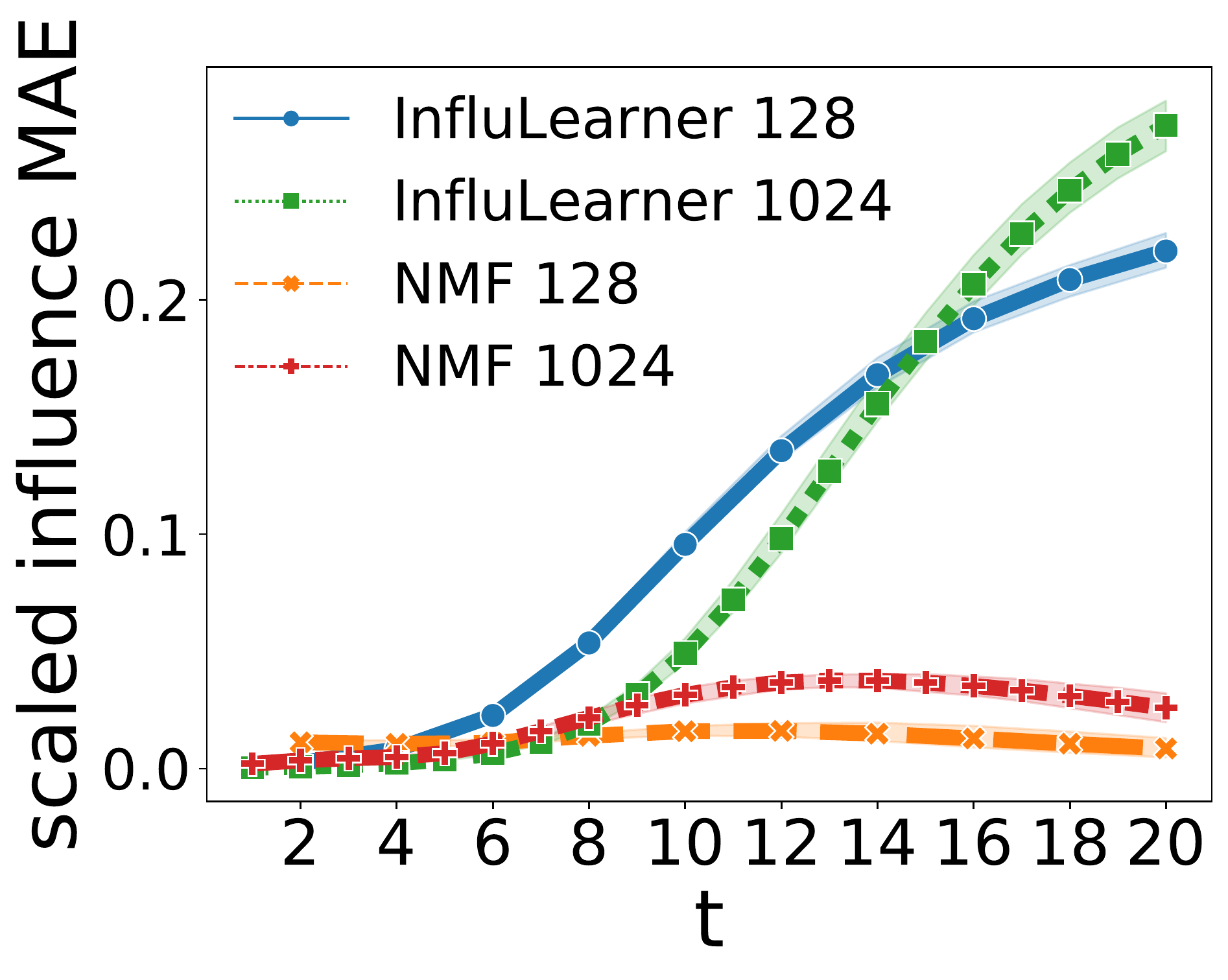}
\end{subfigure}
\begin{subfigure}[b]{.24\textwidth}
\includegraphics[width=\textwidth]{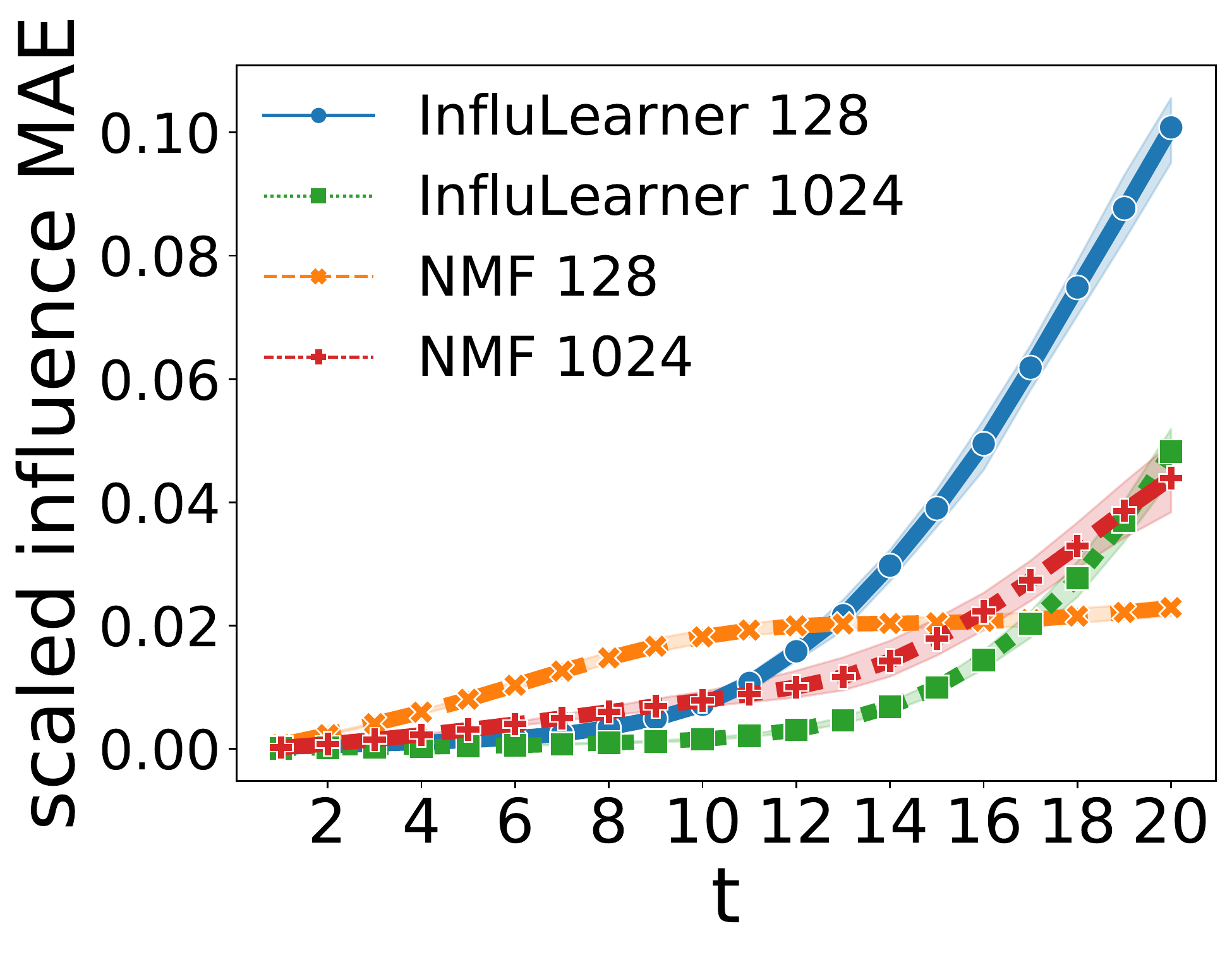}
\end{subfigure}
\begin{subfigure}[b]{.24\textwidth}
\includegraphics[width=\textwidth]{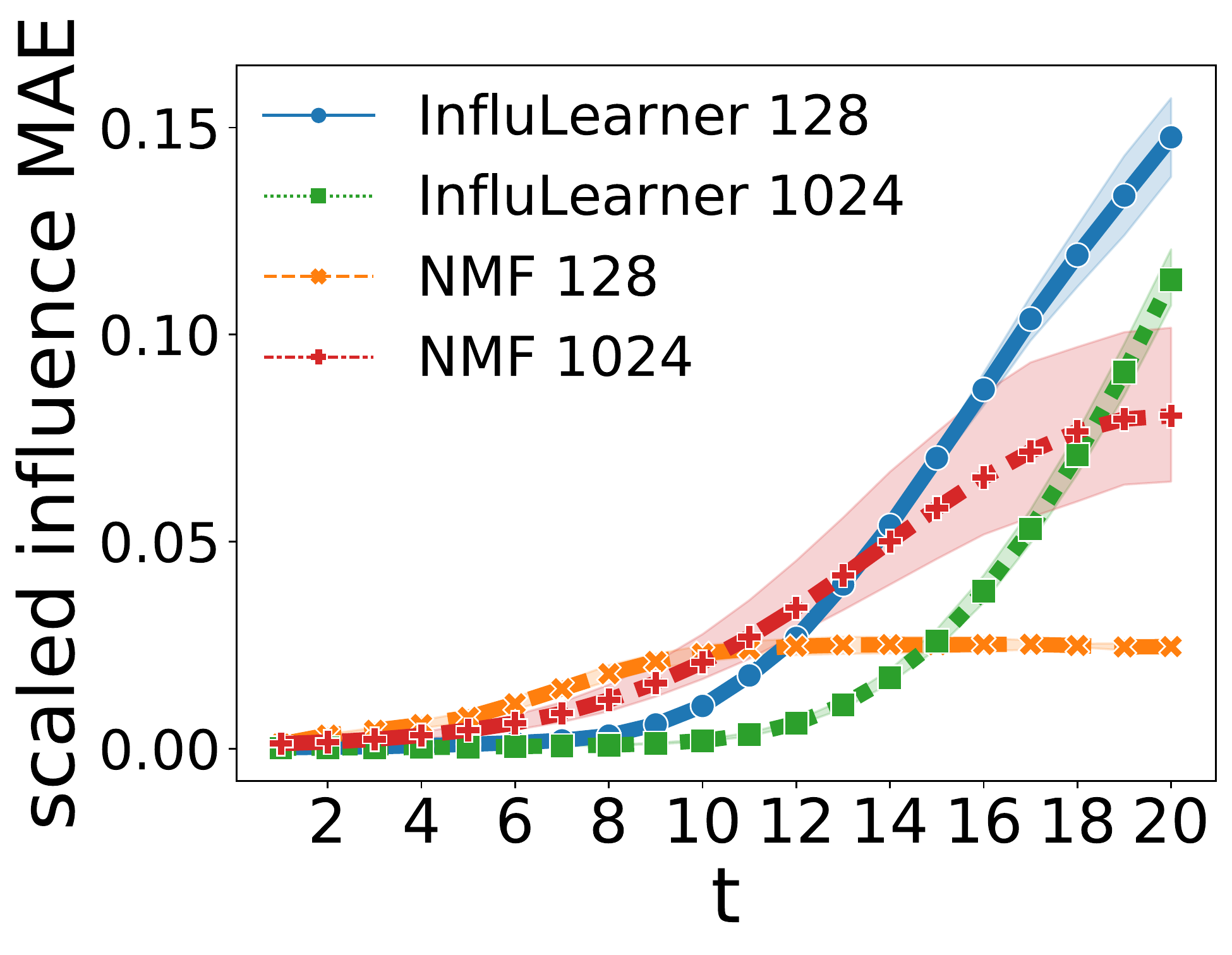}
\end{subfigure}\\
\begin{subfigure}[b]{.24\textwidth}
\includegraphics[width=\textwidth]{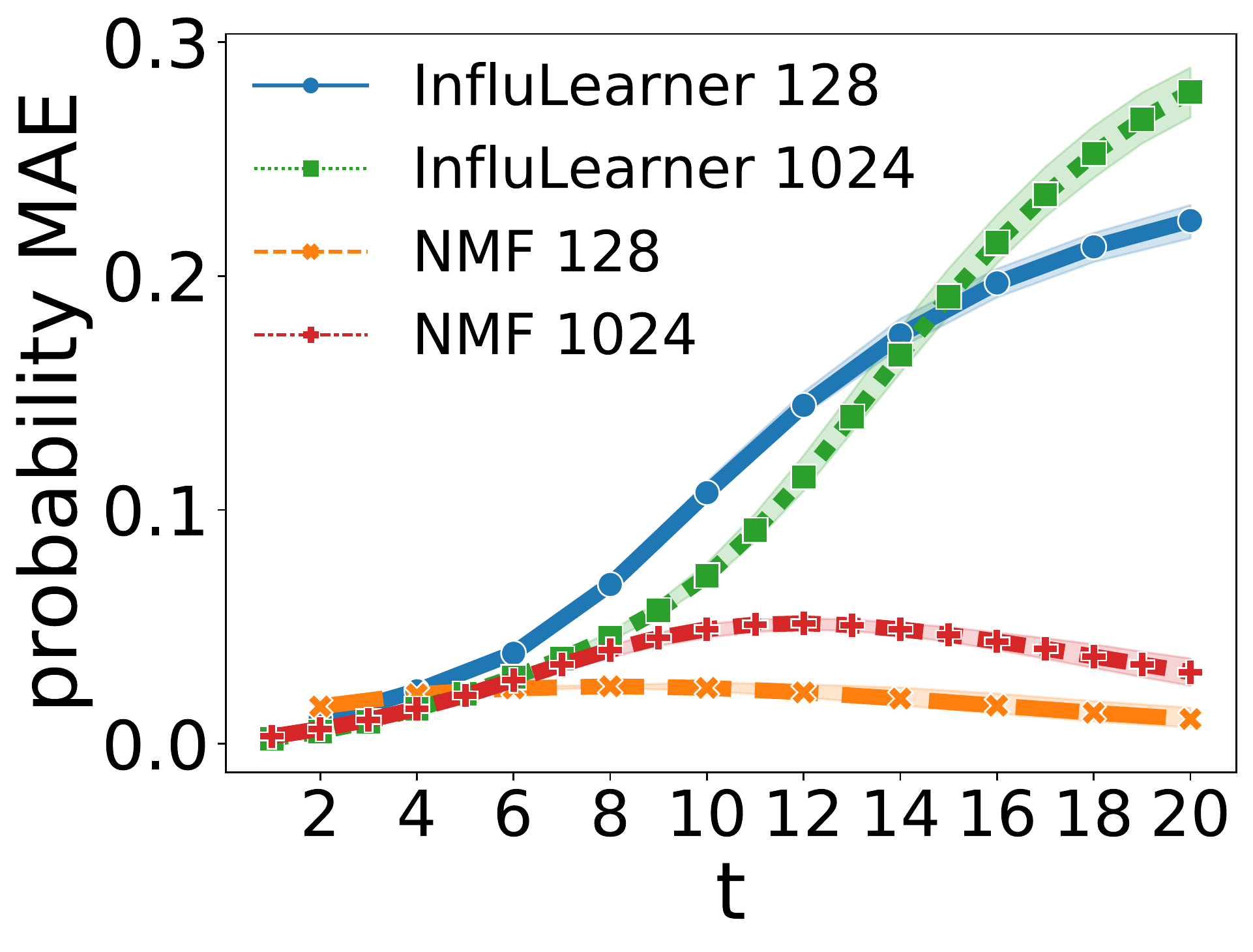}
\caption{Hier + Exp}
\end{subfigure}
\begin{subfigure}[b]{.24\textwidth}
\includegraphics[width=\textwidth]{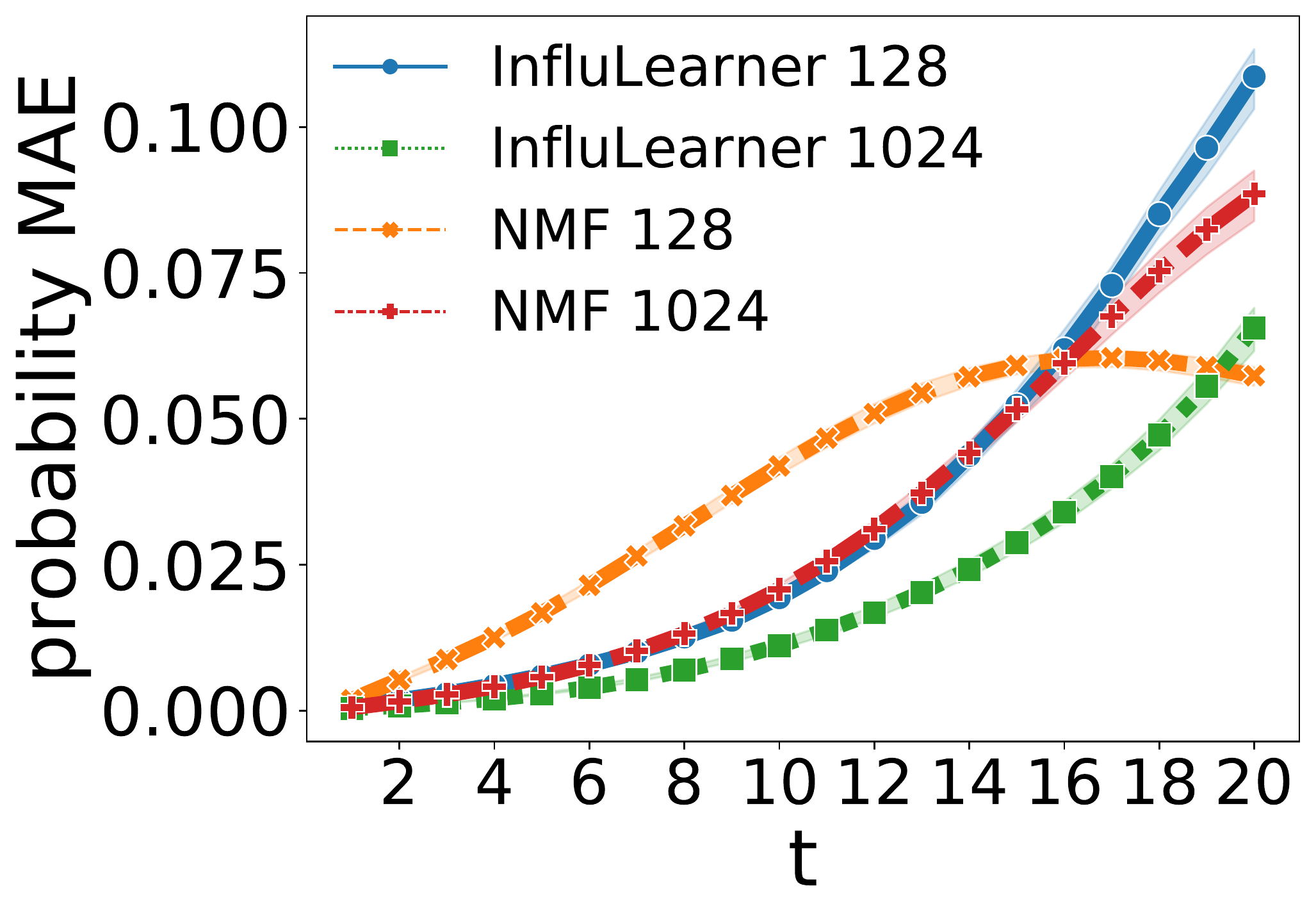}
\caption{Hier + Ray}
\end{subfigure}
\begin{subfigure}[b]{.24\textwidth}
\includegraphics[width=\textwidth]{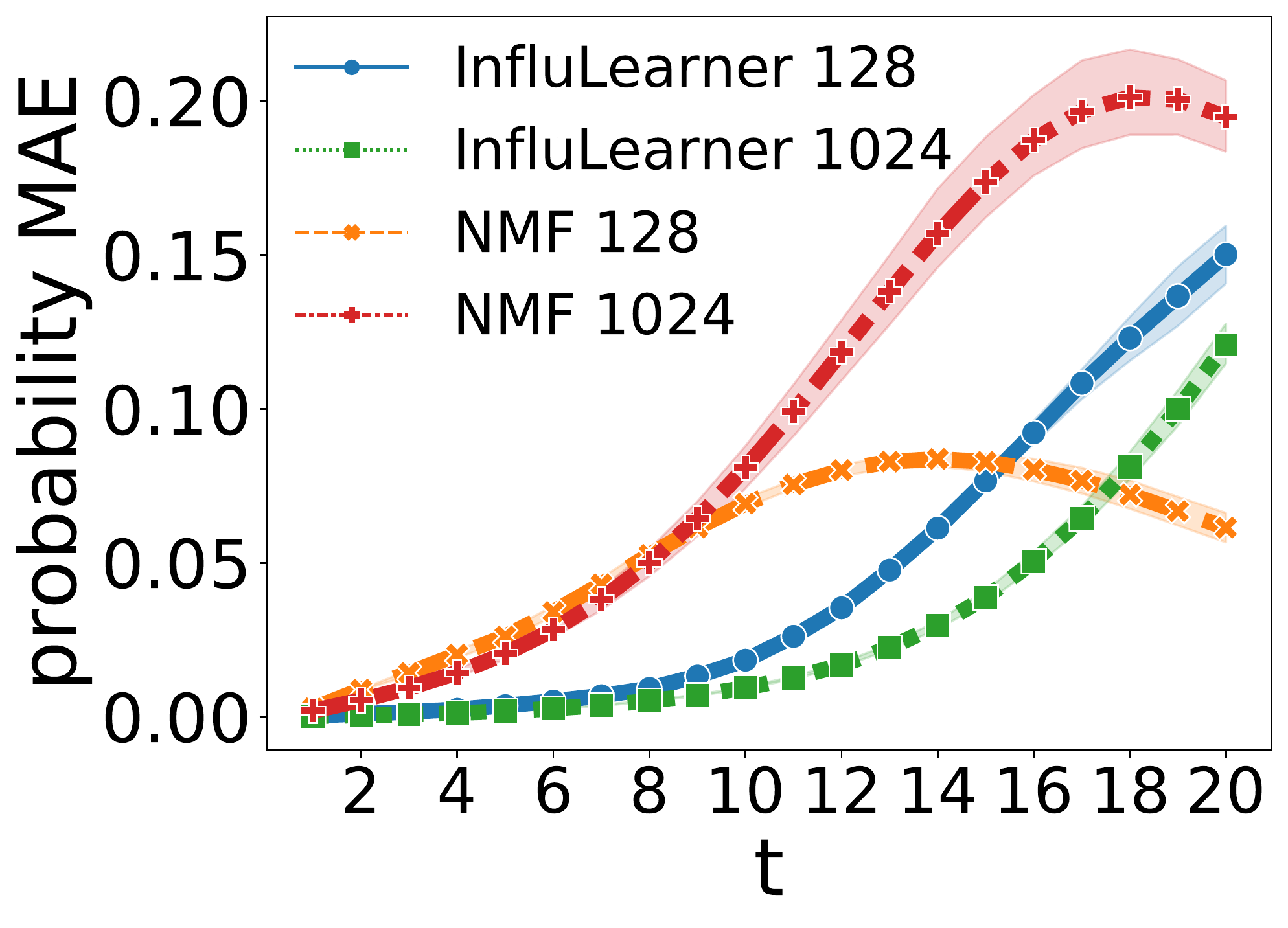}
\caption{Hier + Wbl}
\end{subfigure}
\vspace{-9pt}
\caption{MAE of scaled influence (top) and node infection probability (bottom) by InfluLearner \cite{du2014influence} and NMF on each of the 9 different combinations of Core-periphery (Core), Random (Rand) and Hierarchical (Hier) networks, and exponential (Exp), Rayleigh (Ray) and Weibull (Wbl) diffusion models. Mean (centerline) and standard deviation (shade) over 50 test source sets are shown. Each network has two configurations of $(n,d)$: $(128,4)$ and $(1024,4)$, where $n$ is the number of nodes in the diffusion network, and $d$ is the average out-degree per node.}
\label{fig:main}
\end{figure}

We show the numerical results of InfluLearner and NMF for influence estimation on the three aforementioned synthetic diffusion networks (i.e., Hier, Core, and Rand) in Figure \ref{fig:main}. For each of these three networks, we simulate three types of diffusion times (i.e., Exp, Ray, and Wbl). Therefore, we have 9 network/diffusion combinations in total. For each of these 9 combinations, we show the scaled influence MAE (top) and probability MAE (bottom) of InfluLearner and NMF on networks of size 128 and 1024 as explained above. In each plot of Figure \ref{fig:main}, we show the mean (center line) and standard deviation (shade) averaged over 5 instances. 

As we can observe in Figure \ref{fig:main}, the error of NMF is much smaller than that of InfluLearner for almost all times, except at some early stages and on Hierarchical network with Weibull distribution. 

This demonstrates that NMF is a much more accurate method in influence estimation.

%%%%%%%%%%%%%%%%%%%%%%%%%%%%%%%%128-1024%%%%%%%%%%%%%%%%%%%%%%%%%%%%

\subsection{Network structure inference}
In addition to influence estimation, the proposed NMF can also learn the network structure and the transmission rate matrix $\Abm$ as a byproduct during training.
In this test, we examine the quality of the learned $\Abm$. We set the recovered adjacency matrix $\Ecal$ to the binary indicator matrix $\Abm^{\top} \ge \epsilon$. More precisely, once we learned $\Abm$ in NMF training, we set the edge $\Ecal$ as $\Ecal_{ij} = 1$ if $\alpha_{ij}=(\Abm)_{ji} \ge  0.01$ and $0$ otherwise. We set the threshold $\epsilon = 0.01$ because all the transmission rates are between $[0.01,1]$. 

\paragraph{Evaluation criteria}
To evaluate the quality of $\Ecal$ and $\Abm$, we use four metrics: precision (Prc), recall (Rcl), accuracy (Acc), and correlation (Cor), defined as follows,
\begin{align*}
    \text{Prc}(\Ecal,\Ecal^*) & = \textstyle\frac{|\Ecal \cap \Ecal^*|}{|\Ecal^*|}, \ \ \qquad \qquad
    \text{Rcl}(\Ecal,\Ecal^*) = \textstyle\frac{|\Ecal \cap \Ecal^*|}{|\Ecal|}, \\
    \text{Acc}(\Ecal,\Ecal^*) & = 1 - \textstyle\frac{|\Ecal - \Ecal^*|}{|\Ecal| + |\Ecal^*|}, \qquad
    \text{Cor}(A,A^*) = \textstyle\frac{|\mathrm{tr}(A^{\top} A^*)|}{\|A\|_F \|A^*\|_F},
\end{align*}
where $|\Ecal|$ counts the number of nonzero entries in $\Ecal$, and the $\Ecal^*$ and $\Abm^*$ are the ground truths, respectively.
In Cor, $\|A\|_F^2=\mathrm{tr}(A^{\top}A)$ is the Frobenius norm of the matrix $A$.
Prc is the ratio of edges in $\Ecal^*$ that are recovered in $\Ecal$.
Rcl is the ratio of correctly recovered edges in $\Ecal$.
Acc indicates the ratio of the number of common edges shared by $\Ecal$ and $\Ecal^*$ against the total number of edges in them.
Cor measures similarity between $A$ and $A^*$ by taking their values into consideration.
All metrics are bounded between $[0,1]$, and higher value indicates better accuracy.

\paragraph{Comparison algorithm}
For comparison purpose, we also applied N\textsc{et}R\textsc{ate} \cite{gomez-rodriguez2011uncovering}, a state-of-the-art algorithm that uncovers the network structure and transmission rates from cascade data.
It is worth noting that N\textsc{et}R\textsc{ate} requires the knowledge of the specific diffusion model (e.g., Exp, Ray, or Wbl), so that the likelihood function can be explicitly expressed.
Moreover, N\textsc{et}R\textsc{ate} can only estimate $\Abm$ of diffusion networks, but not the influence. In contrast, NMF tackles both network inference and influence extimation simultaneously.
In terms of computation efficiency, we observed that the implementation of N\textsc{et}R\textsc{ate} provided in \cite{gomez-rodriguez2011uncovering} runs very slowly for large networks. Therefore, we only perform comparisons on networks of size $n=128$ in this experiment.   

\begin{figure}[ht]
    \begin{center}
    \begin{subfigure}{.24\textwidth}
        \includegraphics[width=\textwidth]{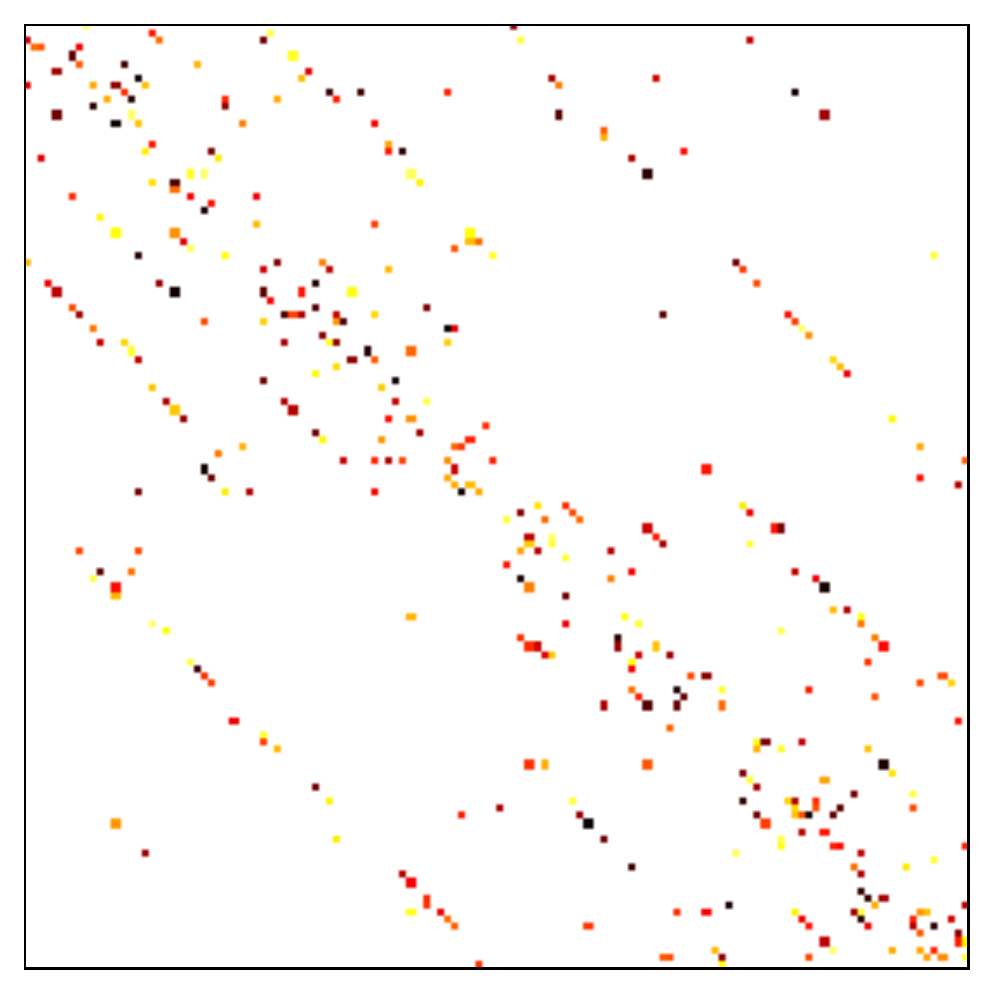}
        \caption{True}
        \label{subfig:A_true}
    \end{subfigure}  
    \begin{subfigure}{.24\textwidth}
        \includegraphics[width=\textwidth]{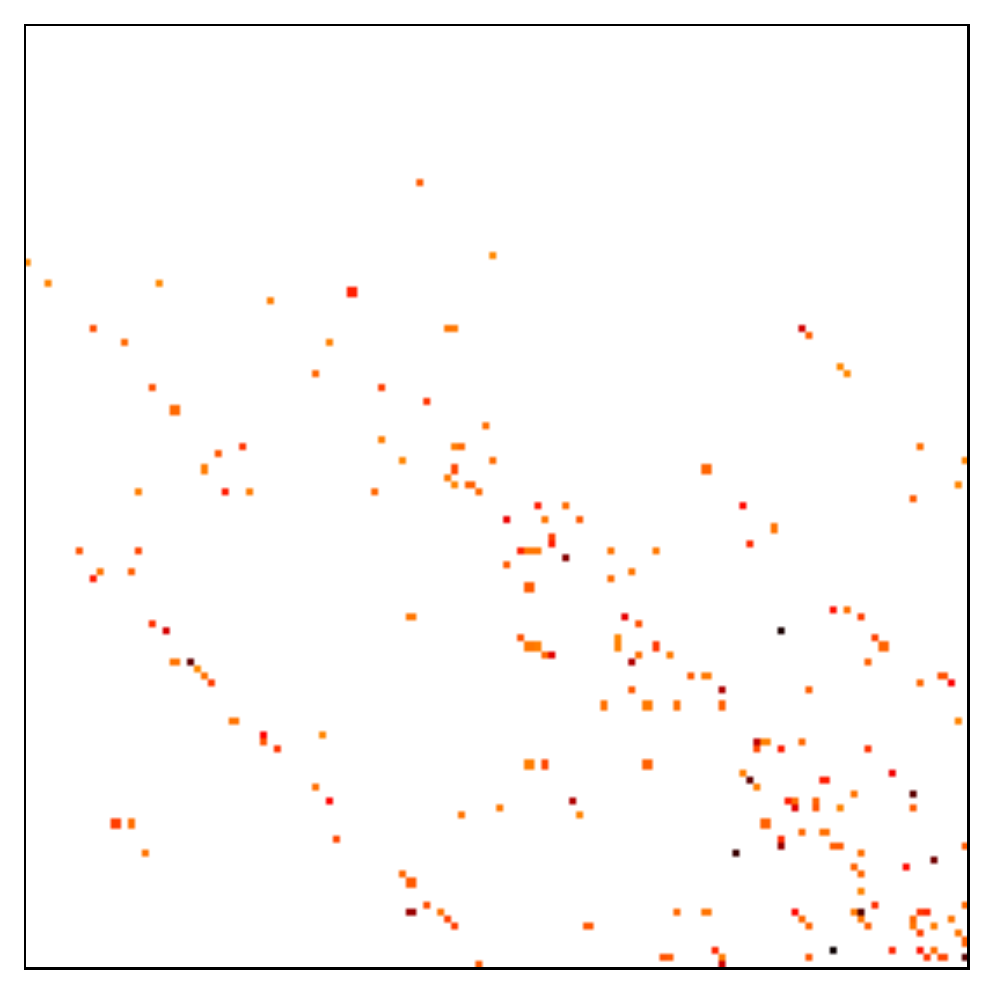}
        \caption{ N\textsc{et}R\textsc{ate}}
        \label{subfig:A_Net}
    \end{subfigure}
    \begin{subfigure}{.24\textwidth}
        \includegraphics[width=\textwidth]{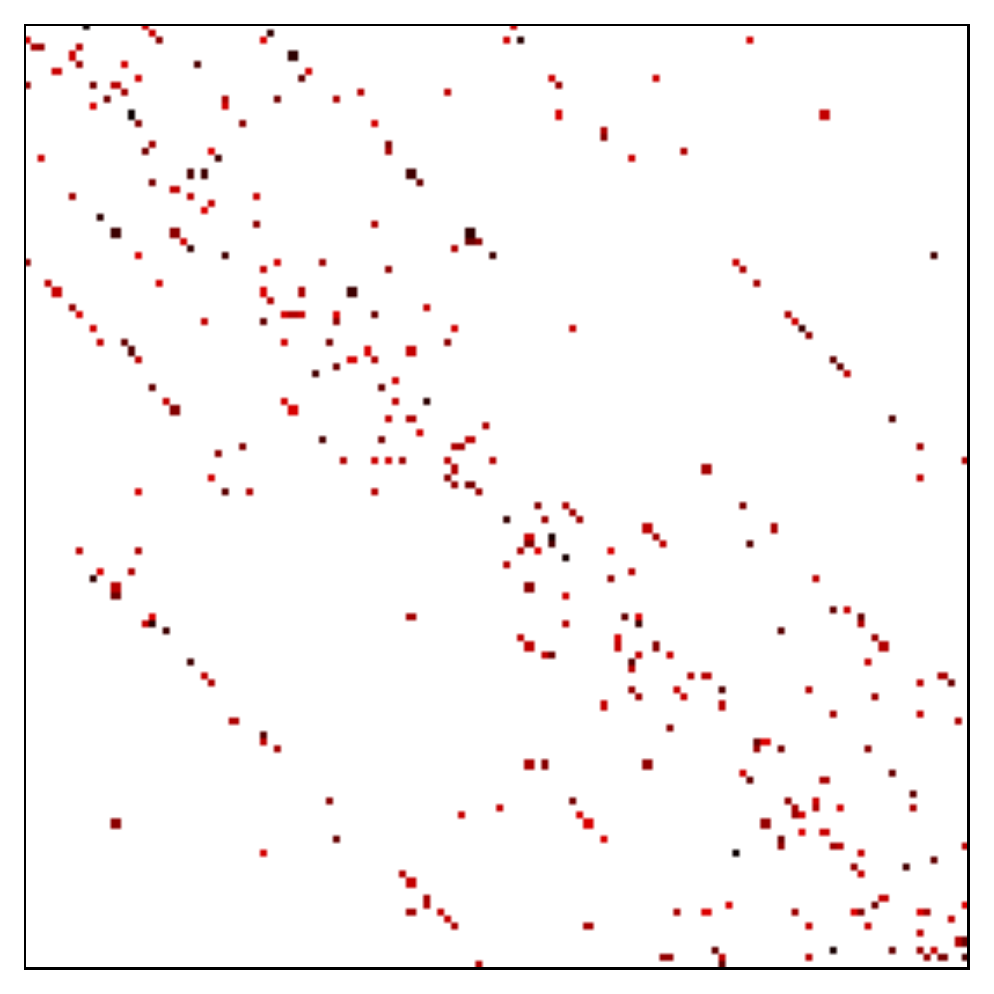}
        \caption{NMF}
    \label{subfig:A_NMF}
    \end{subfigure}
     \end{center}
    \caption{Ground truth $\Abm^*$ (left) and $\Abm$ inferred by N\textsc{et}R\textsc{ate} (middle) and NMF (right) in same color scale using cascades from a Hierarchical network consisting of 128 nodes and 512 edges with exponential diffusion model. Darker pixel indicates larger value of an entry of $\Abm$.}
    \label{fig:CompA}
\end{figure}

\paragraph{Comparison results}
We compared the estimated $\Ecal$ and $\Abm$ using N\textsc{et}R\textsc{ate} and NMF using the four criteria mentioned above in Table \ref{tab:withNetrate} for three types of networks (Random, Hierarchical, and Core-periphery) and two diffusion models (Exponential and Rayleigh).
In all of these tests, NMF consistently outperforms N\textsc{et}R\textsc{ate} in all accuracy metrics.
\begin{table}[t]
    \caption{Performance of network structure inference using N\textsc{et}R\textsc{ate} \cite{gomez-rodriguez2011uncovering} and the proposed NMF on Random, Hierarchical, and Core-periphery networks consisting of 128 nodes and 512 edges with Exponential and Rayleigh as diffusion distribution on edges. Quality of the learned edge set $\Ecal$ and distribution parameter $\Abm$ are measured by precision (Prc), recall (Rcl), accuracy (Acc), and correlation (Cor). Larger value indicates higher accuracy.}
    \label{tab:withNetrate}
    \begin{center}
    \begin{tabular}{llccccr}
    \toprule
    Diffusion & Network & Method & Prc & Rcl & Acc & Cor\\
    \midrule
         \multirow{6}{6.2em}{Exponential} &\multirow{2}{6.2em}{Random} &N\textsc{et}R\textsc{ate} & 0.457&0.821         &0.515 & 0.438\\
        &    &NMF &\textbf{0.459}&\textbf{0.997} &\textbf{0.622}&\textbf{0.910} \\
	\cmidrule{2-7}
    %%%%%%%%%
       &\multirow{2}{6.2em}{Hierarchical} &N\textsc{et}R\textsc{ate} & 0.395 &0.748 &0.515 &0.739\\
	%%%%%%%%%%%%% 
    &    &NMF &\textbf{0.595}&\textbf{0.997} &\textbf{0.745}&\textbf{0.928}\\
	\cmidrule{2-7}
    %%%%%%%%%%
    &\multirow{2}{6.2em}{Core-periphery} &N\textsc{et}R\textsc{ate} & 0.277&0.611 & 0.264& 0.264\\
    &    &NMF & \textbf{0.292}&\textbf{0.997}&\textbf{0.450}&\textbf{0.839} \\
        \midrule
    %%%%%%%%%%
     \multirow{6}{6.2em}{Rayleigh} &\multirow{2}{6.2em}{Random} &N\textsc{et}R\textsc{ate} & 0.481&0.399 &0.434 & 0.465\\
        &    &NMF &\textbf{0.883}&\textbf{0.905} &\textbf{0.894}&\textbf{0.909} \\
           \cmidrule{2-7}            
       &\multirow{2}{6.2em}{Hierarchical} &N\textsc{et}R\textsc{ate}&0.659&0.429 &0.519&0.464\\
    &    &NMF &\textbf{0.889}&\textbf{0.936} &\textbf{0.911}&\textbf{0.913}\\
        \cmidrule{2-7}
       &\multirow{2}{6.2em}{Core-periphery}     &N\textsc{et}R\textsc{ate} & 0.150&0.220& 0.178& 0.143\\
    &    &NMF & \textbf{0.649}&\textbf{0.820} &\textbf{0.724}&\textbf{0.820} \\
        \bottomrule 
    \end{tabular}
    \end{center}
\end{table}
We also draw $\Abm$ inferred by N\textsc{et}R\textsc{ate} and NMF for a visual comparison in Figure \ref{fig:CompA}. In Figure \ref{fig:CompA}, we show the ground truth $\Abm^*$ (left), the matrix $\Abm$ inferred by N\textsc{et}R\textsc{ate} (middle), and $\Abm$ learned by NMF (right). The values of $\alpha_{ij}$ are indicated by the color---the darker the red is, the higher the value of $\alpha_{ij}$---and the white pixels represent where $\alpha_{ij}$ is zero.
As we can see, $\Abm$ learned by NMF is much more faithful to $\Abm^*$ than that by N\textsc{et}R\textsc{ate}. %
This result shows that NMF is very versatile and robust in learning network structure from cascade data. 

Since N\textsc{et}R\textsc{ate} code \cite{gomez-rodriguez2011uncovering} was implemented in MATLAB and is executed on CPU in our experiment, the computation times of N\textsc{et}R\textsc{ate} and NMF cannot be directly compared. However, we notice that N\textsc{et}R\textsc{ate} takes approximately 10+ hours on average to infer each network structure $\Abm$ in Table \ref{tab:withNetrate}, whereas NMF only requires about 300 seconds on average to return both more accurate $\Abm$ and an influence estimation mechanism.

\begin{figure}[t]
    \begin{center}
    \begin{subfigure}{0.24\textwidth}
    \includegraphics[width=\textwidth]{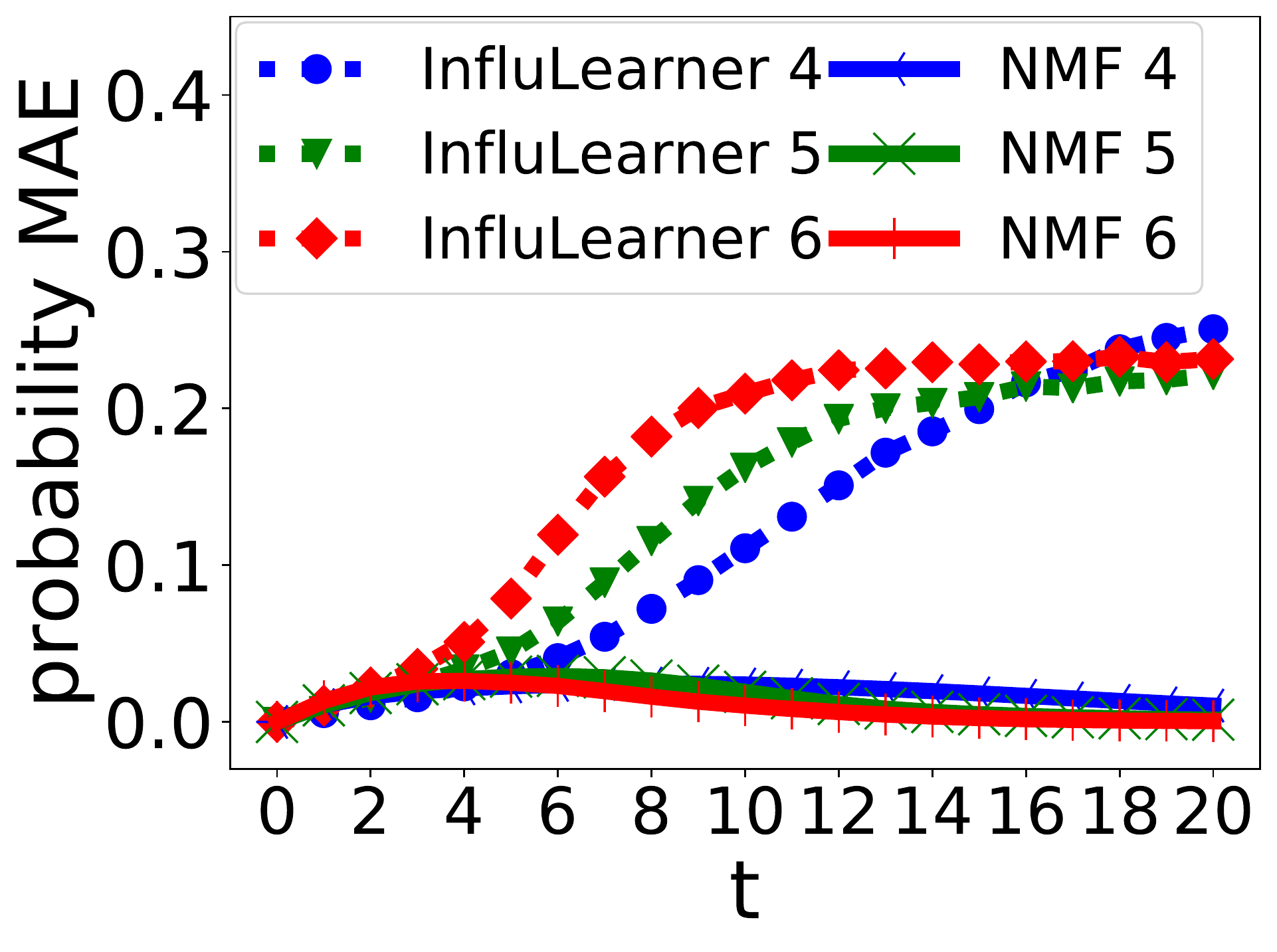}
    \caption{Probability MAE}
    \label{subfig:prob_d}
    \end{subfigure}
    \begin{subfigure}{0.24\textwidth}
    \includegraphics[width=\textwidth]{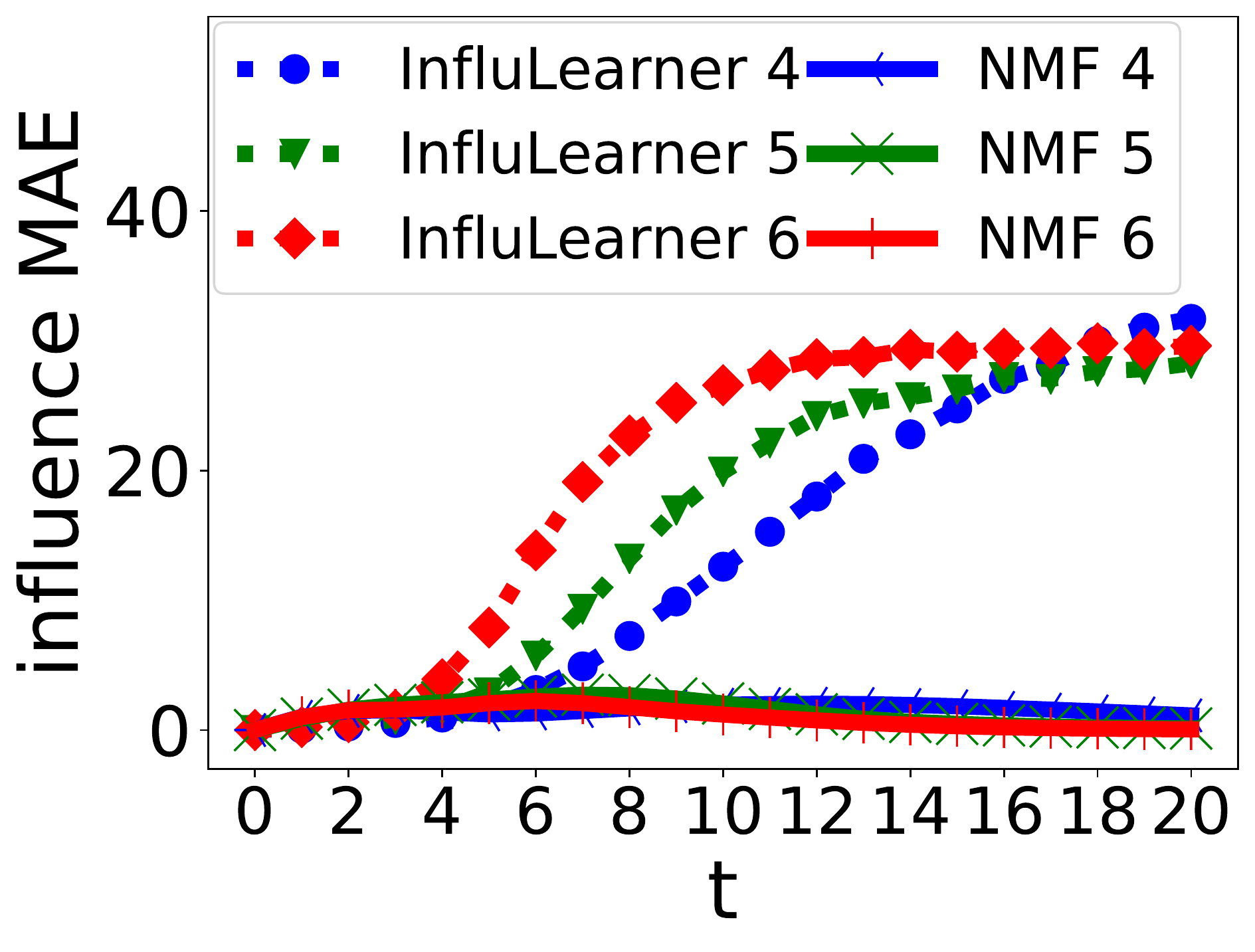}
    \caption{Influence MAE}
    \label{subfig:inf_d}
    \end{subfigure}
    \begin{subfigure}{0.2555\textwidth}
    \includegraphics[width=\textwidth]{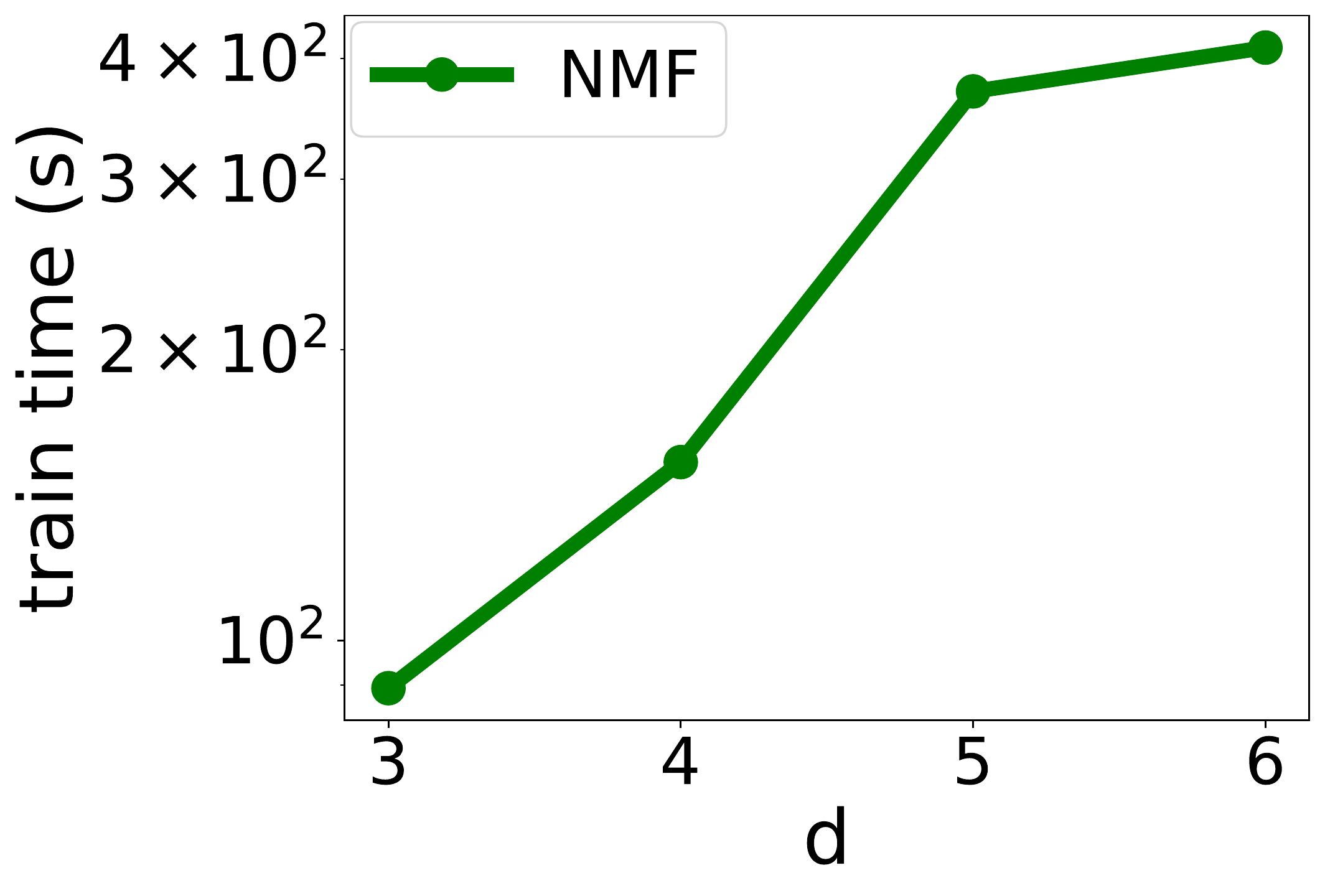}
    \caption{Train time vs $d$}
    \label{subfig:traintime_d}
    \end{subfigure}
    \begin{subfigure}{0.235\textwidth}
    \includegraphics[width=\textwidth]{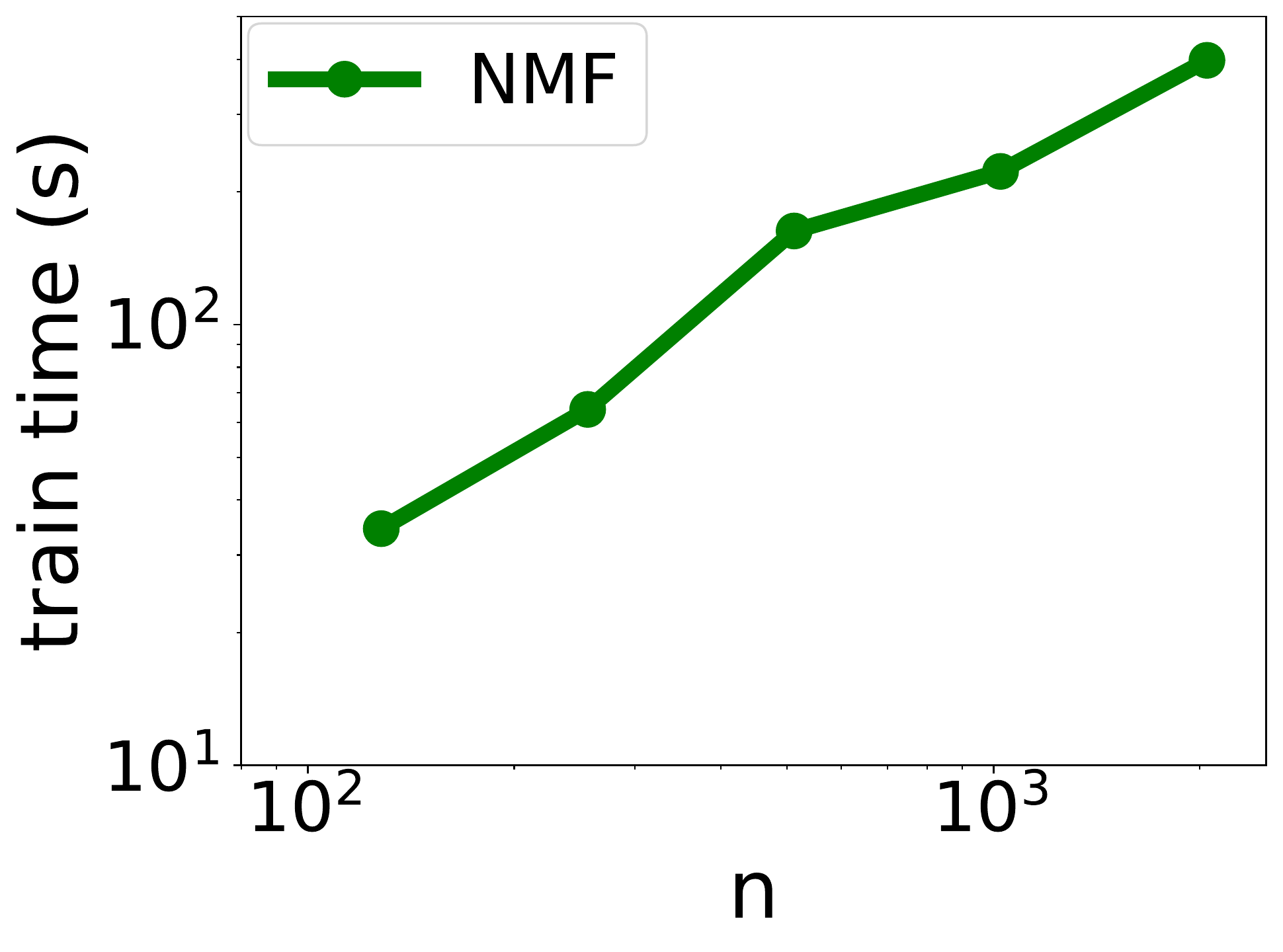}
    \caption{Train time vs $n$}
    \label{subfig:traintime_n}
    \end{subfigure}
    \end{center}
    \caption{(a)--(b) MAE of infection probability and influence obtained by InfluLearner \cite{du2014influence} and NMF on Hierarchical networks of size $n=128$ and increasing $d$ from 4 to 6. (c) Training time (in seconds) of NMF versus density (average out-degree per node) $d$. (d) Training time (in seconds) versus network size $n$. } 
    \label{fig:robust}
\end{figure} 

\begin{figure}[t]
    \begin{center}
    \begin{subfigure}{0.4\textwidth}
    \includegraphics[width=\textwidth]{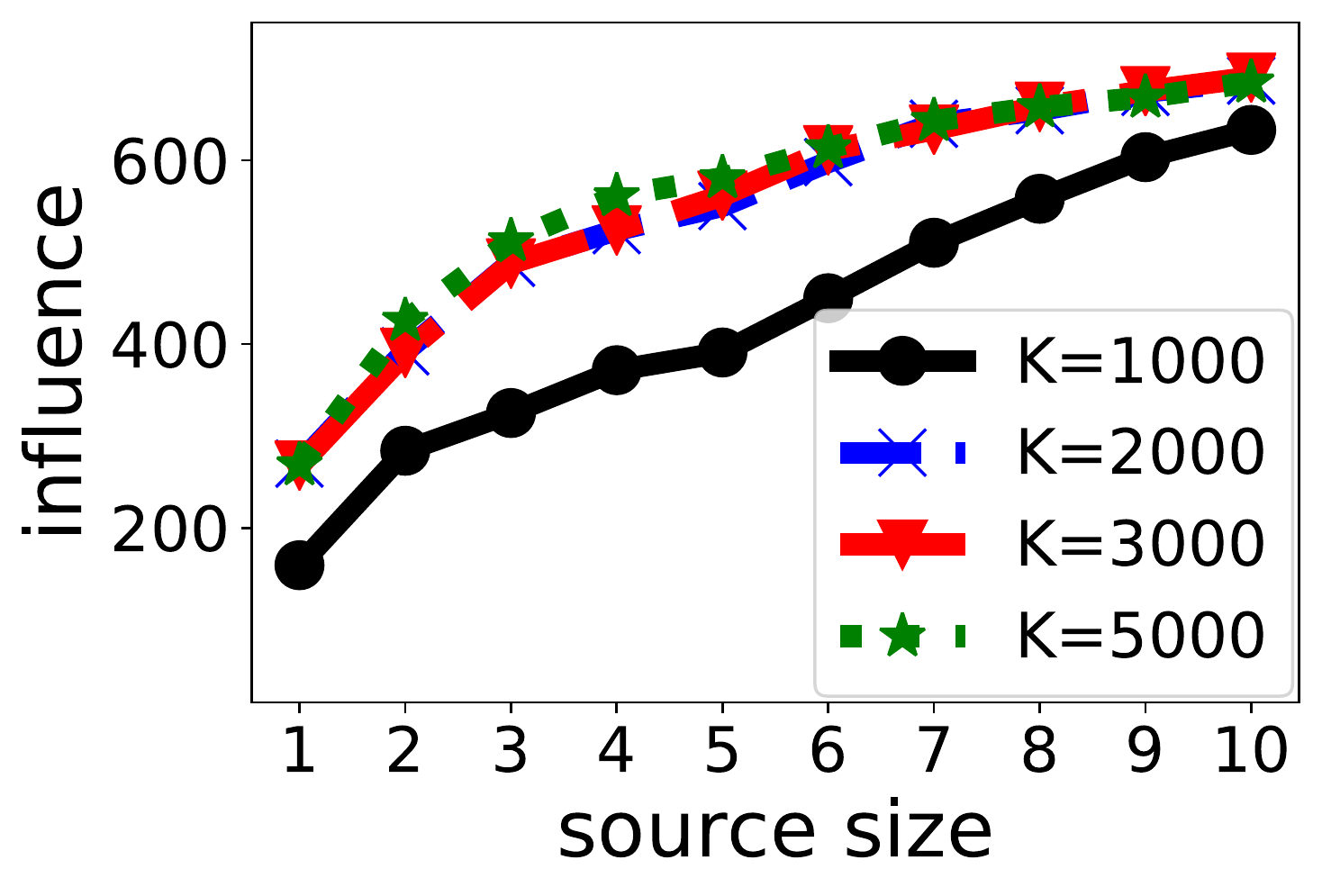}
    \caption{Varying training set size}
    \label{subfig:infmax_samplesize}
    \end{subfigure}
    \begin{subfigure}{0.36\textwidth}
    \includegraphics[width=\textwidth]{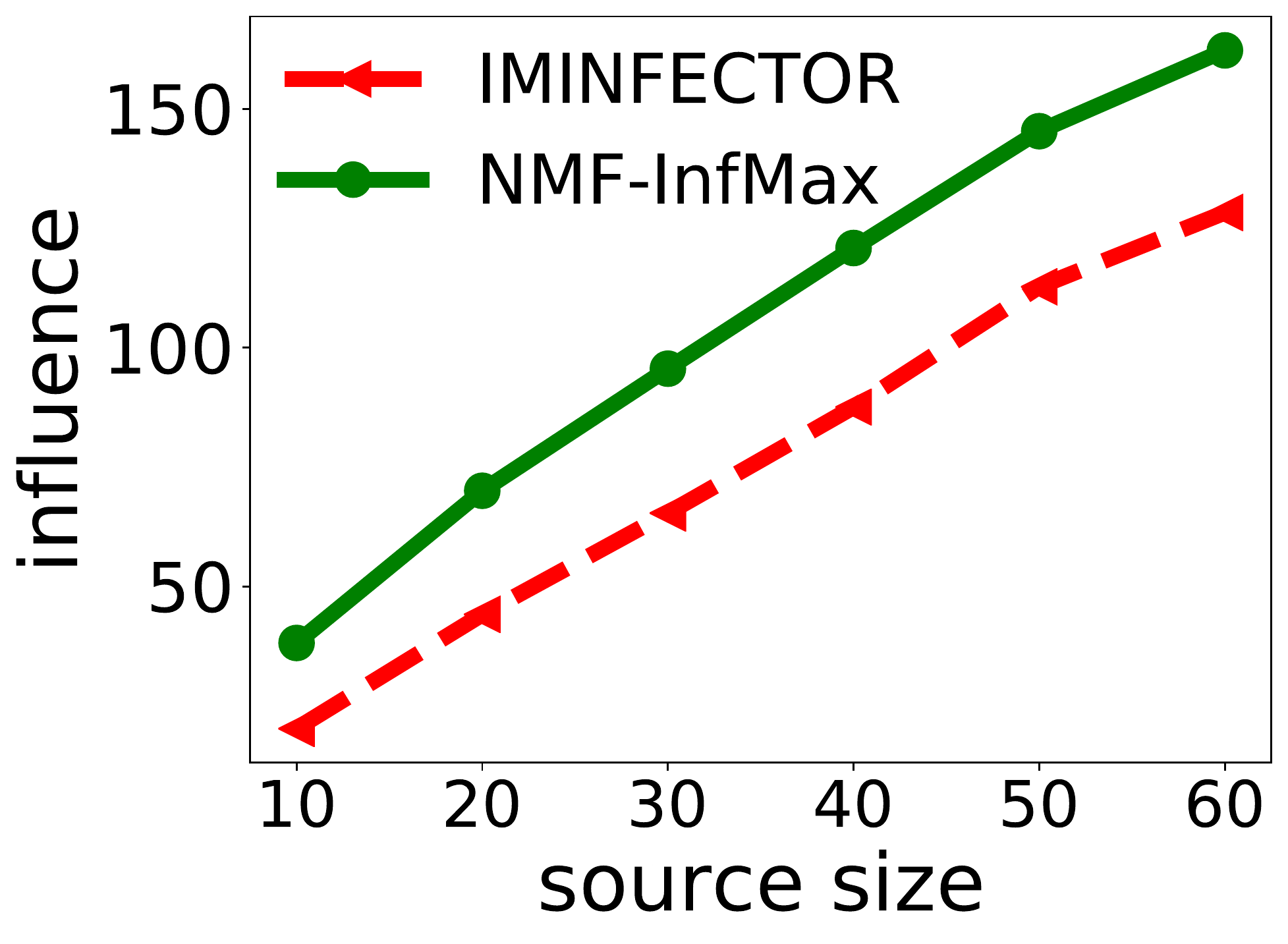}
    \caption{Influence vs $n_{0}$}
    \label{subfig:infmax_real}
    \end{subfigure}
    \end{center}
    \caption{(a) Influence generated the source sets selected by NMF-InfMax trained using increasing number of cascades on Hierarchical networks with 1,024 nodes and 4,096 edges. (b) Influence generated by the source sets selected by IMINFECTOR and NMF-InfMax on the MemeTracker dataset at $T=10$ hours.}
    \label{fig:infmax}
\end{figure}

\subsection{Scalability to network size and density}\label{sec:bynodes}
In this test, we will demonstrate the robustness of NMF in influence estimation when the network size $n$ and density $d$ vary. Recall that $d$ stands for the average out-degree per node. The larger and/or denser the network is, the more challenging the estimation becomes. In all the experiments, we use training data consisting of 9,000 cascades generated from Hierarchical network and exponential diffusion model and set the batch size to 300.

\paragraph{Network size}
Recall that we have showed in Figure \ref{fig:main} that NMF consistently outperforms than InfluLearner when the network size is set to 128 and 1024. To show the scability of NMF, 
we further test NMF on increasing network size $n$ from 128 to 2048 (with density $d=4$). 
To test the training time of NMF, we terminate the computation when the average MAE of infection probability on validation data over 20 timepoints $t_{\ell}=\ell$ ($\ell=1,2,\dots,20$) is below 0.07. 
Euler method with 40 steps is employed as the ODE solver and the learning rate of the Adam optimizer is set to 0.0001 for network with 2048 nodes. 
The training time of NMF is shown in Figure \ref{subfig:traintime_n}, which demonstrate that NMF is scalable for large network size $n$.

\paragraph{Network density}
We also test the performance of NMF for varying network density $d$. We compare the infection probability and influence MAE of InfluLearner and NMF for varying edge density $d$ set to 4, 5, and 6 on a Hierarchical network on exponential diffusion model with 128 nodes. Figure \ref{subfig:prob_d} and Figure \ref{subfig:inf_d} show that the MAE of infection probability and influence estimation obtained by InfluLearner and NMF. These two plots show that NMF is very robust when the density of the network increases by consistently generating estimates of low MAE. Figure \ref{subfig:traintime_d} shows the training time of NMF versus network density $d$ while $n=128$ is fixed. In this plot, the computation time is recorded when the training MAE at time $t_h$ is below 0.04, where $t_h$ is the time when on average half of the nodes on the network are infected as indicated by the ground truth. Here, rk4 method with 40 steps is employed as the ODE solver. Similarly, Figure \ref{subfig:traintime_n} shows the training time versus network size $n$ while $d=4$ is fixed. From Figures \ref{subfig:traintime_d} and \ref{subfig:traintime_n}, we can see that the computational cost of NMF grows approximately quadratic in density $d$ and linear in size $n$.

%%%%%%%%%%%%%%%%%%%%%%%%%%%%%%%%%%%%%%%%%%%%%%%%%%%%%%%%%%%%%%%%%%%%%%%

\subsection{Influence maximization}
This part of the experiment is dedicated to performance evaluation in influence maximization. 
Specifically, we use the trained NMF to find the optimal source set with limited budget for maximal influence by following Algorithm \ref{alg:nmf_infmax} which is referred to as NMF-InfMax. 

\paragraph{Comparison algorithms}
For comparison purpose, we also test the following methods for influence maximization. 
\begin{itemize}
    \item IMINFECTOR \cite{panagopoulos2020multi-task}:
    IMINFECTOR represents the cascade data into two datasets consisting of seed-cascade length pairs and seed-influenced node pairs to approximate the influence spread and infection probability of each node by a regression model and a probability classifier, respectively. The outputs are used to reduce the number of candidate seeds and reformulate the computation of the influence spread in a greedy solution to influence maximization. Like our method, IMINFECTOR only uses cascade data as inputs, with embedding size 50 and sampling percentage 120, trained for 50 epochs with a learning rate of 0.1. The reduction percentage $P$ is set to 100 to keep full information of cascades.
    \item IMM \cite{tang2015influence}: IMM is a reverse reachable (RR) sketch based method which applies the standard greedy algorithm for maximum coverage to derive a budget size node set that covers a large number of RR sets sampled from the given network. We consider the case when IMM return $(1-1/e-\varepsilon)$-approximate solution with $\epsilon= 0.1$ and parameter $\ell=1$, following the experiments in \cite{tang2015influence}. 
    \item InfluMax\cite{gomez-rodriguez2012influence,gomez-rodriguez2016influence}: 
    InfluMax speed up the greedy influence maximization algorithm by exploiting submodularity.
    We incorporate it with the influence estimation algorithm ConTinEst\cite{du2013scalable}. For ContinEst, we draw 10,000 random samples, each of which has 5 random labels for each node. 
\end{itemize}
Since IMM and InfluMax both require the knowledge of the transmission matrix $\Abm$, we apply N\textsc{et}R\textsc{ate} method to learn $\Abm$ from cascade data first, then feed $\Abm$ to these two methods.
We remark that N\textsc{et}R\textsc{ate} and IMINFECTOR are both in favor of training data consisting of cascades with the source sets of size 1 (i.e., only one source node).
In contrary, NMF-InfMax does not have this restriction and thus is more flexible. However, for comparison purpose, we only feed cascade data with single source node to all methods in this experiment.

\paragraph{Experiment setting}
We again use three types of Kronecker graph models: Hierarchical (Hier), Core-periphery (Core) and Random (Rand) networks, and simulate the diffusion processes using exponential distribution with transmission rates randomly sampled from Unif[0.1,1]. 
For each type of network model, we generate two networks of size $n=1,024$ with $d=2$ and $d=4$, respectively.
We sample 100 source nodes, and for each source node we simulate 10 cascades. Hence we have a total of $K$=1,000 cascades for training. 
To train NMF, we set the batch size to 300 and the number of epochs to 50. 
The coefficients of the regularization term on $\Abm$ is set to 0.001, and the rk4 method with 40 time steps is employed as the ODE solver.
To train NMF-InfMax, we set the step size to constant 0.01, and terminate PGD if either the iteration number reaches 500 or the computed influence does not change for 10 consecutive iterations.
In Figure \ref{subfig:infmax_samplesize}, we show the accuracy of NMF-InfMax when the number of training cascades increases from 1,000 to 5,000 for each fixed source set of size from 1 to 10. As we can observe, the accuracy increases significantly when the number of cascades grows from 1,000 to 2,000 but then improvements become insignificant. This suggests that 2,000 cascades is necessary to obtain more accurate influence maximization results for NMF-InfMax. However, due to the limited scalibilty of N\textsc{et}R\textsc{ate} which performs extremely slowly when the number of cascades is over 1,000 and average out-degree is over $4$. We also tested IMINFECTOR with larger training data set, but unlike our method, the accuracy of IMINFECTOR does not improve over 1,000. Hence we still only feed 1,000 cascades to all the compared methods despite that this choice is only in favor of three existing methods.

\begin{figure}[t]
    \begin{center}
    \begin{subfigure}{.32\textwidth}
        \centering\includegraphics[width=\textwidth]{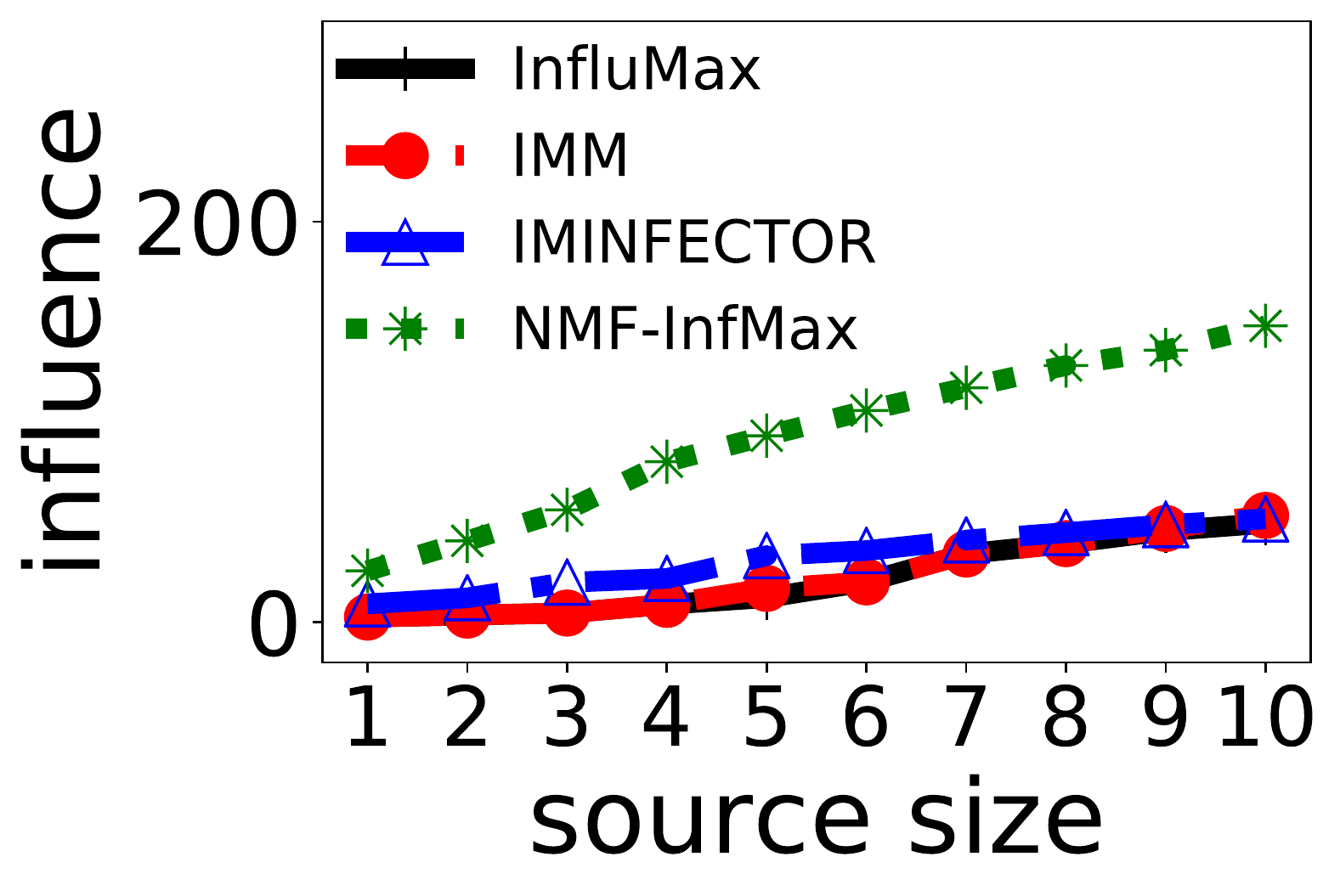}
        %\caption{Hierarchical}
        \label{infmax:he2048}
    \end{subfigure}
    \begin{subfigure}{.32\textwidth}
        \centering\includegraphics[width=\textwidth]{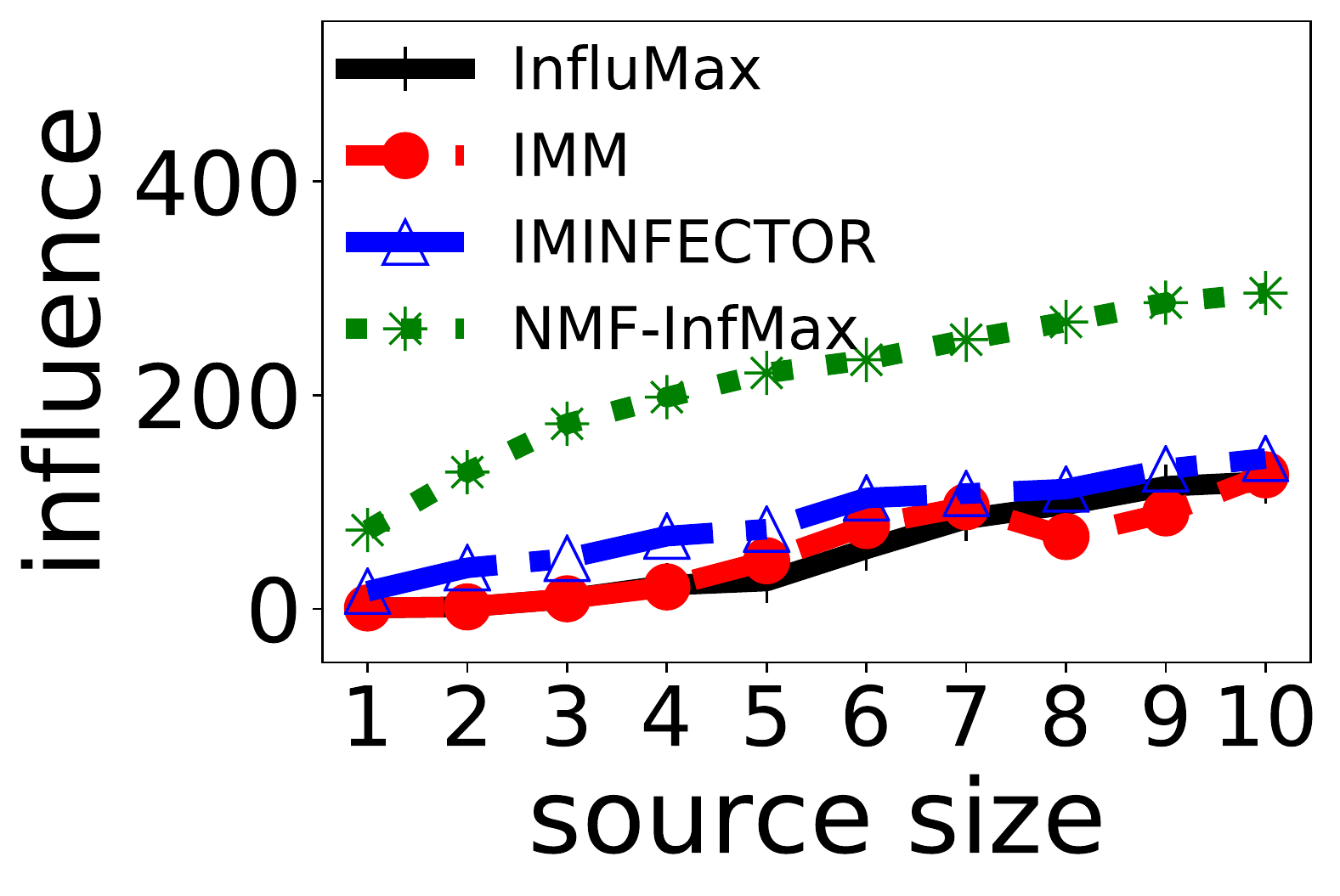}
        %\caption{Random}
        \label{infmax:re2048}
    \end{subfigure}  
    \begin{subfigure}{.32\textwidth}
        \centering\includegraphics[width=\textwidth]{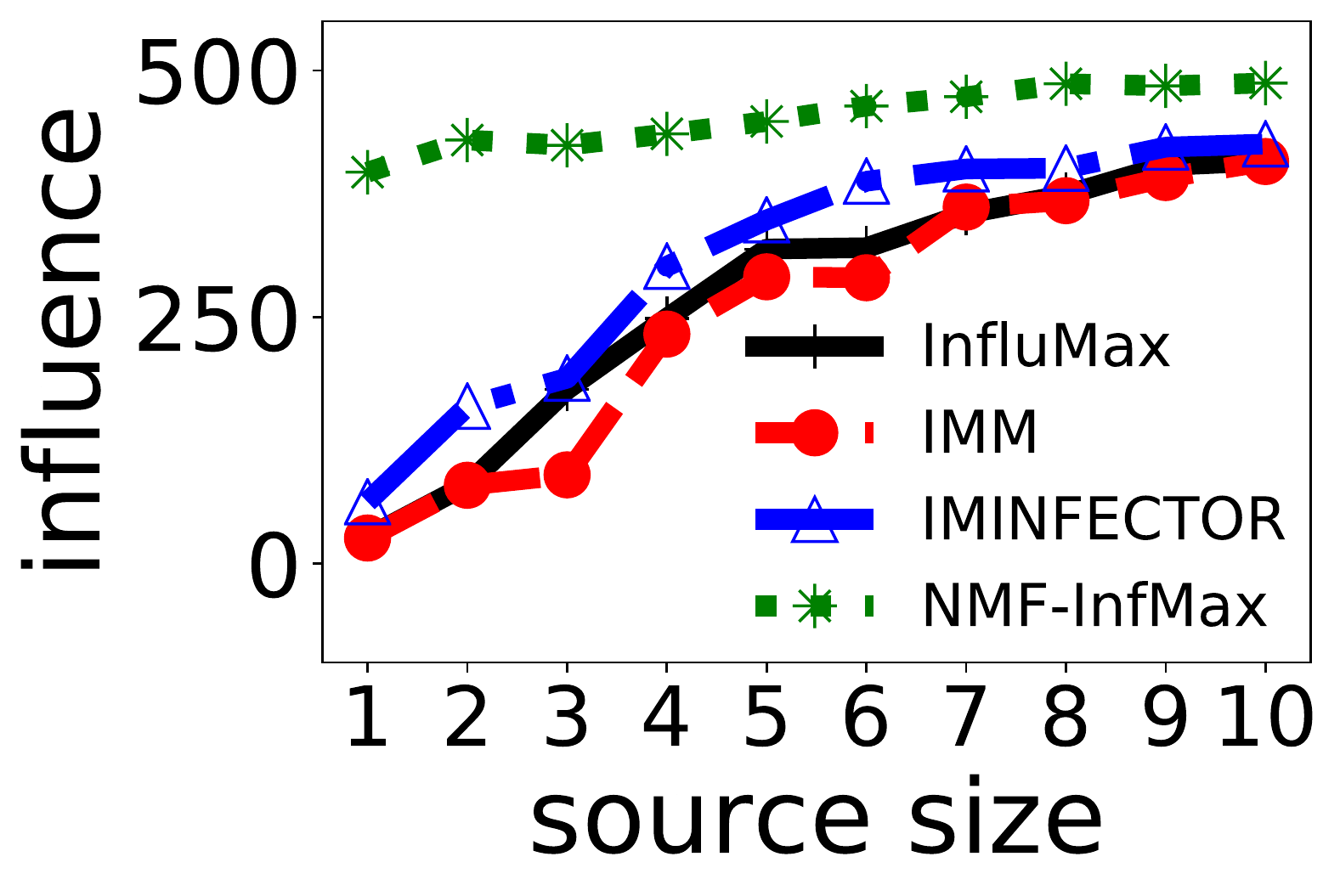}
        %\caption{Core-periphery}
        \label{infmax:cpe2048}
    \end{subfigure}\\
        \begin{subfigure}{.32\textwidth}
        \centering\includegraphics[width=\textwidth]{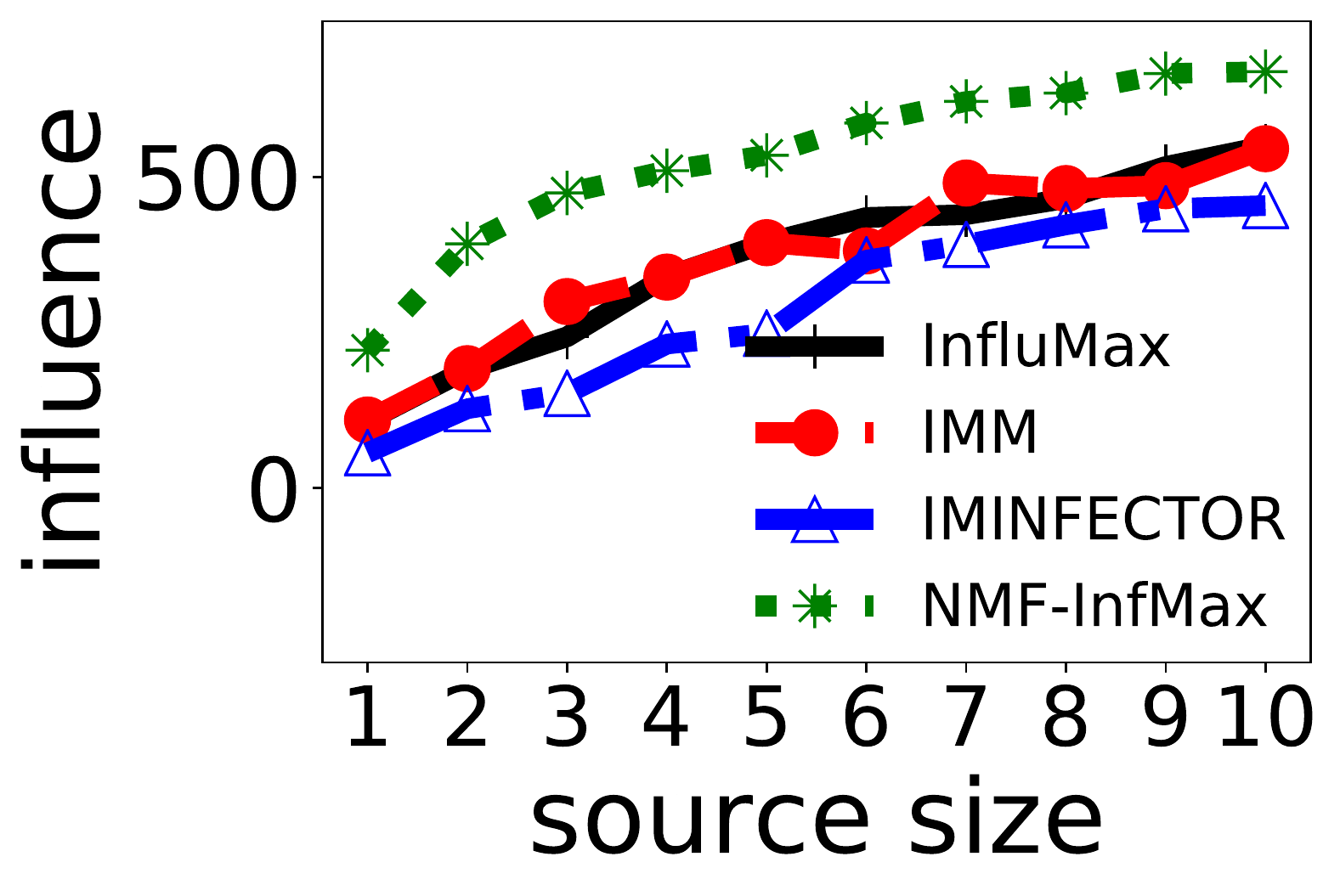}
        \caption{Hierarchical}
        \label{infmax:he4096}
    \end{subfigure}
    \begin{subfigure}{.32\textwidth}
        \centering\includegraphics[width=\textwidth]{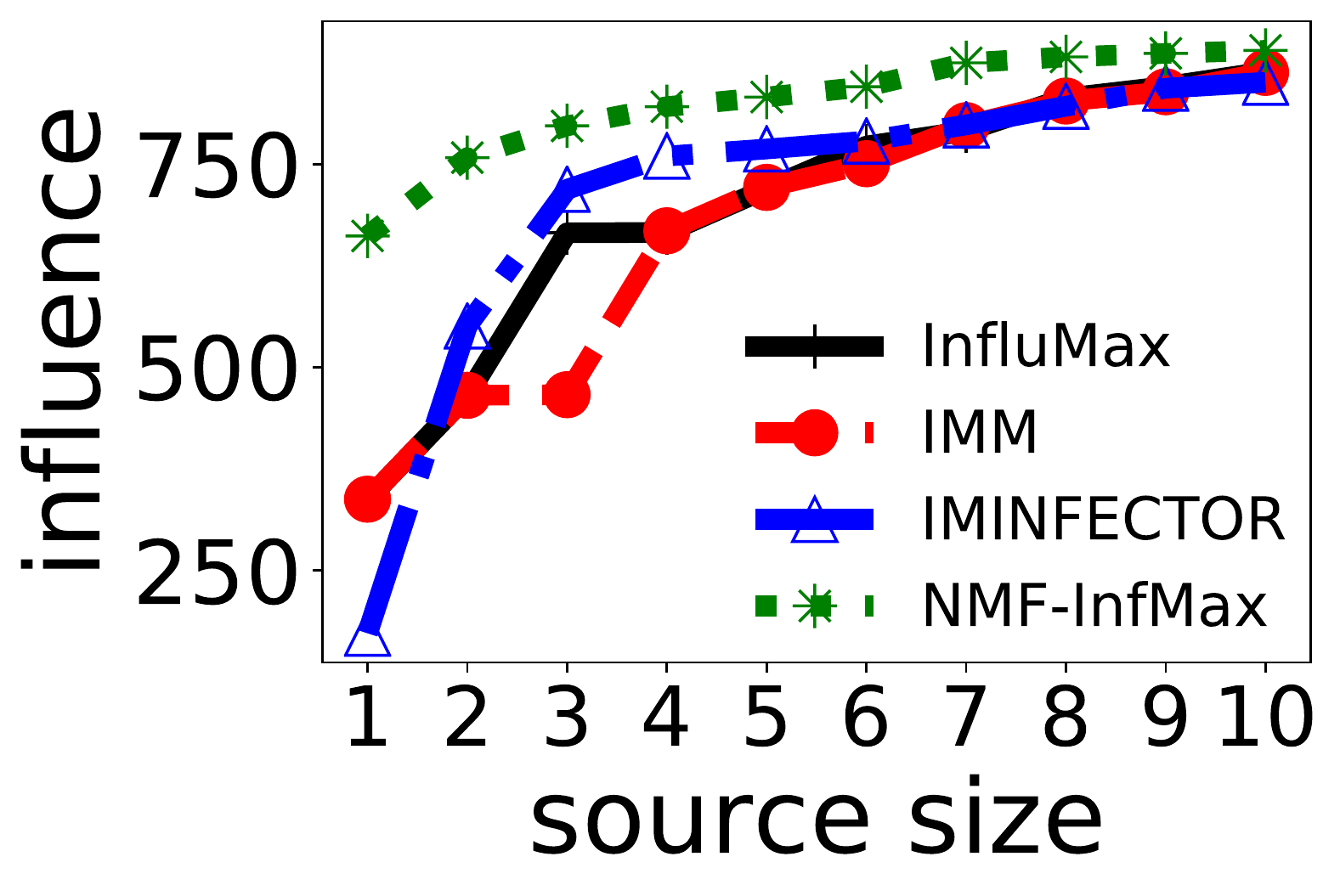}
        \caption{Random}
        \label{infmax:re4096}
    \end{subfigure}  
    \begin{subfigure}{.32\textwidth}
        \centering\includegraphics[width=\textwidth]{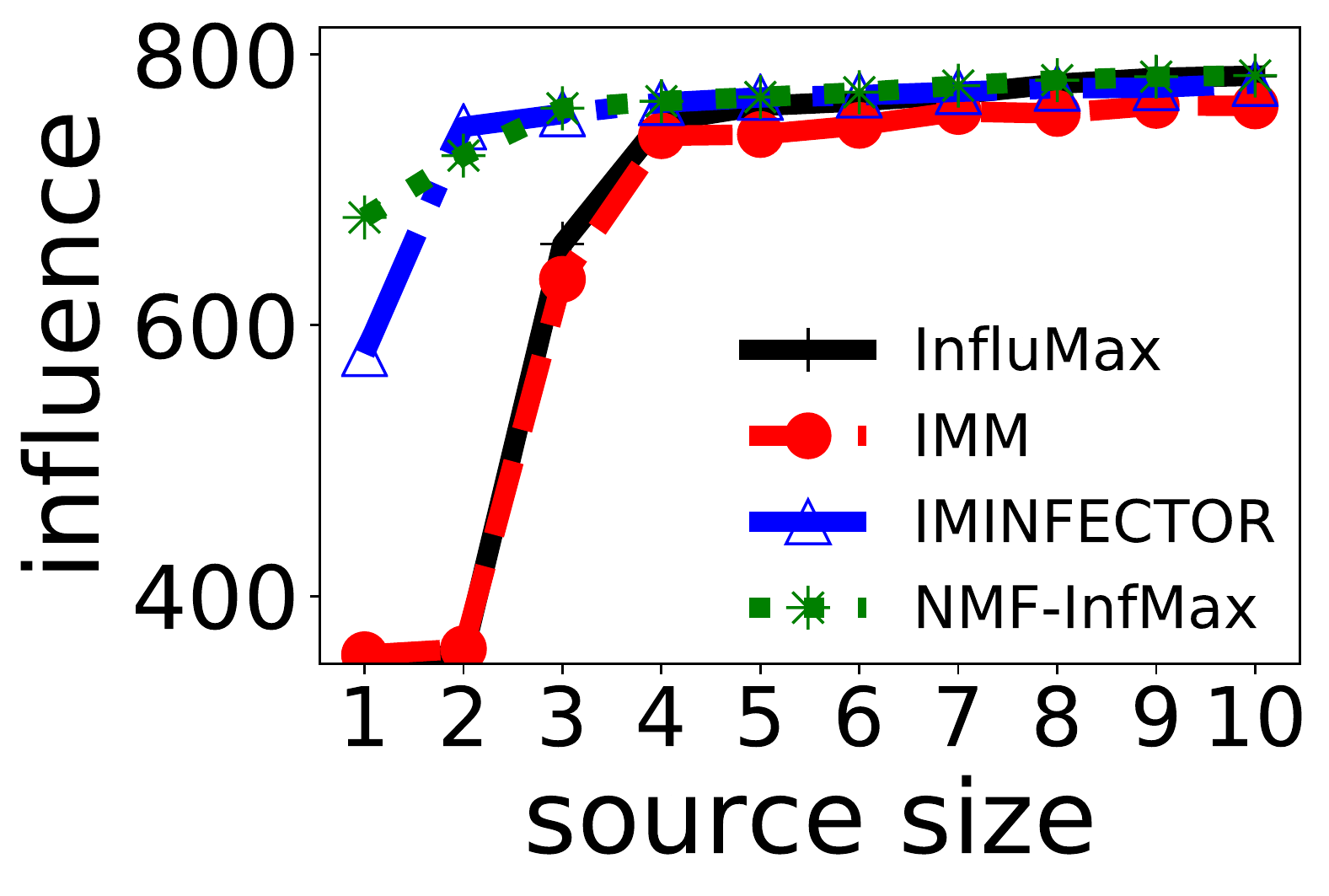}
        \caption{Core-periphery}
        \label{infmax:cpe4096}
    \end{subfigure}
     \end{center}
%%%%%%%%%%%%%%%%%%%%%%%%%%%%%%%%%%%%%%%%%%%%%%%%%%%%%%%%%%%%%%%%%%%%%%%%%%%%%%%%
\caption{Influence of the source sets selected by the compared methods on three different types of networks: (a) Hierarchical, (b) Random, and (c) Core-periphery, with exponential diffusion model at $T=10$ and varying source sizes $n_0$ from 1 to 10. Each network consists of 1024 nodes and 2048 edges (top) or 4096 edges(bottoms).}
\label{fig:infmax2}
\end{figure}

It is also important to note that both InfluMax and IMM require the knowledge of diffusion model given by the shape and scale parameters of edges for the computation of N\textsc{et}R\textsc{ate} and their own. Thus, they are more vulnerable to model mis-specification. In this test, we assume they know the ground truth diffusion model, so they can attain their highest accuracy. However, it is worth noting that the network inference by N\textsc{et}R\textsc{ate} can be very expensive computationally. For example, it took N\textsc{et}R\textsc{ate} up to 160 hours to infer the network structure from 1,000 cascades of a core-periphery network of $n=1,024$ and $d=4$ in Figure  \ref{infmax:cpe4096}. In contrast, the computational cost of IMINFECTOR is very low, but IMINFECTOR is more restrictive on data because it requires that the training cascades contain the nodes to be selected. This may not be feasible in practice. Moreover, the influence maximization results obtained by IMINFECTOR also appear to be worse than that by NMF, as shown below.

\paragraph{Comparison results}
The influence maximization results obtained by the aforementioned algorithms and NMF are shown in Figure \ref{fig:infmax2}. As we can see, NMF-InfMax consistently returns more influential source sets with all budget $n_0$ for all varying network structure, density, and budget. 

%IMM and InfluMax both require network structure to perform influence maximization. When   N\textsc{et}R\textsc{ate} to infer the network structure which is not too scalable for some networks. For example, it takes 577,170s to train the 1000 cascade data of Core-periphery network with 4096 edges to get the results in \ref{infmax:cpe4096}. is the most efficient method, but the candidate nodes to maximize the influence are limited to the source node set of training data while the most influential nodes could show up at any timepoints in the cascades. Moreover, two simple linear models with activation functions are used to rank the importance of nodes which benefits to efficiency but might not to effectiveness. On summary, NMF is the best one to balance efficiency and effectiveness.
%

\paragraph{Real data}
We extract diffusion cascades from the MemeTracker dataset \cite{leskovec2009meme-tracking} which includes 300 million blog posts and articles collected from 5,000 active media sites between March 2011 and February 2012. 
Following \cite{du2014influence}, 
we select the group of cascades with the keyword "apple and jobs"  and then split them as 60\%-train and 40\%-validation for the influence maximization models. 
As the diffusion model of real-world cascade data is unknown, we only test IMINFECTOR and NMF-InfMax. We follow the setting in \cite{du2014influence} to compute the influence of any selected source set: we uniformly sample one cascade from the data for each node in the set and take the union of all sampled cascades as the set of infected nodes. We repeat this process for 1,000 times and take the average as the true influence of the selected set. 
Figure \ref{subfig:infmax_real} shows the result of influence maximization results. In Figure \ref{subfig:infmax_real}, we set $T=10$ and the source size $n_0=10,20,\cdots, 60$, and plot the influence of the source sets selected by IMINFECTOR and NMF-InfMax. As we can see, NMF-InfMax consistently selected more influential combination of nodes that generate greater influence than those selected by IMINFECTOR do.

\section{Related Work}
\label{sec:related}

In this section, we conduct a comprehensive review of the literature related to the present work. There are several topics involved in the proposed method, namely, influence estimation, network inference, and influence maximization, which just emerged within the past decade. These topics are considered independently, and the methods developed are mostly heuristic or sample-demanding. In what follows, we discuss these topics and their related work in order.

\subsection{Influence estimation}
Sampling-based influence estimation methods have been considered for discrete-time and continuous-time diffusion models. 
Discrete-time models assume node infections only occur at discrete time points. Under this setting, the independent cascade (IC) and linear threshold (LT) models are considered and the propagation spread of a source set $S$ is simply estimated by the expected reachable set size of $S$ taken over the randomness of the influence propagation process in \cite{kempe2003maximizing}. 
To improve the efficiency of Monte Carlo simulations used in influence estimation, a method with provable performance guarantee is developed which iterates over a sequence of guesses on the true influence until the verifier accepts in \cite{lucier2015influence}. In \cite{lucier2015influence}, the verifier estimates the influence on multiple sampled graphs using a standard Riemann sum of the influence function, and accepts if this value is close to the guesses. 
In \cite{borgs2013maximizing}, the reverse reachable (RR) sets of nodes are adopted which proved the expected spread equals $n$ times the fraction of sampled RR sets covered by the source set. The sample size is controlled by a given threshold \cite{borgs2013maximizing}, a pre-calculated parameter \cite{tang2014influence}, or some stop conditions \cite{tang2015influence} to achieve a balance between efficiency and accuracy. 
Instead of using the full network structure as the methods above, sketch-based approaches only characterize propagation instances for influence computation, such as the method in \cite{cohen2014sketch-based}, which considers per-node summary structures defined by the bottom-$k$ min-bash \cite{cohen1997size-estimation} sketch of the combined reachability set. 
In contrast to discrete-time models, continuous-time diffusion models allow arbitrary event occurrence times and hence are more accurate in modeling real-world diffusion processes. In continuous-time independent cascade (CIC) models, influence estimation can be reformulated as  the problem of finding the least label list which contains information about the distance to the smallest reachable labels from the source \cite{du2013scalable,gomez-rodriguez2016influence}. Compared to methods using a fixed number of samples, a more scalable approximation scheme with a built-in block is developed to minimize the number of samples needed for the desired accuracy \cite{Nguyen2017outward}.  
Inspired by \cite{tang2015influence}, algorithms proposed in \cite{borgs2013maximizing,tang2015influence,tang2014influence} can be extended from the IC model to other discrete-time models and CIC models by generalizing the definition of RR sets. In \cite{he2020network}, a neural mean-field dynamics approach is proposed, which employs the Mori-Zwanzig (MZ) formalism to derive the node infection probabilities in discrete-time setting.
The influence function can also be approximated by solving a jump stochastic differential equation \cite{zang2020jump} or a deterministic differential equation that governs the evolution of the influence counter \cite{chow2018influence}. 

The aforementioned methods require knowledge of cascade traces \cite{cohen2014sketch-based} or the diffusion networks, such as node connectivity and node-to-node infection rates, as well as various assumptions on the diffusion of interests. However, such knowledge about the diffusion networks may not be available in practice, and the assumptions on the propagation or data formation are often application-specific and do not hold in most other problems. 
InfluLearner \cite{du2014influence} is a state-of-the-art method that does not require knowledge of the underlying diffusion network. InfluLearner estimates the influence directly from cascades data in the CIC models by learning the influence function with a parameterization of the coverage functions using random basis functions. However, the estimation of random basis function suggested by \cite{du2014influence} requires knowledge of the original source node for every infection, which can be difficult or impossible to be tracked in real-world applications, such as epidemic spreads.

In recent years, deep learning techniques have been employed to improve the scalability of influence estimation on large networks. 
In particular, convolutional neural networks (CNNs) and attention mechanism are incorporated with both network structures and user specific features to learn users' latent feature representation in \cite{qiu2018deepinf}. 
By piping represented cascade graphs through a gated recurrent unit (GRU), the future incremental influence of a cascade can be predicted \cite{li2017deepcas}. RNNs and CNNs are also applied to capture the temporal relationships on the user-generated contents networks (e.g., views, likes, comments, reposts) and extract more powerful features in \cite{zhu2020predicting}. In methods based on graph structures, graph neural networks (GNNs) and  graph convolution networks (GCNs) are widely applied. In particular, two coupled GNNs are used to capture the interplay between node activation states and the influence spread \cite{cao2020popularity}, while GCNs integrated with teleport probability from the domain of page rank in \cite{leung2019personalized} enhanced the performance of method in \cite{qiu2018deepinf}. However, these methods depend critically on the structure or content features of cascades which is not available in many real-world applications.

\subsection{Network structure inference}
Inference of diffusion network structure is an important problem closely related to influence estimation. In particular, if the network structure and infections rates are unknown, one often needs to first infer such information from a training dataset of sampled cascades, each of which tracks a series of infection times and locations on the network. Existing methods have been proposed to infer network connectivity \cite{gomez-rodriguez2012inferring,gomez-rodriguez2012submodular,liang2017inferring,du2012learning} and also the infection rates between nodes  \cite{myers2010convexity,gomez-rodriguez2011uncovering,gomez-rodriguez2012structure}. 
Submodular optimization is applied to infer network connectivity \cite{gomez-rodriguez2012inferring,gomez-rodriguez2012submodular,liang2017inferring} by considering the most probable \cite{gomez-rodriguez2012inferring} or all \cite{gomez-rodriguez2012submodular,liang2017inferring} directed trees supported by each cascade. One of the early works that incorporate spatio-temporal factors into network inference is introduced in \cite{liang2017inferring}. 
Utilizing convex optimization, transmission functions \cite{du2012learning}, the prior probability \cite{myers2010convexity}, and the transmission rate \cite{gomez-rodriguez2011uncovering} over edges are inferred from cascades.
In addition to static networks, the infection rates are considered but also in the unobserved dynamic network changing over time \cite{gomez-rodriguez2012structure}. 
Besides cascades, other features of dynamical processes on networks have been used to infer the diffusion network structures.
To avoid using predefined transmission models, the statistical difference of the infection time intervals between nodes in the same cascade versus those not in any cascade was considered in \cite{rong2016model-free}. A given time series of the epidemic prevalence, i.e., the average fraction of infected nodes was applied to discover the underlying network. The recurrent cascading behavior is also explained by integrating a feature vector describing the additional features \cite{wang2012feature-enhanced}.
A graph signal processing (GSP) approach is developed to infer graph structure from dynamics on networks \cite{mateos2019connecting,dong2019learning}.

\subsection{Influence maximization} 
Influence maximization is an important but very challenging problem in real-world applications of diffusion networks, such as commercial advertising and epidemic controls. Influence maximization is shown to be an NP-hard problem under most of diffusion models \cite{li2018influence} (e.g., LT, IC, CIC). It was first formulated in \cite{kempe2003maximizing} as a combinatorial optimization problem. 
Under certain assumptions, the influence function $\sigma(\cdot)$ is a non-negative monotone submodular function, and a standard greedy method \cite{kempe2003maximizing,gomez-rodriguez2016influence} can be applied to obtain provable sub-optimal solution.
Specifically, the greedy method starts from an empty set $\Scal$ and gradually add one node $i$ that maximizes the marginal gain $\sigma(\Scal\cup\{i\})-\sigma(\Scal)$ to $\Scal$.
Note that this requires repeatedly evaluation of influences $\sigma(\Scal)$ which affects the result of influence maximization significantly.

Instead of searching all the nodes in the greedy iterations, a GCN is trained by a probabilistic greedy mechanism, such that it selects a node with probability proportional to its marginal gain to identify noise and predict the node quality for the propagation spread \cite{manchanda2020gcomb}. The computations on the reward for adding the node to set $\Scal$ is performed in another Q-learning network. 
The importance of nodes can also be measured by exploiting the submodularity \cite{goyal2011celf,leskovec2007cost-effective} or only considering one-hop and two-hoop spread benefit measures on nodes in \cite{he2019tifim}.
The influence maximization problem is also modeled as the maximum coverage problem of selecting the budget number of nodes to cover the maximum number of sampled RR sets in \cite{borgs2013maximizing,tang2015influence,tang2014influence}. 
For instances without the information of network structure, the influence relationships between nodes are representation learned from cascade date initiated by a single node to derive a greedy solution in \cite{panagopoulos2020multi-task,panagopoulos2020influence}.

\section{Conclusion}
\label{sec:conclusion}
We propose a novel framework using neural mean-field dynamics for inference and estimation on diffusion networks. Our new framework is derived from the Mori-Zwanzig formalism to obtain exact evolution of node infection probabilities. The Mori-Zwanzig memory can be approximated by convolutions, which renders the system as a delay differential equation for highly interpretable parameterization. Directly using information diffusion cascade data, our framework outperforms many state-of-the-art methods in network structure inference and influence estimation. Our framework can also effectively tackle influence maximization on networks, which is known to be a challenging NP-hard problem. Extensive numerical experiments were conducted to show the promising accuracy and efficiency of the proposed framework on both synthetic and real-world data sets. We expect that the proposed framework can be applied to many other optimization and control problems arising from diffusion network applications, such as optimal campaigning, propagation control, and source identification, which will also be investigated in our future work.

% \newpage

\appendix
\section{Proofs}
\label{app:proof}

\subsection{Proof of Theorem \ref{thm:z_ode}}
\label{subsec:pf_z_ode}
\begin{proof}
Let $\lambda_i^*(t)$ be the conditional intensity of node $i$ at time $t$, i.e., $\ex[\dif X_i(t)|$ $\Hcal(t)]= \lambda_i^*(t) \dif t$. 
In the standard diffusion model, the conditional intensity $\lambda_i^*(t)$ of a healthy node $i$ (i.e., $X_i(t)=0$) is determined by the total infection rate of its infected neighbors $j$ (i.e., $X_j(t)=1$). That is,
\begin{equation}\label{eq:lambda_exp}
\lambda_i^*(t) = \sum_{j} \alpha_{ji} X_j(t) (1 - X_i(t)).
\end{equation}
By taking expectation $\ex_{\Hcal(t)}[\cdot]$ on both sides of \eqref{eq:lambda_exp}, we obtain
\begin{align}
\lambda_i(t) 
:=\ &\ex_{\Hcal(t)}[\lambda_i^*(t)] = \ex_{\Hcal(t)}\sbr[2]{\alpha_{ji} X_j(t) (1 - X_i(t)) \big\vert \Hcal(t)} \nonumber \\
=\ &\sum_{j}\alpha_{ji}(x_j - x_{ij}) = \sum_{j}\alpha_{ji}(x_j - y_{ij} - e_{ij}). \label{eq:lambdai1}
\end{align}
On the other hand, there is
\begin{equation}\label{eq:dx_exp}
\lambda_i(t) \dif t = \ex_{\Hcal(t)}[\lambda_i^*(t)] \dif t = \ex_{\Hcal(t)}[\dif X_i(t) | \Hcal(t)] = \dif \ex_{\Hcal(t)}[X_i(t)|\Hcal(t)] = \dif x_i.
\end{equation}
Combining \eqref{eq:lambdai1} and \eqref{eq:dx_exp} yields
\begin{align*}
x_i' & = \frac{\dif x_i(t)}{\dif t} = \sum_{j} \alpha_{ji} (x_j - y_{ij} - e_{ij}) = (\Abm \xbm)_i - (\diag(\xbm) \Abm \xbm)_i - \sum_{j}\alpha_{ji}e_{ij}
\end{align*}
for every $i\in [n]$, which verifies the $\xbm$ part of \eqref{eq:z_ode}.
Similarly, we can obtain 
\begin{align}\label{eq:xI1}
x_I' 
& =  \sum_{i \in I} \sum_{j \notin I} \alpha_{ji}( x_I - x_{I \cup \{j\}}) =  \sum_{i \in I} \sum_{j \notin I} \alpha_{ji}( y_I + e_I - y_{I \cup \{j\}} - e_{I \cup \{j\}}).
\end{align}
Moreover, by taking derivative on both sides of $x_I(t) = y_I(t) + e_I(t)$, we obtain
\begin{align}\label{eq:xI2}
x_I' = \sum_{i\in I} y_{I \setminus \{i\}} x_i' + e_I' = \sum_{i\in I} y_{I \setminus \{i\}} \sum_{j\ne i} \alpha_{ji}(x_j - x_ix_j - e_{ij}) + e_I'.
\end{align}
Combining \eqref{eq:xI1} and \eqref{eq:xI2} yields the $\ebm$ part of \eqref{eq:z_ode}.

It is clear that $\xbm_0 = {\chibm}_{\Scal}$. For every $I$, at time $t=0$, there is $x_I(0)=\prod_{i\in I} X_i(0)=1$ if $I\subset \Scal$ and $0$ otherwise; and the same for $y_I(0)$. Hence $e_I(0)=x_I(0)-y_I(0)=0$ for all $I$. Hence $\zbm_0 = [\xbm_0; \ebm_0] = [{\chibm}_{\Scal}; \zerobm]$, which verifies the initial condition of \eqref{eq:z_ode}.
\end{proof}

\subsection{Proof of Theorem \ref{thm:mz}}
\label{subsec:pf_mz}
\begin{proof}%[Proof of Theorem \ref{thm:mz}]
Consider the system \eqref{eq:z_ode} over a finite time horizon $[0,T]$, which evolves on a smooth manifold $r \subset \mathbb{R}^N$. For any real-valued phase (observable) space function $g:r \to \mathbb{R}$, the nonlinear system \eqref{eq:z_ode} is equivalent to the linear partial differential equation, known as the Liouville equation:
\begin{equation}\label{eq:liouville}
\begin{cases}
\partial_t u(t,\zbm) = \Lcal [u](t,\zbm), \\
u(0,\zbm) = g(\zbm),
\end{cases}
\end{equation}
where the Liouville operator $\Lcal[u] := \bar{\fbm}(\zbm) \cdot \nabla_{\zbm}u$. The equivalency is in the sense that the solution of \eqref{eq:liouville} satisfies $u(t,\zbm_0) = g(\zbm(t;\zbm_0))$, where $\zbm(t;\zbm_0)$ is the solution to \eqref{eq:z_ode} with initial value $\zbm_0$.

Denote $e^{t\Lcal}$ the Koopman operator associated with $\Lcal$ such that $e^{t\Lcal} g(\zbm_0) = g(\zbm(t))$ where $\zbm(t)$ is the solution of \eqref{eq:z_ode}.
Then $e^{t\Lcal}$ satisfies the semi-group property, i.e.,
\begin{equation}\label{eq:semigroup}
e^{t \Lcal} g (z) = g( e^{t \Lcal} z)
\end{equation}
for all $g$.
On the right hand side of \eqref{eq:semigroup}, $\zbm$ can be interpreted as $\zbm= \iotabm(\zbm) = [\iota_1(\zbm),\dots,\iota_N(\zbm)]$ where $\iota_j(\zbm) = z_j$ for all $j$.

Now consider the projection operator $\Pcal$ as the truncation such that $\Pcal g(\zbm) = \Pcal g(\xbm,\ebm) = g(\xbm,0)$ for any $\zbm = (\xbm,\ebm)$, and its orthogonal complement as $\Qcal = I - \Pcal$ where $I$ is the identity operator. 
Note that $\zbm'(t) = \frac{\dif \zbm(t)}{\dif t} = \frac{\partial }{\partial t} e^{t \Lcal}\zbm_0$, and $\bar{\fbm}(\zbm(t)) = e^{t \Lcal} \fbm(\zbm_0) = e^{t \Lcal} \Lcal \zbm_0$ since $\Lcal \iota_j (\zbm) = \fbm_j(\zbm)$ for all $\zbm$ and $j$. 
Therefore \eqref{eq:z_ode} implies that
\begin{equation}\label{eq:z_ode_equiv}
    \frac{\partial }{\partial t} e^{t \Lcal} \zbm_0 = e^{t \Lcal}\Lcal \zbm_0 = e^{t \Lcal} \Pcal \Lcal \zbm_0 + e^{t \Lcal} \Qcal \Lcal \zbm_0.
\end{equation}
Note that the first term on the right hand side of \eqref{eq:z_ode_equiv} is
\begin{equation} \label{eq:z_ode_rhs1}
    e^{t \Lcal} \Pcal \Lcal \zbm_0 = \Pcal \Lcal e^{t \Lcal} \zbm_0 = \Pcal \Lcal \zbm(t).
\end{equation}
For the second term in \eqref{eq:z_ode_equiv}, we recall that the well-known Dyson's identity for the Koopman operator $\Lcal$ is given by
\begin{equation}\label{eq:dyson}
e^{t \Lcal} = e^{t \Qcal \Lcal} + \int_{0}^{t} e^{s\Lcal} \Pcal \Lcal e^{(t-s)\Qcal \Lcal} \dif s.
\end{equation}
Applying \eqref{eq:dyson} to $\Qcal \Lcal \zbm_0$ yields
\begin{align}
e^{t \Lcal} \Qcal \Lcal \zbm_0  
& = e^{t \Qcal \Lcal} \Qcal \Lcal \zbm_0 + \int_{0}^{t} e^{s\Lcal} \Pcal \Lcal e^{(t-s)\Qcal \Lcal} \Qcal \Lcal \zbm_0\dif s \nonumber \\
& = e^{t \Qcal \Lcal} \Qcal \Lcal \zbm_0 + \int_{0}^{t}  \Pcal \Lcal e^{(t-s)\Qcal \Lcal} \Qcal \Lcal e^{s\Lcal}  \zbm_0\dif s  \label{eq:z_ode_rhs2} \\
& = e^{t \Qcal \Lcal} \Qcal \Lcal \zbm_0 + \int_{0}^{t}  \Pcal \Lcal e^{(t-s)\Qcal \Lcal} \Qcal \Lcal \zbm(s)\dif s \nonumber. 
\end{align}
Substituting \eqref{eq:z_ode_rhs1} and \eqref{eq:z_ode_rhs2} into \eqref{eq:z_ode_equiv}, we obtain
\begin{equation}\label{eq:z_ode_equiv2}
    \frac{\partial }{\partial t} e^{t \Lcal} \zbm_0 = \Pcal \Lcal \zbm(t) + e^{t \Qcal \Lcal} \Qcal \Lcal \zbm_0 + \int_{0}^{t}  \Pcal \Lcal e^{(t-s)\Qcal \Lcal} \Qcal \Lcal \zbm(s)\dif s,
\end{equation}
where we used the fact that $e^{t \Lcal} \Pcal \Lcal \zbm_0 = \Pcal \Lcal e^{t \Lcal} \zbm_0 = \Pcal \Lcal \zbm(t)$.
Denote $\phibm(t,\zbm) := e^{t \Lcal} \Qcal \Lcal \zbm$, then we simplify \eqref{eq:z_ode_equiv2} into
\begin{equation}
    \frac{\partial }{\partial t} e^{t \Lcal} \zbm_0 = \Pcal \Lcal \zbm(t) + \phibm(t,\zbm_0) + \int_{0}^{t}  \kbm(t-s, \zbm(s))\dif s,
\end{equation}
where $\kbm(t,\zbm) := \Pcal \Lcal \phibm(t,\zbm) = \Pcal \Lcal e^{t \Lcal} \Qcal \Lcal \zbm $.

Now consider the evolution of $\phibm(t,\zbm)$, which is given by
\begin{equation} \label{eq:phi_ode}
    \partial_t \phibm(t,\zbm_0) = \Qcal \Lcal \phibm(t,\zbm_0),
\end{equation}
with initial condition $\phibm(0,\zbm_0) = \Qcal \Lcal \zbm_0 = \Lcal\zbm_0 - \Pcal \Lcal \zbm_0 = \bar{\fbm}(\xbm_0,\ebm_0) - \bar{\fbm}(\xbm_0,\zerobm) = \zerobm$ since $\ebm_0 = \zerobm$.
Applying $\Pcal $ on both sides of \eqref{eq:phi_ode} yields
\begin{equation*}
    \partial_t \Pcal \phibm(t,\zbm_0) = \Pcal \Qcal \Lcal \phibm(t,\zbm_0) = \zerobm,
\end{equation*}
with initial $\Pcal \phibm(0,\zbm_0) = \zerobm$. This implies that $\Pcal \phibm(t,\zbm_0) = \zerobm$ for all $t$.
Hence, applying $\Pcal$ to both sides of \eqref{eq:z_ode_equiv2} yields
\begin{equation}\label{eq:mz_full}
\frac{\partial }{\partial t}\Pcal \zbm(t) = \frac{\partial}{\partial t}\Pcal e^{t \Lcal}\zbm_0 = \Pcal \Lcal \zbm(t) + \int_{0}^{t} \Pcal \kbm(t-s,\zbm(s)) \dif s.
\end{equation}
Restricting to the first $n$ components, $\Pcal \zbm(t)$ reduces to $\xbm(t)$ and $\Pcal \kbm(t-s,\zbm(s))$ reduces to $\kbm(t-s,\xbm(s))$.
Recalling that $\Pcal \Lcal \zbm(t) = \Pcal \bar{\fbm}(\zbm(t)) = \bar{\fbm}(\xbm(t),\zerobm) = \fbm(\xbm(t))$ completes the proof.
\end{proof}

\subsection{Proof of Proposition \ref{prop:rnn}}
\label{subsec:pf_rnn}

\begin{proof}%[Proof of Theorem \ref{prop:rnn}]
From the definition of $\hbm(t)$ in \eqref{eq:h}, we obtain
\begin{equation}\label{eq:h}
\hbm(t) = \int_{0}^t \Kbm(t-s;\wbm)\xbm(s)\dif s = \int_{-\infty}^t \Kbm(t-s;\wbm)\xbm(s)\dif s = \int_{0}^{\infty} \Kbm(s;\wbm)\xbm(t-s)\dif s
\end{equation}
where we used the fact that $\xbm(t)=0$ for $t<0$.
Taking derivative on both sides of \eqref{eq:h} yields
\begin{align*}
\hbm'(t) 
& = \int_{0}^{\infty} \Kbm(s;\wbm) \xbm'(t-s) \dif s = \int_{0}^{\infty} \Kbm(s;\wbm) \tilde{\fbm}(\xbm(t-s),\hbm(t-s); \Abm, \etabm) \dif s \\
& = \int_{-\infty}^t \Kbm(t-s;\wbm) \tilde{\fbm}(\xbm(s),\hbm(s); \Abm, \etabm) \dif s = \int_{0}^t \Kbm(t-s;\wbm) \tilde{\fbm}(\xbm(s),\hbm(s); \Abm, \etabm) \dif s
\end{align*}
where we used the fact that $\xbm'(t)=\tilde{\fbm}(\xbm(t),\hbm(t);\Abm,\etabm)=0$ for $t<0$ in the last equality.

If $\Kbm(t; \wbm) = \sum_{l}\Bbm_l e^{-\Cbm_l t}$, then we can take derivative of \eqref{eq:h} and readily deduce that $\hbm' = \sum_{l=1}^L (\Bbm_l \xbm - \Cbm_l \hbm)$.
\end{proof}

\subsection{Proof of Theorem \ref{thm:pmp}}
\label{subsec:pf_pmp}

\begin{proof}%[Proof of Theorem \ref{thm:pmp}]
Let $\zetabm \in \mathbb{R}^{m}$ and $\varepsilon \ge 0$ be arbitrary. 
Consider the variation of any control $\thetabm$ given by $\thetabm_\varepsilon:=\thetabm + \varepsilon \zetabm$ and denote $\mbm_{\varepsilon} (t)$ the state process following \eqref{eq:oc_m} with $\thetabm_\varepsilon$. Then we have
\[
\mbm_{\varepsilon}(t)=\mbm(t)+\varepsilon\ybm(t)+o(\varepsilon),\qquad 0\leq t\leq T,
\]
where the first-order perturbation $\ybm(t)$ satisfies 
\[
\begin{cases}
\ybm'(t) = \nabla_{\mbm}\gbm(\mbm(t);\thetabm) \ybm(t)+\nabla_{\thetabm}\gbm(\mbm(t);\thetabm) \zetabm, \qquad 0 \leq t \leq T, \\
\ybm(0) = \zerobm.
\end{cases}
\]
Therefore, the directional derivative of $\ell$ defined in \eqref{eq:oc_obj} at $\thetabm$ along the direction $\zetabm$ is
\begin{align}
\label{eq:pos}
  \frac{\dif }{\dif \varepsilon}\ell (\thetabm_\varepsilon) \Big\vert_{\varepsilon=0}
  &=\int_0^T \del[2]{\nabla_{\mbm}r(\mbm_t,\thetabm) \ybm(t) + \nabla_{\thetabm}r(\mbm(t),\thetabm)\zetabm} \dif t
  +\pbm(T)\ybm(T).
\end{align}
On the other hand, we have
\begin{align*}
  (\pbm\cdot\ybm)'
  = \pbm' \cdot \ybm + \pbm \cdot \ybm' & = -\del[2]{\nabla_{\mbm} \gbm(\mbm(t); \thetabm) \pbm(t) + \nabla_{\mbm} r(\mbm(t),\thetabm)} \cdot \ybm \\
  &\quad + \pbm \cdot \del[2]{\nabla_{\mbm}\gbm(\mbm(t);\thetabm)^{\top} \ybm(t)+\nabla_{\thetabm}\gbm(\mbm(t);\thetabm)^{\top} \zetabm} \\
  & =-\nabla_{\mbm} r(\mbm(t), \thetabm)^{\top}\ybm(t)+\pbm(t)^\top\nabla_{\thetabm}\gbm(\mbm(t);\thetabm)\zetabm.
\end{align*}
Since $\ybm(0) = \zerobm$, we know
\begin{align}\label{eq:new}
  \pbm(T)\cdot\ybm(T)
  &= \int_0^T \del[2]{ -\nabla_{\mbm} r(\thetabm,\mbm(t))^{\top}\ybm(t)+\pbm(t)^\top\nabla_{\thetabm}\gbm(\mbm(t);\thetabm)\zetabm} \dif t. 
\end{align}
Substituting \eqref{eq:new} into \eqref{eq:pos} yields
\[
\frac{\dif}{\dif \varepsilon}\ell(\thetabm_\varepsilon)\Big\vert_{\varepsilon=0}=\cbr[2]{\int_0^T \del{\nabla_{\thetabm}\gbm(\mbm(t);\thetabm)^{\top}\pbm(t) + \nabla_{\thetabm}r(\thetabm,\mbm(t))}  \dif t }\cdot \zetabm.
\]
As $\zetabm$ is arbitrary, we know that the gradient $\nabla_{\thetabm} \ell(\thetabm)$ is as claimed in \eqref{eq:grad-P}. 

Note that the integrand in \eqref{eq:grad-P} is
\begin{equation*}
    \del{\nabla_{\thetabm}r(\thetabm,\mbm(t))+ \nabla_{\thetabm}\gbm(\mbm(t);\thetabm)^{\top}\pbm(t) }= \nabla_{\thetabm} H(\mbm(t), \pbm(t); \thetabm).
\end{equation*}
Hence, at the optimal $\thetabm^*$ of $\ell$, we have
\begin{align}\label{eq:min}
    \frac{\dif}{\dif \varepsilon}\ell(\thetabm^*_\varepsilon)\Big\vert_{\varepsilon=0} =   \del[2]{\int_0^T \nabla_{\thetabm} H(\mbm^*(t),\pbm^*(t);\thetabm^*) \dif t } \cdot \zetabm \geq 0,
\end{align}
for all $\zetabm \in \mathbb{R}^{m}$, from which we readily deduce the identity regarding $H$ at $\thetabm^*$.
\end{proof}

\subsection{Proof of Theorem \ref{thm:infmax-grad}}
\label{subsec:pf_infmax-grad}

\begin{proof}
Let $\vbm \in \mathbb{R}^n$ be arbitrary and consider the variation $\ubm_{\epsilon}:=\ubm+ \epsilon \vbm+o(\epsilon)$ of $\ubm$ with $\epsilon>0$. 
Let $\xbm_\epsilon(t)$ be the $\xbm$-part of the solution $\mbm_{\epsilon}(t)$ to \eqref{eq:oc_m} with initial $[\ubm_{\epsilon};\zerobm]$.
Suppose $\xbm_\epsilon(t) = \xbm(t) + \epsilon \wbm(t) + o(\epsilon)$ for all $t \in [0,T]$ as $\epsilon\to 0$, then $\wbm(t)$ solves
\begin{equation}
\label{eq:infmax-w}
\begin{cases}
\wbm'(t) = \nabla_{\xbm} \gbm_{\xbm}(\mbm(t);\thetabm) \wbm(t),\quad 0 \le t \le T, \\
\wbm(0) = \vbm.
\end{cases}
\end{equation}
Note that \eqref{eq:infmax-w} is a linear ODE of $\wbm$ and thus has an analytic solution as follows:
\[
\wbm(T) = e^{\int_0^T \nabla_{\xbm} \gbm_{\xbm}(\mbm(t);\thetabm) \dif t} \vbm.
\]
Next, we compute the directional derivative of $L$ defined in \eqref{eq:im_relax} at $\ubm$ along direction $\vbm$:
\begin{align*}
\frac{\dif}{\dif \epsilon} L(\ubm_{\epsilon})\Big|_{\epsilon = 0} 
& = \frac{\dif}{\dif \epsilon} \del[2]{\Rcal(\ubm_{\epsilon}) - \onebm \cdot \xbm_{\epsilon}(t)}\Big|_{\epsilon = 0} \\
& = \nabla_{\ubm} \Rcal(\ubm)\cdot \vbm-\onebm \cdot \wbm(T) \\
& =\del[2]{\nabla_{\ubm} \Rcal(\ubm)-e^{\int_0^T \nabla_{\xbm} \gbm_{\xbm}(\xbm(t);\thetabm)^{\top} \dif t} \onebm } \cdot \vbm.    
\end{align*}
As $\vbm$ is arbitrary, we know the gradient $\nabla_{\ubm} L(\ubm)$ is
\begin{align}\label{eq:ex_sol}
    \nabla_{\ubm} L(\ubm) = \nabla_{\ubm} \Rcal(\ubm)- e^{\int_0^T \nabla_{\xbm} \gbm_{\xbm}(\xbm(t);\thetabm)^{\top} \dif t} \onebm.
\end{align}
It is clear that the second term on the right hand side of \eqref{eq:ex_sol} is $\sbm(T)$ solved from
\begin{align}\label{eq:forward}
  \begin{cases}
\sbm'(t) = \nabla_{\xbm} \gbm_{\xbm}(\xbm(t);\thetabm)^{\top} \sbm(t),\quad 0\leq t\leq T, \\
\sbm(0) = \onebm.
\end{cases} 
\end{align}
This completes the proof.
\end{proof}

\bibliographystyle{abbrv}
\bibliography{nmf_ref}

\end{document}